\documentclass[letterpaper]{article} 
\usepackage{aaai23}  
\usepackage{times}  
\usepackage{helvet}  
\usepackage{courier}  
\usepackage[hyphens]{url}  
\usepackage{graphicx} 
\usepackage{natbib}  
\usepackage{caption} 
\usepackage{algorithm}
\usepackage{algorithmic}
\usepackage{subcaption}

\usepackage{newfloat}
\usepackage{listings}
\usepackage{amsmath}
\usepackage{amsfonts}
\usepackage{amsthm}
\PassOptionsToPackage{options}{natbib}
\DeclareCaptionStyle{ruled}{labelfont=normalfont,labelsep=colon,strut=off} 
\floatstyle{ruled}
\newfloat{listing}{tb}{lst}{}
\floatname{listing}{Listing}
\newcommand{\bv}{\boldsymbol{v}}
\newcommand{\bx}{\boldsymbol{x}}
\newcommand{\by}{\boldsymbol{y}}
\newcommand{\norm}[1]{\| #1\|}

\newcommand{\rd}{\mathrm{d}}
\newcommand{\half}{\frac{1}{2}}

\newcommand{\mL}{\mathcal{L}}
\newcommand{\mN}{\mathcal{P}_{NN}}
\newcommand{\mU}{\mathcal{U}}
\newcommand{\mP}{\mathcal{P}}
\newcommand{\mX}{\mathcal{X}}

\newcommand{\Lop}{\mathcal{L}}

\newcommand{\eps}{\varepsilon}
\newcommand{\average}[1]{ \left\langle#1 \right\rangle}
\newcommand{\RN}[1]{\textup{\uppercase\expandafter{\romannumeral#1}}
}

\theoremstyle{definition} 
\newtheorem{remark}{Remark}
\newtheorem{theorem}{Theorem}

\usepackage[utf8]{inputenc} 
\usepackage[T1]{fontenc}    
\usepackage{hyperref}       
\usepackage{cleveref}
\usepackage{url}            
\usepackage{booktabs}       
\usepackage{amsfonts}       
\usepackage{nicefrac}       
\usepackage{microtype}      
\usepackage{xcolor}         

\title{Transfer Learning Enhanced DeepONet for Long-Time Prediction of Evolution Equations}
\author {
	Wuzhe Xu,\textsuperscript{\rm 1}\thanks{Wuzhe Xu is the corresponding author.}
	Yulong Lu, \textsuperscript{\rm 1}
	Li Wang \textsuperscript{\rm 2}
}
\affiliations {
	\textsuperscript{\rm 1} Department of Mathematics and Statistics, University of Massachusetts Amherst\\
	\textsuperscript{\rm 2} School of Mathematics, University of Minnesota\\
	wuzhexu@umass.edu, yulonglu@umass.edu, wang8818@umn.edu
}

\usepackage{bibentry}

\begin{document}
	
\nocopyright

\maketitle

\begin{abstract}
	Deep operator network (DeepONet) has demonstrated great success in various learning tasks, including learning solution operators of partial differential equations. In particular, it provides an efficient approach to predict the evolution equations in a finite time horizon. Nevertheless, the vanilla DeepONet suffers from the issue of stability degradation in the long-time prediction. This paper proposes a {\em transfer-learning} aided DeepONet to enhance the stability. Our idea is to use transfer learning to
	sequentially update the DeepONets as the surrogates for propagators learned in different time frames. The evolving DeepONets can better track the varying complexities of the evolution equations, while only need to be updated by efficient training of a tiny fraction of the operator networks. Through systematic experiments, we show that the proposed method not only improves the long-time accuracy of DeepONet while maintaining similar computational cost but also substantially reduces the sample size of the training set. 
\end{abstract}

\section{Introduction}
Solving partial differential equations (PDEs) through deep learning approach has attracted extensive attention recently. Thanks to the universal approximation theorem of neural networks, it is natural to approximate solutions of PDEs using neural network. Many popular neural network based methods have been proposed recently, such as Deep Ritz Method \cite{yu2018deep}, Deep Galerkin Method \cite{sirignano2018dgm}, Physics Informed Neural Networks (PINNs) \cite{raissi2019physics} and the Weak Adversarial Networks \cite{zang2020weak}. In spite of the great success of these methods in solving various PDEs, the neural networks need to be re-trained if one seeks solutions corresponding to different initial conditions (ICs), boundary conditions (BCs) or parameters for the same PDEs. Instead, the recently proposed parametric operator learning methods, such as DeepONet \cite{lu2019deeponet} and FNO \cite{li2020fourier} enable learning of PDEs corresponding to varying BCs or ICs without re-training the networks. 
However, there is one important caveat in the aforementioned operator neural networks. Namely they are essentially supervised learning and often require solving large number of PDEs to form the training data, which can be extremely expansive, especially when PDEs of interest lie in high dimensional spaces. To overcome this issue, Wang et al. \cite{wang2021learning,wang2021long} proposed the physics-informed DeepONet, which uses only the physical information (for instance the governing law of the PDEs) to construct loss function and thus making DeepONet self-supervised. Nevertheless, in practice the physics-informed DeepONets are more difficult to train compared to its vanilla version since the exact differential operators act on the networks and make the convergence behavior highly depends on the underlying physics problem.

Recently DeepONet has also been applied to learning the propagators of evolution equations; see e.g. \cite{liu2022deeppropnet,wang2021long}. The basic idea is to employ DeepONets to learn the solution operator of a PDE within a short time interval subject to a collection of (random) initial conditions. The solution of the PDE at later times can be computed as recursive actions of the trained network operator on solutions obtained at the prior steps. However, the approximation accuracy of solutions can deteriorate in the long-run for at least two reasons. First, due to the approximation error, the trained DeepONet, as a surrogate propagator, may be expansive even if the exact propagator is non-expansive, which leads to the accumulation of approximation error in time and hence makes it difficult to predict the solution in the long-run. Second, during the time-evolution of PDEs, the functions that a propagator inputs and outputs can vary in time, even though the form of the propagator within a fixed time-slot may remain unchanged (e.g. when the dynamics is autonomous).  Taking diffusion equation as an example, one observes that the functions in the range space of the propagator or the semigroup are much smoother  than those in the domain, and for this reason, the solutions in later times become increasingly more regular than those in earlier times. Furthermore, some evolution equations may develop various complexities in a long time-horizon, such as turbulence and scale separations. For those equations,  iterating a DeepONet surrogate that is usually only trained in a single (short) time-frame using a finite collection of initial functions may fail to capture the correct regularity or complexity of the solutions in the long time. 

Transfer learning \cite{bozinovski1976influence,do2005transfer} is an important class of  machine learning techniques that use one neural network trained for one task for a new neural network trained for a different related task. The idea is that the knowledge or important features of one problem gained by training the former neural nets  can be transferred to other problems. Transfer learning has been widely used in image recognition \cite{yin2019feature,jin2020deep}, natural language processing \cite{ruder2019transfer} and recently in PINNs \cite{goswami2019transfer,obiols2021surfnet,song2022transfer,desai2021one}. To the best of our knowledge, the present work is the first work to employ transfer learning for learning solution operators of evolutionary PDEs.

\subsection{Our contributions}
We propose a novel physics-informed DeepONet approach based on the transfer learning for predicting time-dependent PDEs. Different from the existing usage of DeepONets in learning the propagators of PDEs where the learned propagators are treated constant in time, we use transfer learning to sequentially update the learned propagators as time evolves. The resulting time-changing DeepONets offer several advantages compared to the vanilla counterparts: (1) the evolving DeepONets can better adapt to the varying complexities associated to the evolution equations; (2) the DeepONets are updated in a computationally efficient way that the hidden layers are frozen once trained and only the parameters in the last layer are re-trained. 

We hereby highlight the major contributions of the proposed method:\begin{itemize}
	\item Time marching with the transfer-learning tuned DeepONet gives more accurate and robust long-time prediction of solutions of PDEs while still maintaining low computational cost.

	\item The proposed method is applied to various types of evolutionary PDEs, including the reaction diffusion equations, Allen-Cahn and Cahn-Hilliard equations, the Navie-Stokes equation and multiscale linear radiative transfer equations. 
	
	\item Through extensive numerical results, we show that  our method can significantly reduce the training sample size needed by DeepONet to achieve the same (or even higher) accuracy. 

\end{itemize}

\subsection{Related works}\label{sec:relatedwork}
Transfer-learning has been previously combined with physics informed neural networks for solving PDEs problems arising from diverse fields, including the phase-field modeling of fracture \cite{goswami2019transfer}, super-resolution of turbulent flows \cite{obiols2021surfnet}, training of CNNs on multi-fidelity data (e.g. multi-resolution images of PDE solutions on fine and coarse meshes) \cite{song2022transfer}, etc. In \cite{chakraborty2022domain}, transfer-learning was also applied 
as a domain adaption method for learning solutions of PDEs defined on complex geometries. The recent paper \cite{desai2021one} proposed a one-shot transfer learning strategy that freezes the hidden layers of a pre-trained PINN and reduces the training neural networks for solving new differential equations to optimizing only the last (linear) layer. This approach  eliminates the need of re-training the whole network parameters while still produces  high-quality solutions by tuning a small fraction of parameters in the last layer.  The present paper marry this transfer learning idea with  DeepONet for learning the propagators of evolution equations in order to predict the long time evolution. 

While we are finalizing the current paper, we are aware of a recent preprint  \cite{goswami2022deep} where transfer learning was exploited together with DeepONet for learning  PDEs under conditional shift. The purpose there is to train a source PDE model with sufficient labeled data from one source domain and transfer the learned parameter to a target domain with limited labeled data. The technology developed there is mainly applied for transferring the knowledge of a solution operator trained on a system of PDEs from one domain to another. Different from \cite{goswami2022deep}, we leverage transfer learning to successively tuning the surrogate models of  propagators learned via physics-informed DeepONet so that the tuned operator networks can adaptively track the evolving propagators that carry evolving inputs and outputs. 
The proposed approach is proven to be more accurate and robust for learning the long-time evolution of PDEs. 

\section{Numerical method}\label{sec:numerical_method}

\paragraph{Problem set-up}
Consider the initial boundary value problem for a general evolution equation:
\begin{equation} \label{111}
	\left\{\begin{array}{l}
		\partial_t f(t, \bx) = \mathcal{L}(f(t, \bx))  , \\
		f(t, \bx) = \phi(\bx), ~ \bx \in \partial \Omega_x, \\
		f(0, \bx) = f_0(\bx)\,, ~ \bx \in  \Omega_x\,.
	\end{array}\right.
\end{equation}
Throughout the paper, we assume that the equations are dissipative in the sense that $\int_{\Omega_x} f \mathcal{L} f d\bx \leq 0$. 
Given a time step size $\Delta t$, we consider the semi-discrete approximation $f^n(\bx)$ of the solution $f(n \Delta t,\bx)$ to \eqref{111} defined by the backward Euler discretization: 
\begin{equation}\label{eqn:mP}
	f^{n+1}(\bx) = (I - \Delta t \mathcal{L})^{-1}f^n(\bx)
	:=\mathcal{P}^{\Delta t} f^n(\bx)\,.
\end{equation}
Our goal is to approximate the propagator 
\[
\mP^{\Delta t}: f^n(\bx) \mapsto f^{n+1}(\bx)
\]
by an operator neural network $\mP_{NN}$ so that only one forward pass of the neural network achieves time-marching solutions from one step to the next, and that the evolution dynamics can be captured in a long time-horizon. 

It is important to point out that the backward Euler scheme is not the only choice for time-marching. One can extend it to high order time discretization schemes such as Runge-Kutta methods, as long as $\Delta t$ is chosen such that $\mP^{\Delta t}$ is a non-expanding operator. We will make this point more clear in Section 3 and Appendix~\ref{2nd}. To ease the notation, the superscript $n$, $n+1$ and $\Delta t$ will be omitted in the following context if it does not cause any confusion.  

\subsection{Physics-informed DeepONet}

\begin{figure}[th!] 
	\centering
	{\includegraphics[width=0.3\textwidth, height=0.15\textheight]{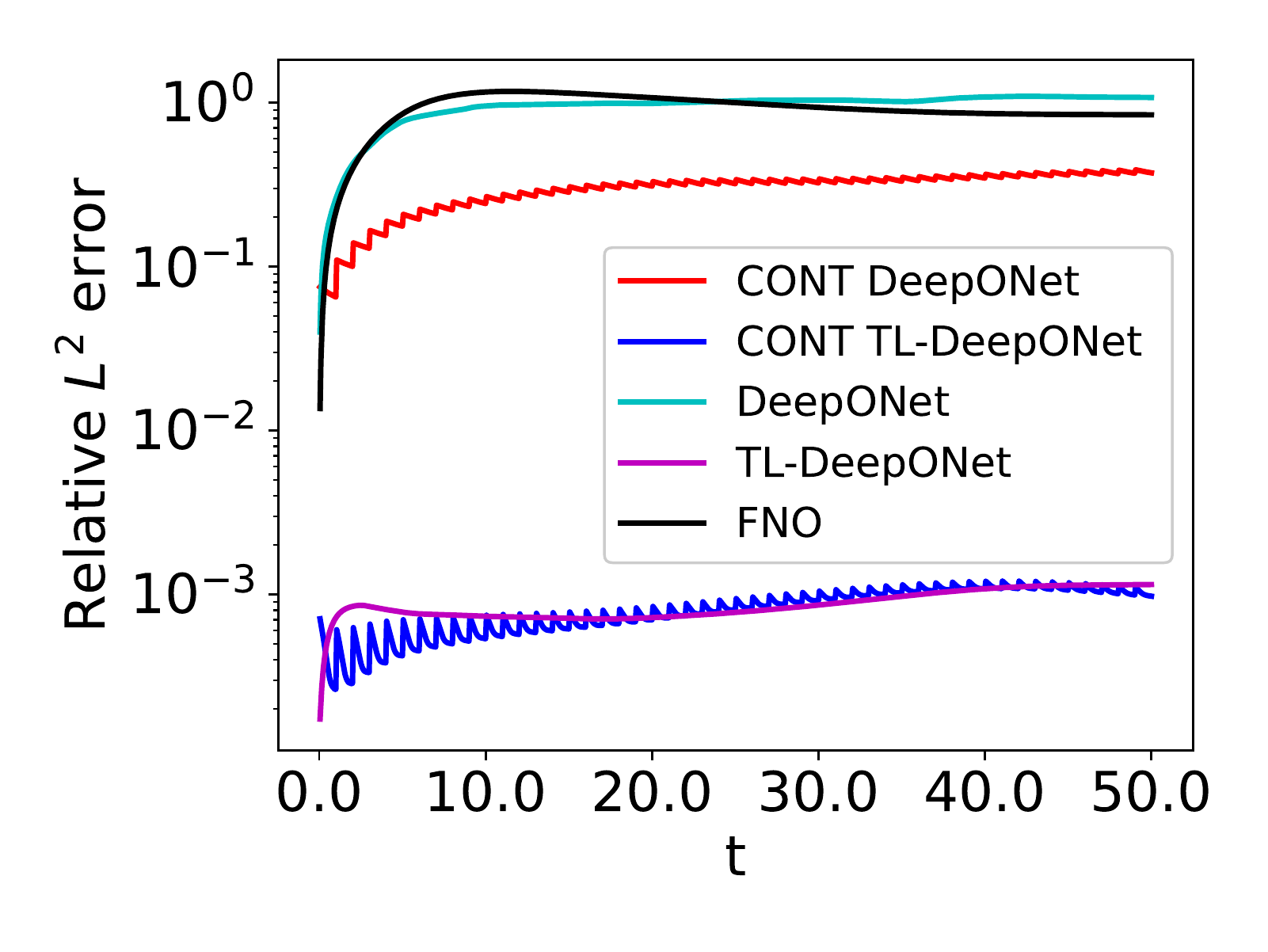}}
	\caption{The relative $L^2$ error in time (defined in $\eqref{eqn:avg_l2_loss_t}$) for 1D Allen Cahn $\eqref{eqn:ac}$. The networks with ``CONT" refer to networks trained by using \eqref{eqn:bc_loss} whereas the others are trained by using  \eqref{eqn:emp_loss}. The prefix ``TL" means tuned by transfer learning. See implementation details in Appendix \ref{sec:detail_ac_ch}.}

\label{fig:failure}
\end{figure}

Let $\Omega_x$ be a compact set in $\mathbb R^d$ and let $\mathcal{X}$ be a compact subspace of the space $C(\Omega_x)$ of continuous function defined on $\Omega_x$. Then according to the universal approximation theorem \cite{CC95}, an operator $\mathcal{P}: \mathcal X \rightarrow \mathcal X$ can be approximated by a parametric operator $\mathcal{P}_{NN}$ with arbitrary accuracy. That is, for any $\varepsilon>0$, there exists a sufficiently large parametric neural network $\mP_{NN}$, such that
\[
\int_{\mX} \int_{\Omega_x} | \mP(f)(\bx) - \mN(f)(\bx) |^2 d \bx d \mu(f) < \varepsilon\,.
\]
Here $\mu$ denotes a probability measure on $\mX$. 
In practice,  $\mu$ is chosen as a Gaussian measure induced by the law of a Gaussian random field. Several operator networks have been proposed recently, including DeepONets \cite{lu2019deeponet,wang2021learning} and various neural operators \cite{bhattacharya2020model,li2020fourier,kovachki2021neural}. In this paper, we adopt DeepONet as the basic architecture and refine it with transfer learning. The vanilla DeepONet  takes the following form:
\begin{align}\label{eqn:approx2}
	\mathcal{P}_{NN} (f)(&\bx;\theta, \xi) \nonumber \\
	&= \sum_{k=1}^p b^{NN}_k(f(\by_1), \cdots, f(\by_N); \theta) t^{NN}_k(\bx; \xi) \nonumber
	\\ & =:\sum_{k=1}^p b^{NN}_k(f; \theta) t^{NN}_k(\bx; \xi).
\end{align}
The operator network $\mP_{NN}$ consists of two sub-networks: the branch net $b^{NN}$ is paramterized by $\theta$ and maps an encoded input function $\{f(\by_i)\}_{i=1}^N$ to $p$ scalars $b_k^{NN}$, and the trunk net $t^{NN} = \{t_k^{NN}\}_{k=1}^p$  is parameterized by $\xi$ and forms a directionary of functions in the output space. Both networks can be modified in practice to fit with various set-ups of PDEs, such as the boundary condition. A diagram of the DeepONets architecture we use in this paper is shown in Figure~\ref{fig:tl_pidon_arch}. The vanilla DeepONets is often trained in a supervised fashion and  requires pairs of input-output functions. To be more specific, given $N_s$ randomly sampled functions $\{ f_s(\bx) \}_{s=1}^{N_s}$ one needs to prepare reference solutions $\{ \mP (f_s)(\bx) \}_{s=1}^{N_s}$ either analytically or using conventional high-fidelity numerical solvers. Then one trains the $\mP_{NN}$ by minimizing the  loss function
\[
\begin{aligned}
	& \frac{1}{2N_s} \sum_{s=1}^{N_s}\Big( \| \mathcal{P}_{NN}(f_s)(\cdot; \theta, \xi) - \mP (f_s)(\cdot) \|_{L^2(\Omega_x)}^2 \\
	\qquad & + \|\mathcal{P}_{NN}(f_s)(\cdot; \theta, \xi)  - \phi_s(\cdot)\|_{L^2(\partial \Omega_x)}^2\Big). 
\end{aligned}
\]
However, in reality it can be extremely expensive to obtain the outputs $\mathcal{P}(f_s)$, especially when the underlying physical principles are complicated and the dimension of the problem is high. To this end, \cite{wang2021learning} proposed a physics-informed DeepONet which makes the  learning procedure above  self-supervised. More precisely, we turn to minimizing the new loss function 
\begin{equation}
	\begin{aligned}
		& \frac{1}{2N_s} \sum_{s=1}^{N_s}\Big(  \| \mathcal{P}^{-1}\big(\mathcal{P}_{NN}(f_s)(\cdot; \theta, \xi)\big) -  f_s(\cdot) \|_{L^2(\Omega_x)}^2 \\
		\qquad & + \|\mathcal{P}_{NN}(f_s)(\cdot; \theta,\xi)  - \phi_s(\cdot)\|_{L^2(\partial \Omega_x)}^2\Big).  \label{eqn:emp_loss}
	\end{aligned}
\end{equation}
Note that the introduction of $\mP^{-1}$ in \eqref{eqn:emp_loss} completely avoids the evaluations of $\mP(f_s)$. The boundary term in \eqref{eqn:emp_loss} can be further eliminated in practice because the networks can be modified to satify the  boundary conditions (see e.g. \cite{lu2022comprehensive}). We also observe through numerical experiments that eliminating the  the boundary loss can substantially improves the training efficiency. 
The physics-informed DeepONet has  been applied to learning evolution equations \cite{wang2021long}. For equation \eqref{111}, instead of first discretizing it in time, they consider time as an additional input variable, and try to learn an operator $\mP^I$ that maps the initial condition to the solutions over an time-interval $[0,t_0]$: 
\begin{align*}
	\mP^I: f(0,\bx) \mapsto f(t,\bx), \quad \text{for} \quad t \in [0,t_0]\,.
\end{align*}
The corresponding loss function to be minimized is
\begin{align}
	L(\theta,  w,  \xi) & =  \frac{1}{2N_s} \sum_{s=1}^{N_s}   ( \| \partial_t (\mP_{NN}^I (f_s))(t,\bx)  \nonumber \\ & - \mathcal L (\mP_{NN}^I (f_s))(t,\bx) \|_{L^2(\Omega_x \times [0,t_0])}^2 \nonumber \\
	& + \| \mP_{NN}^I f_s(\bx) - g_s(\bx) \|_{L^2(\partial\Omega_x \times [0,t_0])}^2 ),  \label{eqn:bc_loss}
\end{align}
where $\mP_{NN}^I$ is the neural network approximator to $\mP^I$. Once trained, $\mP_{NN}^I$ can be applied to $f(t_0,\bx)$ to get the solution $f(t,\bx)$ over $[t_0, 2t_0]$. Repeating this process enables one to obtain approximation solutions in any finite time. However, this methodology may suffer from long-time instability. In fact, let us illustrate this using the Allen-Cahn equation \eqref{eqn:ac} in one dimension.  Figure~\ref{fig:failure} shows that the average $L^2$ errors
of approximated solutions learned by  DeepONets using both the single-shot loss \eqref{eqn:emp_loss} (labeled as DeepONet) and the time-integrated loss \eqref{eqn:bc_loss} (labeled as CONT DeepONet). Both errors accumulate rapidly as time increases, indicating the instability of the learned DeepONets in prediction of long-time solution. In contrast, our transfer-learning assisted DeepOnet dramatically reduces the error and stabilizes the prediction. For completeness, we also compare them with the Fourier Neural Operator(FNO) \cite{li2020fourier}, which is another  state-of-the-art operator learning method. 

\subsection{DeepONet with transfer learning}

\begin{figure*}[h!]
\centering
{\includegraphics[width=0.78\textwidth, height=0.22\textheight]{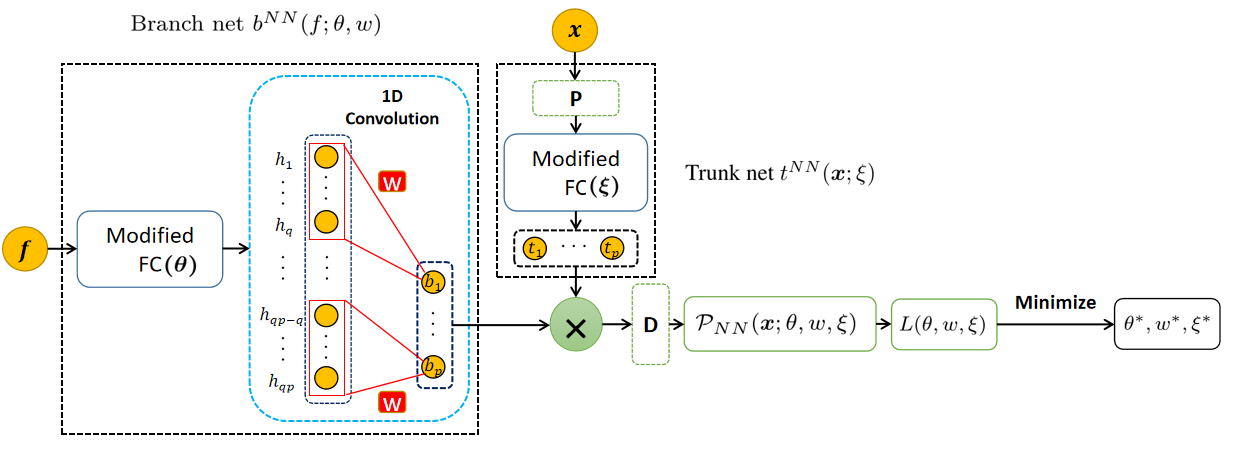}}
\caption{The architecture of transfer learning aided physics-informed DeepONet. Here $P$ and $D$ are optional layers that  enforce periodic and Dirichlet boundary conditions, respectively. The block named Modified FC is a modified fully connected neural networks architecture introduced in \cite{wang2021learning}. The parameter $w$ (in the red box) denotes the tunable weights in the last hidden layer of the branch net. In the transfer learning step, only $w$ will be re-trained while the $\theta, \xi$ are frozen. } 
\label{fig:tl_pidon_arch}
\end{figure*}

The main idea of transfer learning is to train a neural network on a large data set and then partially freeze and apply it to a related but unseen task. Inspired by \cite{desai2021one}, we employ the transfer learning technique to successively correct the trained DeepONet at the prediction steps:  we freeze the majority of the well-trained DeepONet and merely re-train the weights in the last hidden layer of the branch net by fitting the same physics-informed loss \eqref{eqn:emp_loss} defined by the underlying PDEs. 
To be more precise, by separating the parameters $\theta$ in the hidden layers and the parameter $w$ in the last layer of the branch net, we rewrite the  branch net as
\begin{align*}
b_k^{NN}(f;\theta, w) = \sum_{j=1}^q w_j h_{k,j}(f;\theta)\,,
\end{align*}
where $h = \{h_{k,j}\}$ are the outputs of the last hidden layer of the branch net and $w = \{w_j\}$ are the  weights in the last layer. Inserting this into \eqref{eqn:approx2} gives 
\begin{equation}\label{eqn:approx_trans}
\mathcal{P}_{NN} (f)(\bx;\theta, w, \xi) = \sum_{k=1}^p \sum_{j=1}^q w_j h^{NN}_{k, j}(f; \theta) t^{NN}_k(\bx; \xi)\,.
\end{equation}
The architecture of the new operator network is illustrated in Figure~\ref{fig:tl_pidon_arch}.
In the training step, the optimal  parameters $(\theta^*, w^*,\xi^*)$ of the DeepONet \eqref{eqn:approx_trans} can be obtained by minimizing the empirical loss \eqref{eqn:emp_loss}. Later in each prediction step, we freeze the value of $\theta^*$ and $\xi^*$, but update $w^\ast$ by re-training the loss \eqref{eqn:emp_loss} with newly-predicted solution as the initial condition. Namely with the predicted solution $f_n$ at step $n$, we seek $w^\ast_{n+1}$ defined by 
\begin{align}\label{eqn:opt_w}
&  w^*_{n+1}  \in \arg \min_w  \nonumber \\ & \frac{1}{2N_s} \sum_{s=1}^{N_s}\Big(  \| \mathcal{P}^{-1}  \mathcal{P}_{NN} (f)(\bx;\theta^\ast, w, \xi^\ast)  - f_n(\bx) \|_{L^2(\Omega_x)}^2 \nonumber \\
+ & \|  \mathcal{P}_{NN} (f)(\bx;\theta^\ast, w, \xi^\ast)  - \phi(\bx)\|_{L^2(\partial \Omega_x)}^2\Big), n=1,2,\cdots.
\end{align}
Note that $w^\ast_1 = w^\ast$. The optimal sequence of weights $w_n^\ast$ defines a sequence of operator networks $\mP^n_{NN} := \mP^{n}_{NN}(\theta^\ast, w^\ast_n, \xi^\ast)$, which can be used to approximate the solution at $t=n\Delta t$ by 
$$
f(n\Delta t) \approx \mP^{n}_{NN}\circ \mP^{n-1}_{NN}\circ \mP^{1}_{NN} (f_0).
$$
It is interesting to note that the proposed method shares some similarities with the classical Galerkin approximation. In fact, the operator network \eqref{eqn:approx_trans} can be further rewritten as 
\begin{align*}
\mathcal{P}_{NN} (f)(\bx) & = \sum_{j=1}^q w_j \left( \sum_{k=1}^p  h^{NN}_{k, j}(f; \theta) t^{NN}_k(\bx; \xi) \right) \\
&=: \sum_{j=1}^q w_j \phi_j(\bx; f)\,,
\end{align*}
Observe that $\phi_j$ playing the role of basis functions in Galerkin methods, and $w_j$ being the corresponding weight. 
However, unlike most Galerkin methods which often use  handcraft bases, such as piecewise polynomials and trigonometric functions, here the bases are learned from the problem itself, and vary with the function they approximate. This seemly minor change reduces substantially the number of bases needed in the output space, as shown by extensive numerical tests in Section \ref{sec:numerical_exp}. To minimize \eqref{eqn:opt_w}, it amounts to solving a system of $N$ equations with $q$ unknowns, where $N$ is the total number of fixed sensors in the branch net. This is achieved by least square minimization. Since $q \ll N$, the computational complexity of finding the least square solution is only $\mathcal{O}(q^2 N)$.  In practice, we further reduce the computational complexity by sub-sampling $N_c$ grid points out of $N$ in the transfer learning step.

\section{Theoretical result}\label{sec:thm}
In this section, we analyze the long time stability of the learned operator $\mP_{NN}$. First let $\mX $ be a Banach space and assume that the original propagator $\mP$ (i.e., $\mP^{\Delta t}$ in \eqref{eqn:mP}): $\mathcal{X} \rightarrow \mathcal{X}$ is non-expansive such that 
\begin{align} \label{assump}
\norm{\mP}_{\mathcal{X}}:= \sup_{f\in \mathcal{X}, \norm{f}_{\mX} =1} \norm{\mP f}_\mX  \leq 1.
\end{align}
In the case that $\mX = L^2(\Omega_x)$, assumption\eqref{assump} follows from the dissipative assumption of $\mL$; see Appendix \ref{sec:assump} for more details. Let $\mU \subseteq \mX$ be a linear subspace, we also assume that  
\begin{align}\label{assmp2}
\mP f \in \mU, \quad \forall f \in \mU\,.
\end{align}
The theorem below shows that the long-time prediction error of the operator network can be bounded by the loss function. 
\begin{theorem}\label{thm:main}
Assume \eqref{assump} and \eqref{assmp2} hold. If the neural network approximator $\mP^{NN}$ satisfies that the maximum  loss over the set $\mathcal U$ is less than $\delta$, i.e. 
\begin{equation} \label{loss0}
\sup_{f\in \mathcal U, \|f\|_{\mX}=1 } \norm{\mP^{-1} \mP_{NN} f - f}_\mX \leq \delta\,,
\end{equation}
and that 
\begin{equation} \label{eq:Pnnf}
\mP_{NN} f \in \mU, \quad \forall f \in \mU,
\end{equation} 
then the following long-time stability holds 
\begin{equation} \label{sta1}
\sup_{f\in\mU, \|f\|_{\mX}=1}	\norm{(\mP)^K f - (\mP_{NN})^K f}_\mX \leq \delta K(1+\delta )^K.
\end{equation}
Moreover, if we further assume that 
\begin{equation} \label{assump2}
\norm{\mP}_{\mX} \leq \eta < 1
\end{equation}
and  \eqref{loss0} holds with $\delta \leq \half (1-\eta)$, then we have 
\begin{equation} \label{est2}
\sup_{f\in\mU, \|f\|_{\mX}=1} 	\norm{(\mP)^K f - (\mP_{NN})^K f}_\mX \leq \delta K \Big(\frac{ 1+\eta}{2}\Big)^{K-1}\,.
\end{equation}
\end{theorem}

The proof of Theorem \ref{thm:main} is provided in Appendix~\ref{pf}.




\begin{remark}

\begin{itemize}
\item[1.] If the error tolerance $\delta = \Delta t^2$ with $\Delta t$ being the time-discretization stepsize and the number of iterations $K = \frac{T}{\Delta t}$, then  \eqref{sta1} becomes 
$$
\sup_{f\in\mU, \|f\|_{\mX}=1}	\norm{(\mP)^K f - (\mP_{NN})^K f}_\mX \leq e^{T\Delta t} T \Delta t \,.
$$
When  assumption \eqref{assump2} holds, the estimate above improves to 
$$\begin{aligned}
	\sup_{f\in\mU, \|f\|_{\mX}=1} 	\norm{(\mP)^K f - (\mP_{NN})^K f}_\mX & \leq \big(\frac{1+\eta}{2}\big)^{\frac{T}{\Delta t}} T \Delta t\\
	& \leq  C\Delta t,
\end{aligned}
$$
where $C$ is a constant independent of $T$ and $\Delta t$, suggesting that the prediction error is of the order $\mathcal{O}(\Delta t)$ uniformly in time.

\item[2.] We comment on the assumptions made in Theorem \ref{thm:main}. In practice, physics-informed loss \eqref{eqn:opt_w} is trained so that condition \eqref{loss0}  is fulfilled for some subspace space $\mU$, e.g. $\mU = \{e^{i \mathbf{k}\cdot \mathbf{x}}\}_{|\mathbf{k}| \leq K_0}$. Assumption \eqref{assmp2}  holds for such a choice of $\mU$ when the evolution equation involves diffusion. Assumption \eqref{eq:Pnnf} holds in particular when sine or cosine activation function is used in the operator network $\mathcal{P}_{NN}$. 
\end{itemize}


\end{remark}

\section{Numerical experiments}\label{sec:numerical_exp}
In this section, we demonstrate the effectiveness of transfer learning enhanced DeepONet and show its advantages over the vanilla DeepONet through several evolutionay PDEs, including reaction diffusion equation, Allen-Cahn and Cahn-Hilliard equations, Navier-Stokes equation and multiscale linear radiative transfer equation.  
The equations of consideration are equipped with either Dirichlet or periodic boundary conditions. In all the test problems, our goal is to predict the long time evolution of the equations obtained by successive actions of the propagators learned via DeepONets. More concretely, we first build the first-step neural operator approximation $\mP^1_{NN}$ to the propagator $\mP = \mP^{\Delta t}$ by minimizing the physics-informed loss \eqref{eqn:emp_loss} with $M$ training initial data. The operator network $\mP^1_{NN}$ is then gradually tuned to $\mP^j_{NN},j=2,\cdots K$ via updating the weights $w$ in the last-layer of its trunk nets. With the learned (and adjusted) operators $\mP^j_{NN},j=1,\cdots K$, the solution of a PDE at  time $t=K\Delta t$ with an initial condition $f_0$ can then be obtained approximately by $\mP^K_{NN}\circ \cdots \circ \mP^1_{NN} f_0$.  We remark that  the $M$ training data is constructed as a subset of a larger  training set of size $N_s\times N_p$, which consists of pointwise evaluations of $N_s$ randomly sampled functions at $N_p$ physical locations (sensors). We refer to Appendix \ref{sec:data} for detailed discussions on the data generating process and treatment of boundary conditions in various test problems. Choices of parameters for the operator networks and the training process are discussed in the end of Appendix \ref{sec:data}.   In all the numerical results to follow, we quantify the performance of the proposed method by measuring the aggregated relative prediction error over a time horizon $[0,T]$; see the precise definition of the relative error in Appendix \ref{sec:data}.  The codes used for the numerical experiments will be published on the website  \url{https://github.com/woodssss/TL-PI-DeepONet}.

\subsection{Reaction diffusion equation}
Consider the reaction diffusion equation
\begin{equation}\label{eqn:nrd}
\left\{ \begin{array}{l}
\partial_t f = d  \Delta f + k f^2, x\in \Omega_x:= [0,1]
\\ f(t,\bx) = 0, ~~\bx \in \partial \Omega_x = \{0,1\}\,,
\\ f(0,\bx) = f_0(\bx) \,,
\end{array}\right.
\end{equation}
where $d=k=0.001$. In this example, we train DeepOnets and our transfer learning enhanced DeepONets using two different loss functions, with one based on the physics-informed loss within a single time-step $\Delta t = 0.05$ (c.f. \eqref{eqn:emp_loss}), and the other based on the aggregated physics-informed loss \eqref{eqn:bc_loss} in the time window $t\in [0,1]$. We refer to the DeepONets trained using the latter loss as {\em CONT DeepONets} and reserve {\em DeepONets} for the one trained by the former loss.  The numerical results of different DeepONets with varying training sample sizes $M$ are shown in Table~\ref{table:nrd}. Our proposed method provides more accurate and more robust prediction of the solutions with little extra computational cost. In particular, the prediction error of vanilla DeepONets and CONT DeepONets increase dramatically as time increases from $0.2$ to $50$, while transfer learning can significantly reduces the error and stabilizes the prediction in the long time. 
In addition, our method also substantially reduces the size of training data to achieve the same order of prediction accuracy. 
Note that, since trained only within a single time step, DeepONets take far less training time than the corresponding CONT DeepONets while maintain comparable accuracy. The similar trade-off of accuracy and training cost applies to other experiments. For this reason, in subsequent examples we will only report results on our proposed method and the vanilla DeepONet, and exclude the results from CONT DeepONets. Note also that the results obtained in Table ~\ref{table:nrd} are for propagators defined by the backward Euler scheme. One can also consider propagators defined by higher order time-discretization schemes and their neural network approximation. We refer the numerical results obtained using the Crank-Nicolson method to Table ~\ref{table:nrd_2nd} in Appendix \ref{sec:reaction}.

\begin{table}[!ht]
\centering
\begin{tabular}{l|lll|ll}
\toprule
\cmidrule(r){1-2}
Neural network    & $M$  & $t_1$ & $t_2$ &  $T=0.2$ & $T=50$ \\
\midrule
CONT  & 1000  & 43030  & 0.26 & 7.95e-2 & 4.17e-1     \\
DeepONet & 3000   & 79375 & 0.28 & 1.01e-2 & 4.15e-2    \\ 
& 10000  & 83465  & 0.26 & 3.40e-3 & 1.34e-2      \\
CONT    & 1000  &43030 & 0.93 & \textbf{5.38e-3} & \textbf{1.95e-3}    \\
TL-DeepONet & 3000   & 79375 & 0.98 & \textbf{1.87e-3} & \textbf{6.88e-4}   \\ 
& 10000  &83465 & 0.93 & \textbf{1.84e-3} & \textbf{4.87e-4}     \\
DeepONet  & 1000  & 2575  & 5.02 & 2.75e-1 & 1.69e0     \\
& 3000   & 4313 & 4.9 & 1.05e-1 & 1.60e0    \\ 
& 10000  & 5854 & 4.8 & 9.19e-2 & 1.87e0      \\
TL-DeepONet & 1000  & 2575 &  9.86 & \textbf{1.03e-3} & \textbf{1.52e-3}    \\
& 3000   & 4313 & 7.1 & \textbf{8.34e-4}  & \textbf{1.19e-3}    \\ 
& 10000  & 5854 & 8.2  &  \textbf{8.19e-4} &  \textbf{9.05e-4}     \\
\bottomrule
\end{tabular}
\caption{Results on reaction diffusion equation. Here $t_1$ is the training time and $t_2$ is the averaged  time of predicting the solution trajectories among the time interval $[0, 50]$ based on $30$ test initial conditions. The last two columns to the right are the averaged relative $L^2$ error within $[0,T]$.}
\label{table:nrd}
\end{table}

\subsection{Allen-Cahn and Cahn-Hilliard equations}
\begin{figure}
\centering
\setlength{\tabcolsep}{-2pt}
\renewcommand{\arraystretch}{-1}
\begin{tabular}{ccc}
\text{Initical condition} \quad & t=4 & t=10\\
\raisebox{-.5\height}{\includegraphics[width=0.16\textwidth, height=0.1\textheight]{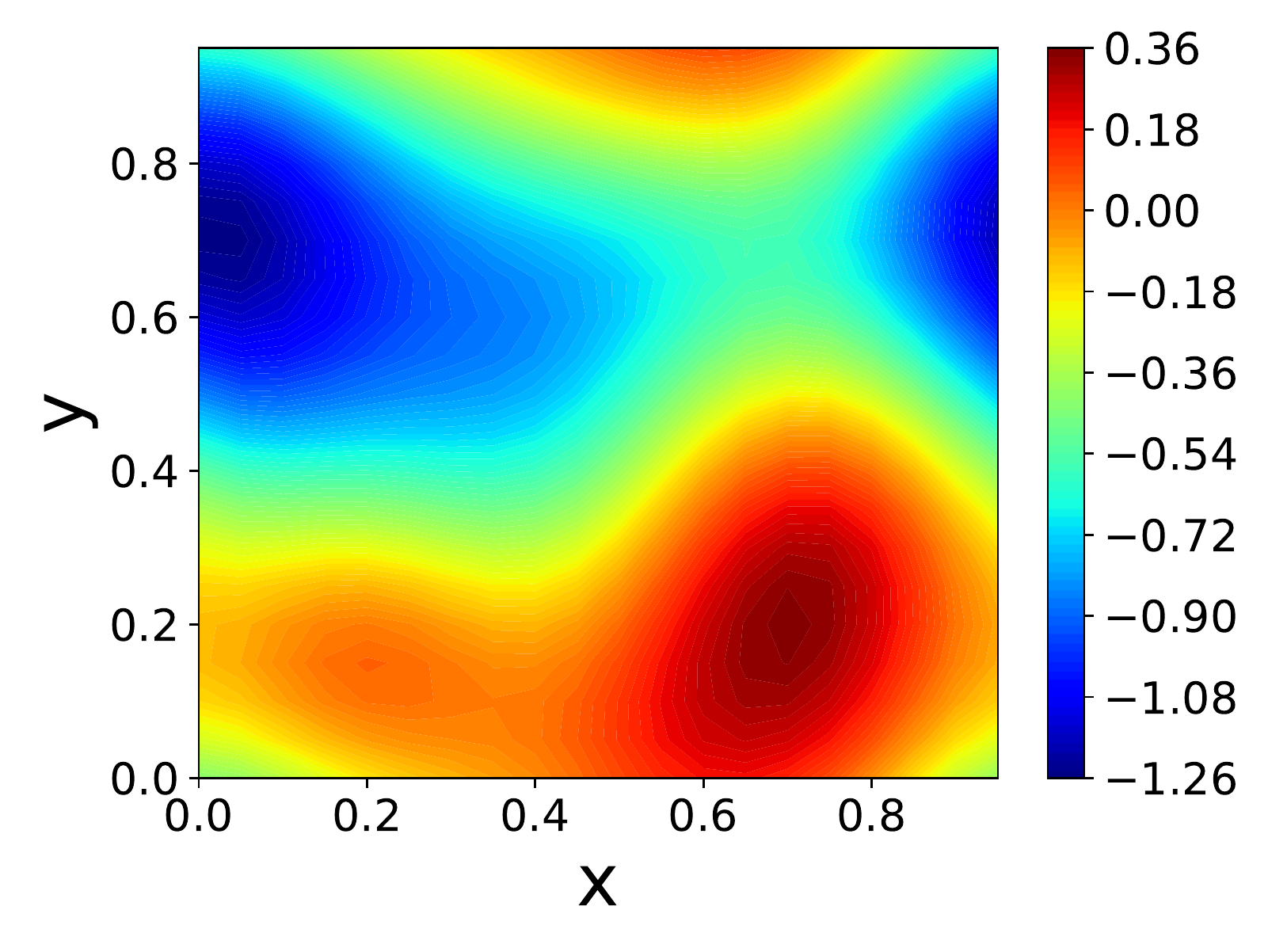}}&
\raisebox{-.5\height}{\includegraphics[width=0.16\textwidth, height=0.1\textheight]{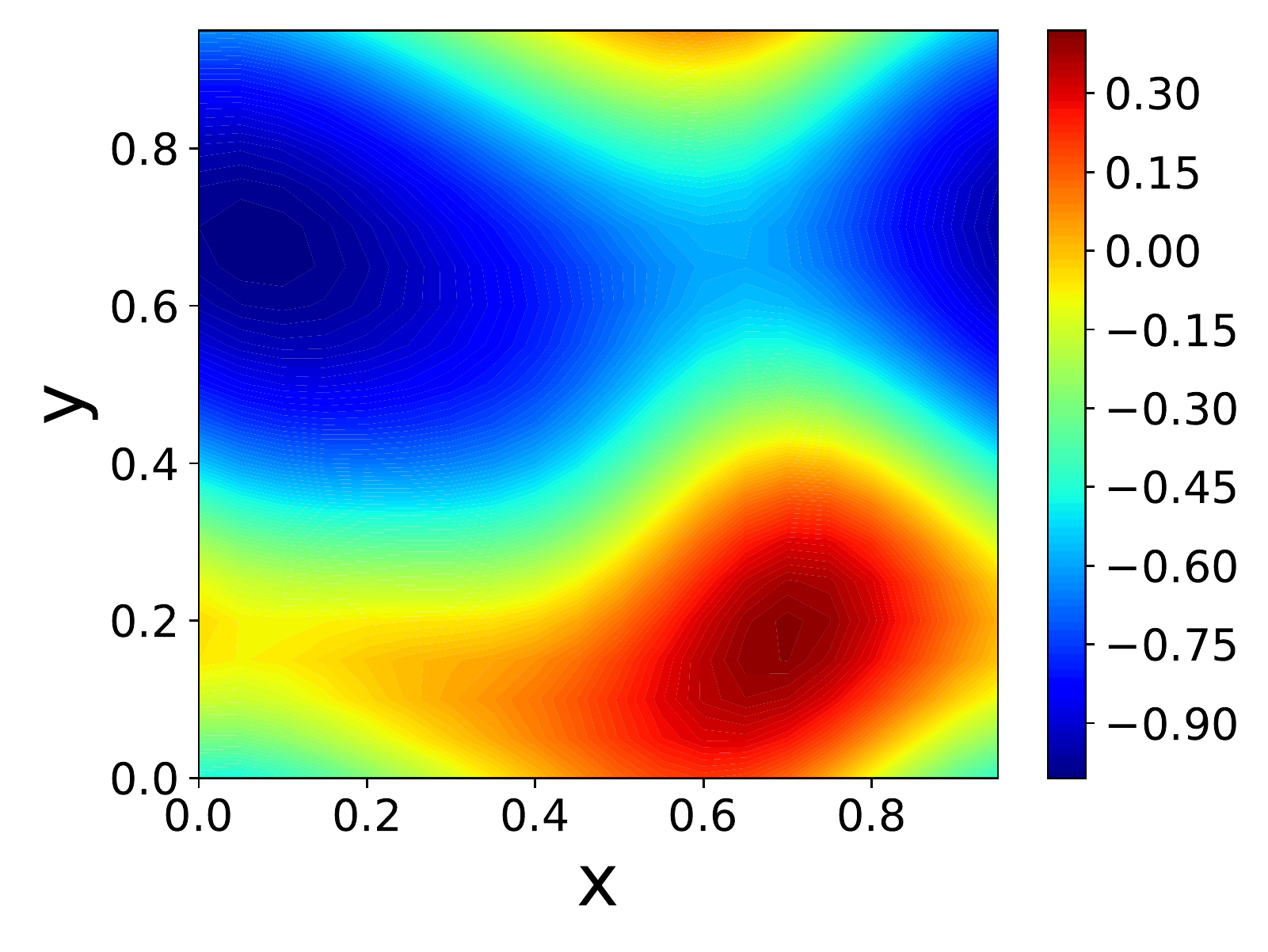}}&
\raisebox{-.5\height}{\includegraphics[width=0.16\textwidth, height=0.1\textheight]{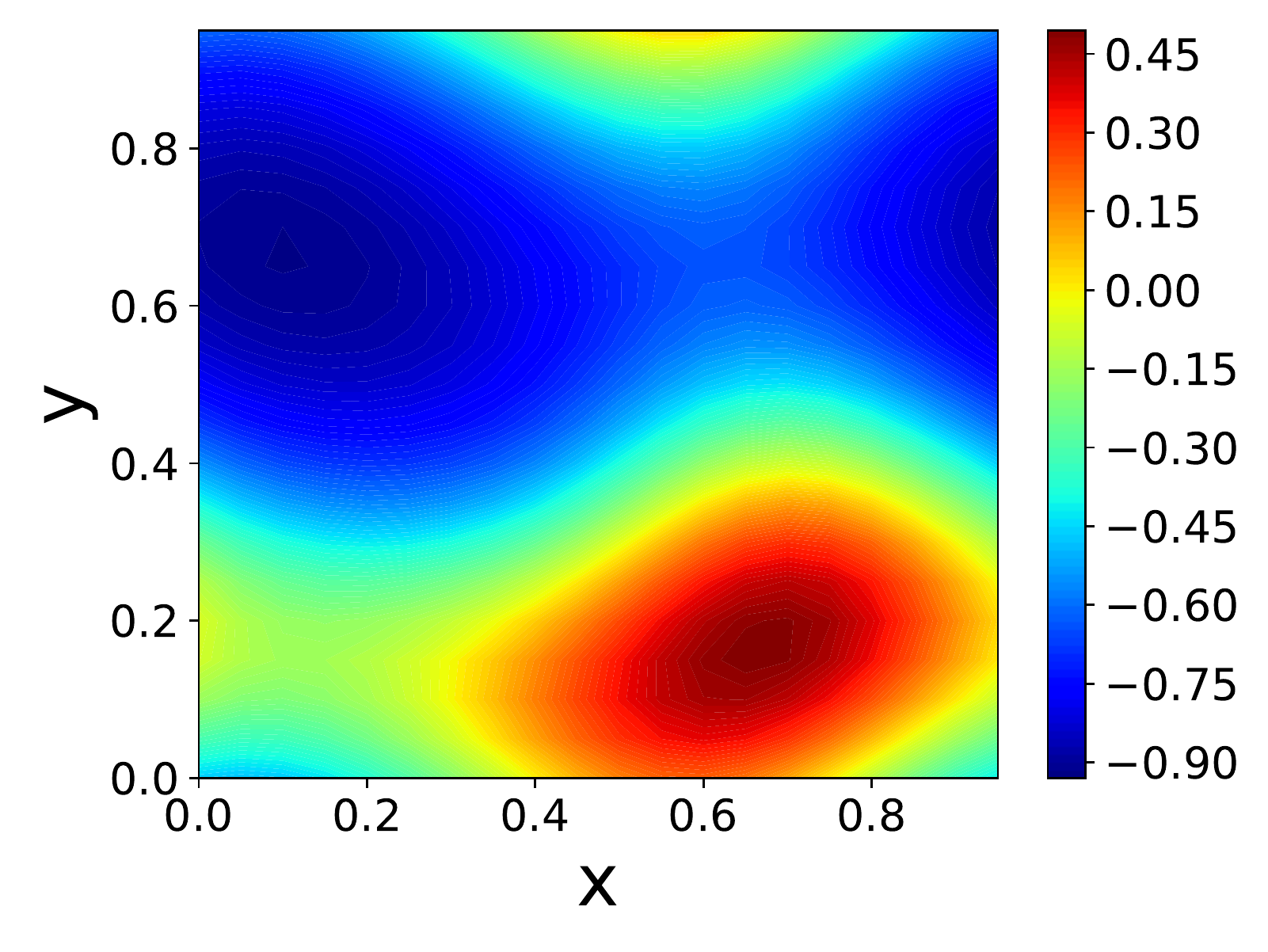}}\\
\text{DeepONet}&
\raisebox{-.5\height}{\includegraphics[width=0.16\textwidth, height=0.1\textheight]{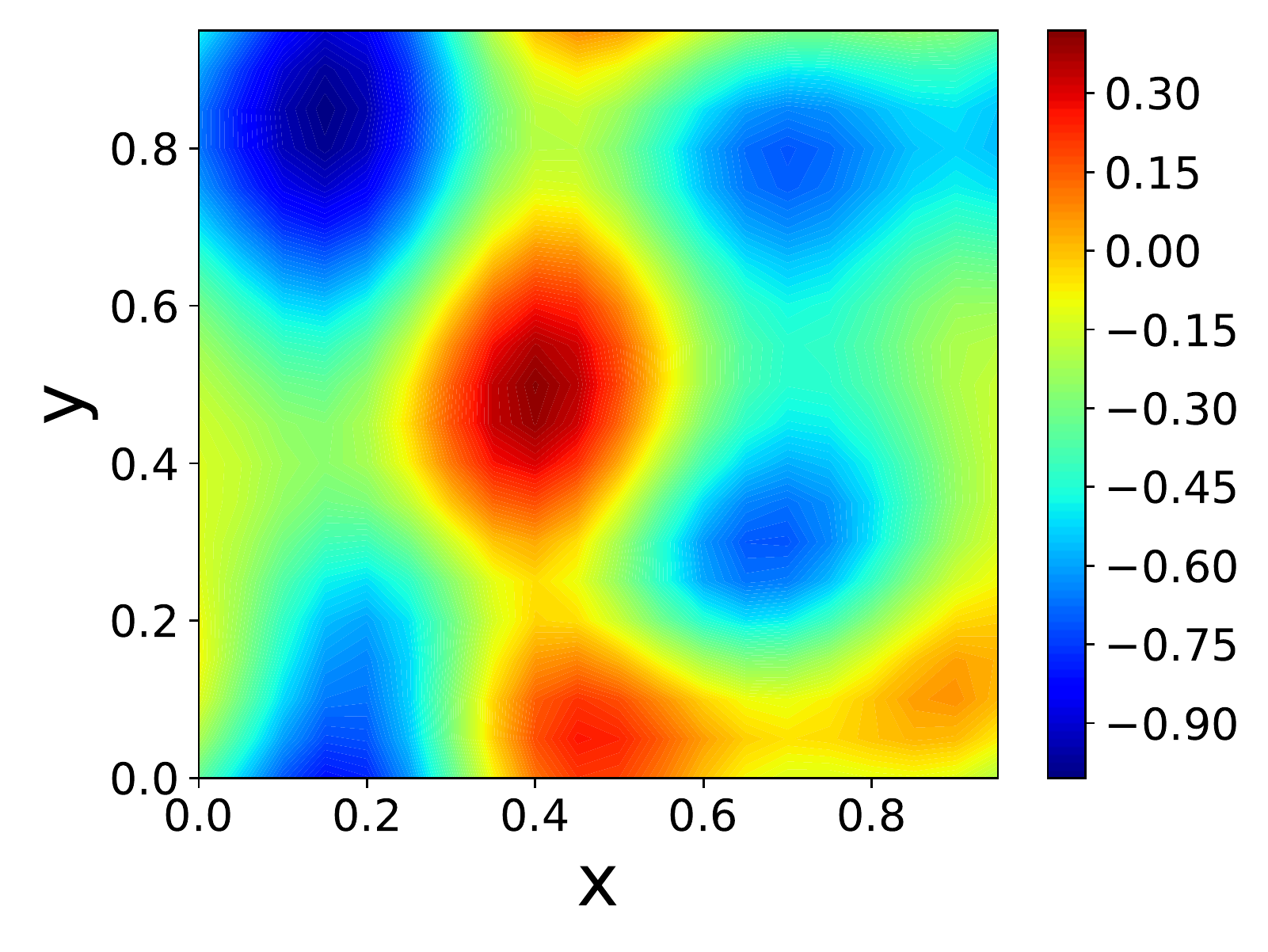}}&
\raisebox{-.5\height}{\includegraphics[width=0.16\textwidth, height=0.1\textheight]{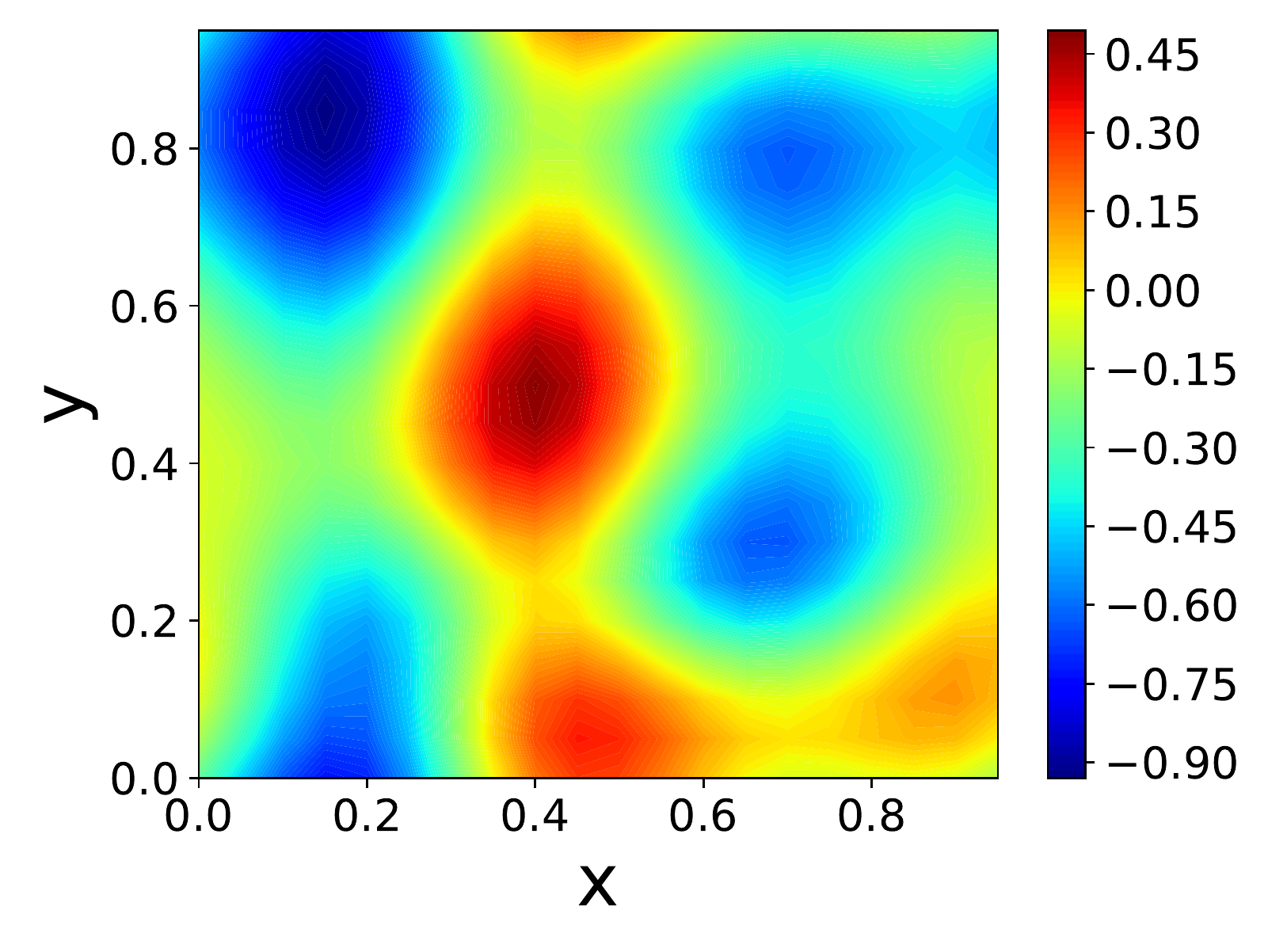}}\\
\text{TL-DeepONet}&
\raisebox{-.5\height}{\includegraphics[width=0.16\textwidth, height=0.1\textheight]{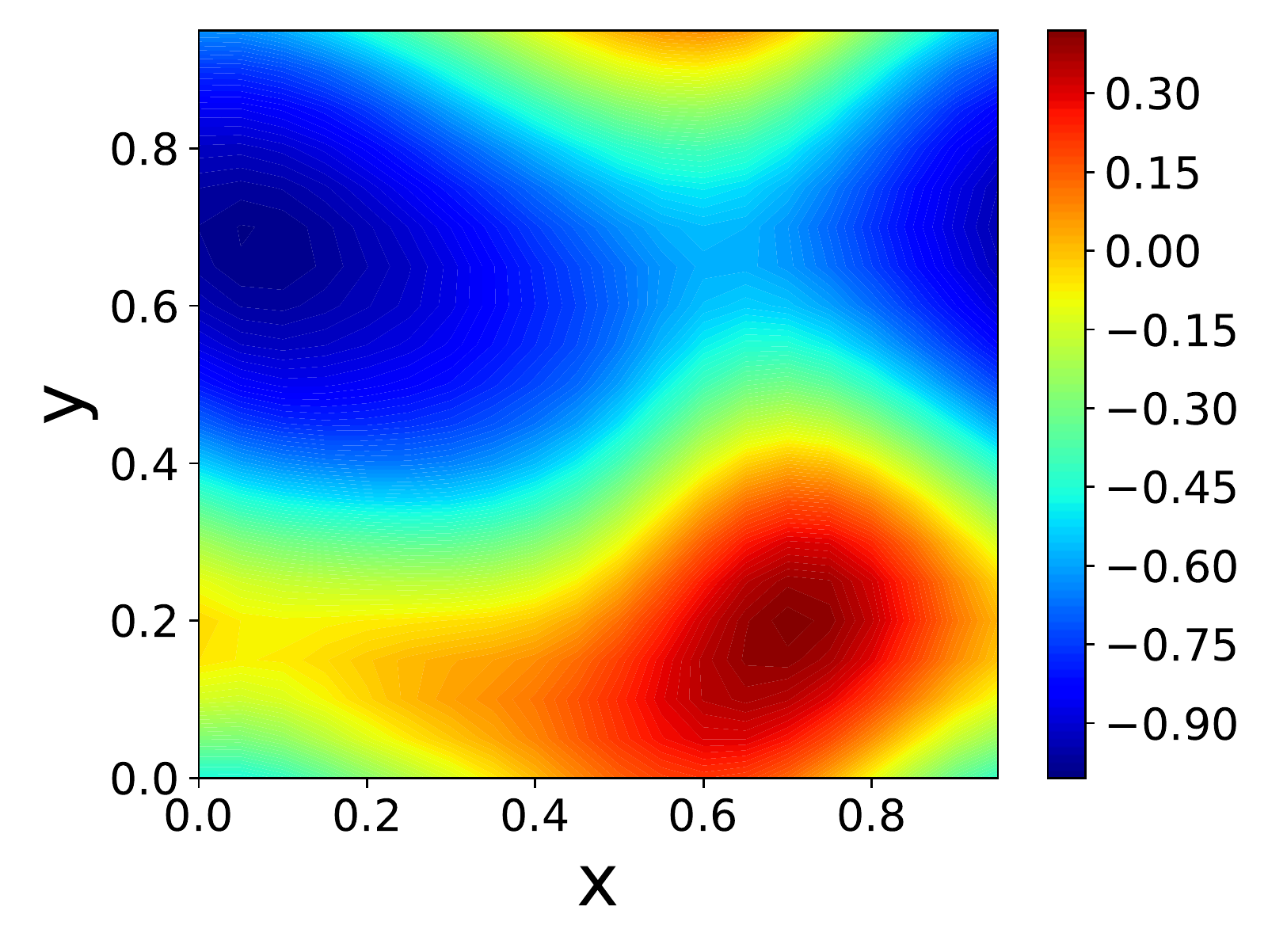}}&
\raisebox{-.5\height}{\includegraphics[width=0.16\textwidth, height=0.1\textheight]{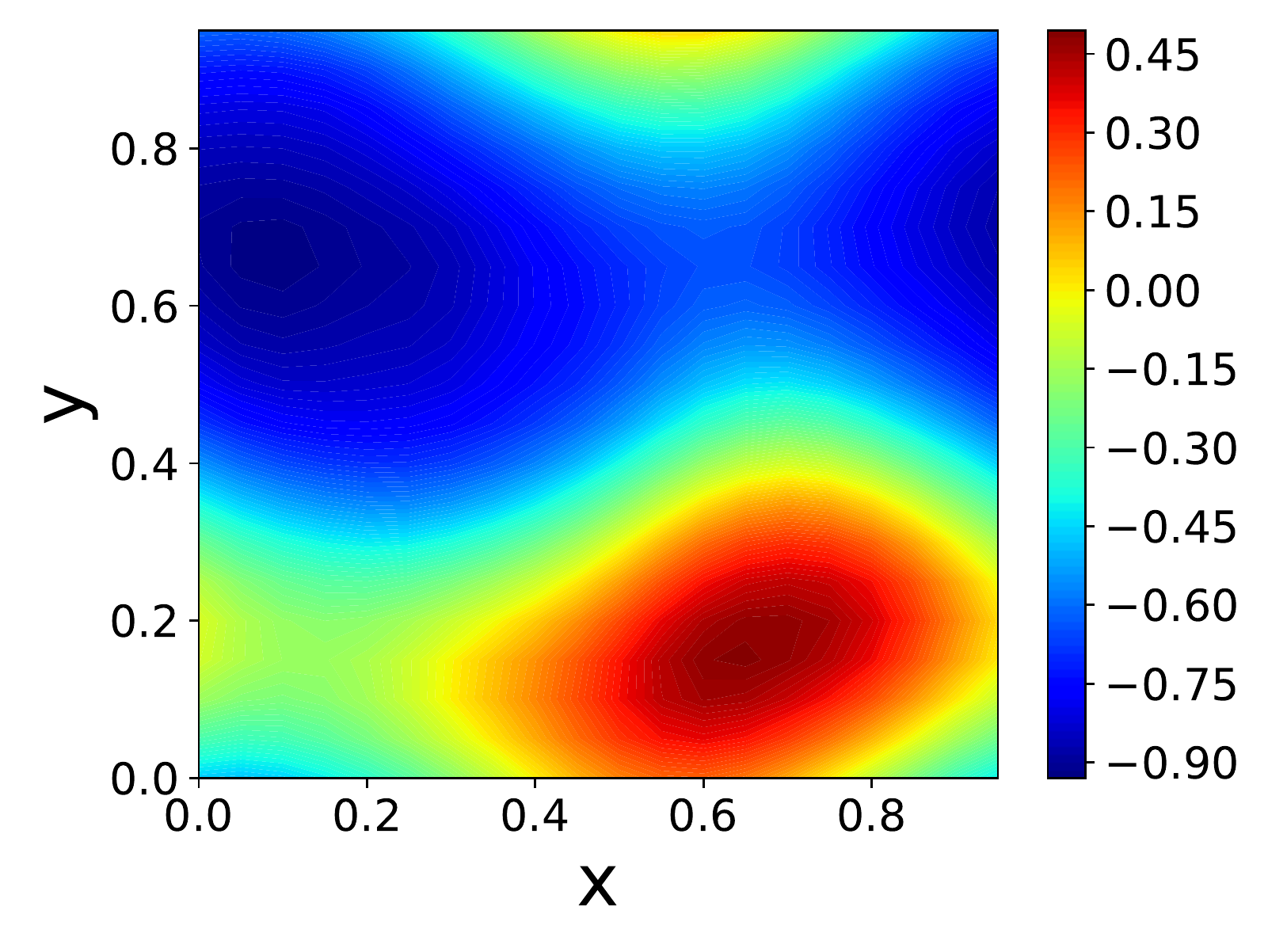}}\\
\end{tabular}
\caption{Results on 2D Cahn-Hilliard equation: snapshots of reference solutions (top), and of approximate solutions predicted by   DeepONet (middle) and TL-DeepONet (bottom).}
\label{fig:ch2d} 
\end{figure}

In the second example, we consider Allen-Cahn equation 
\begin{equation}\label{eqn:ac}
\left\{ \begin{array}{l}
\partial_t f =  d_1 \Delta f +   d_2 f(1-f^2), \\
f(0, \bx) = f_0(\bx) ,
\end{array}\right.
\end{equation}
and Cahn-Hilliard equation 
\begin{equation}\label{eqn:ch}
\left\{ \begin{array}{l}
\partial_t f =   \Delta g, \\
g = - d_1 \Delta f +   d_2 (f^3 - f), \\
f(0, \bx) = f_0(\bx),
\end{array}\right.
\end{equation}
both equipped with periodic boundary conditions. They are prototype models for the motion of anti-phase boundaries in crystalline solids. The computational domain is $ \Omega := [0, 1]^d$ with $d=1,2$. We are interested in learning the propagator $\mP = \mP^{\Delta t}$ with $\Delta t = 0.05$ and used it to predict the solutions $f(t, \bx)$ for every $t\leq T=50$. The results on Allen-Cahn equation are shown in Table~\ref{table:ac} (1D) and Table~\ref{table:ac_2d} (2D). See also Figure~\ref{fig:failure} for a plot of evolving relative errors on 1D Allen-Cahn equation. Similar results for 1D Cahn-Hilliard equation are presented in 
Table~\ref{table:ch} and Figure~\ref{fig:ch2d} compares the snapshots of predicted solutions to the 2D Cahn-Hilliard equation. In all the results, for a fixed $d_2$, the relative errors increase as $d_1$ decreases because the transition layers of solutions are increasingly sharper and hence make the numerical resolution more challenging. Similar to the previous example, our proposed method provides more accurate prediction of solutions than the vanilla DeepONets among all the configurations of parameters. We note that the average trajectory prediction times of TL-DeepONets increase for about 3 times compared to those of the vanilla DeepONets while the prediction errors of the former decrease by at least two orders of magnitude. 
\begin{table}[h]
\centering
\begin{tabular}{l|llll}
\toprule
\cmidrule(r){1-2}
Neural network    & $M$    &  $d_1$=1e-3 &  $d_1$=5e-4 &  $d_1$=1e-4  \\
\midrule
DeepONet  & 1000   &  1.13e0 & 1.33e0 &  1.29e0 \\
& 3000   & 1.33e0 & 1.18e0   &  1.23e0 \\ 
& 10000  & 1.01e0 &  8.95e-1  & 1.43e0     \\
TL-DeepONet  & 1000  & \textbf{9.25e-4} & \textbf{9.64e-4}  & \textbf{2.16e-2}   \\
& 3000 & \textbf{7.78e-4}  & \textbf{8.83e-4} &   \textbf{1.81e-2}   \\ 
& 10000 & \textbf{5.81e-4} & \textbf{7.94e-4} & \textbf{1.16e-2}       \\
\bottomrule
\end{tabular}
\caption{Results on 1D Allen-Cahn equation: the time-average of relative prediction errors within $[0,50]$. The average trajectory prediction time is 5.1s for DeepONet and 16.1s for TL-DeepONet.}
\label{table:ac}
\end{table}

\begin{table}[h]
\centering
\begin{tabular}{l|llll}
\toprule
\cmidrule(r){1-2}
Neural network    & $M$    &  $d_1$=4e-3 &  $d_1$=2e-3 &  $d_1$=1e-3  \\
\midrule
DeepONet   & 1000   &  9.96e-1 & 1.01e0 &  1.02e0 \\
& 10000   & 9.96e-1 & 1.00e0   &  1.00e0 \\ 
TL-DeepONet   & 1000  & \textbf{6.54e-3} & \textbf{8.43e-3}  & \textbf{1.01e-2}   \\
& 10000 & \textbf{4.96e-3}  & \textbf{6.46e-3} &   \textbf{9.01e-3}   \\ 
\bottomrule
\end{tabular}
\caption{Results on 2D Allen-Cahn equation: the time-average of the relative prediction errors within $[0,10]$. The average trajectory prediction time is 6.1s for DeepONet and 29.5s for TL-DeepONet.}
\label{table:ac_2d}
\end{table}

\begin{table}[h]
\centering
\begin{tabular}{l|llll}
\toprule
\cmidrule(r){1-2}
Neural network    & $M$    &  $d_1$=4e-6 &  $d_1$=2e-6 &  $d_1$=1e-6  \\
\midrule
DeepONet  & 1000 & 9.43e-1  & 9.47e-1 & 9.44e-1 \\
& 3000   & 9.54e-1  & 9.58e-1  & 9.39e-1 \\ 
& 10000  & 9.65e-1 & 9.58e-1  & 9.38e-1    \\
TL-DeepONet  & 1000  & \textbf{1.04e-2}  & \textbf{1.36e-2}  & \textbf{4.25e-2}  \\
& 3000 & \textbf{8.11e-3} & \textbf{9.04e-3} & \textbf{3.86e-2}  \\ 
& 10000 &\textbf{2.29e-3}  &\textbf{7.89e-3}  &   \textbf{3.03e-2}   \\
\bottomrule
\end{tabular}
\caption{Results on 1D Cahn-Hilliard equation: the time-average of the relative prediction errors within $[0,50]$. the average trajectory prediction time is 4.4s for DeepONet and 12.3s for TL-DeepONet.}
\label{table:ch}
\end{table}

\subsection{Navier-Stokes equation}
\begin{figure}[ht]
\centering
\begin{center}
\setlength{\tabcolsep}{-1pt}
\renewcommand{\arraystretch}{-1}
\begin{tabular}{ccc}
\text{Initical condition} \quad & $t=4$ &  $t=10$\\
\includegraphics[width=0.16\textwidth, height=0.1\textheight]{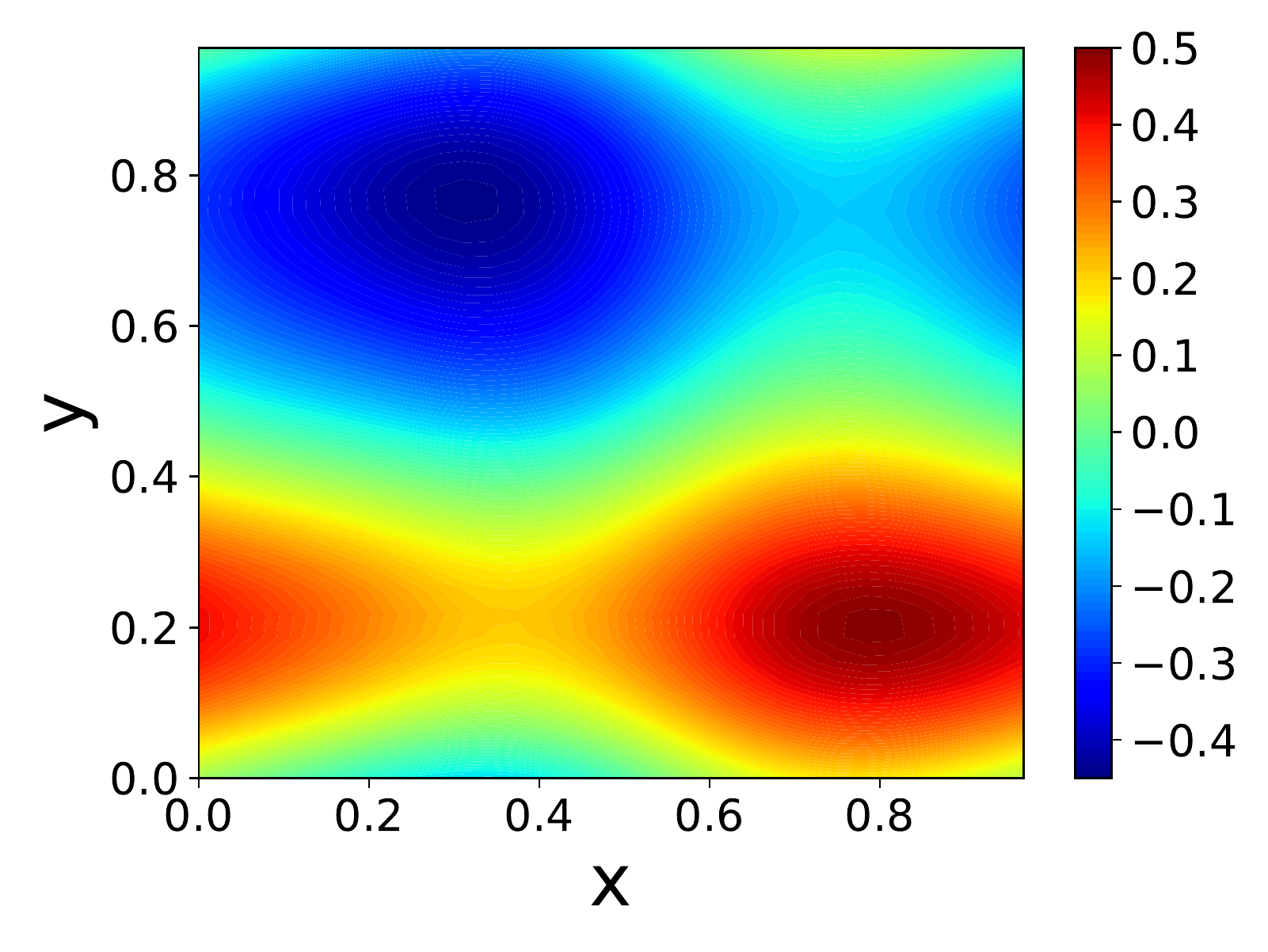}&
\includegraphics[width=0.16\textwidth, height=0.1\textheight]{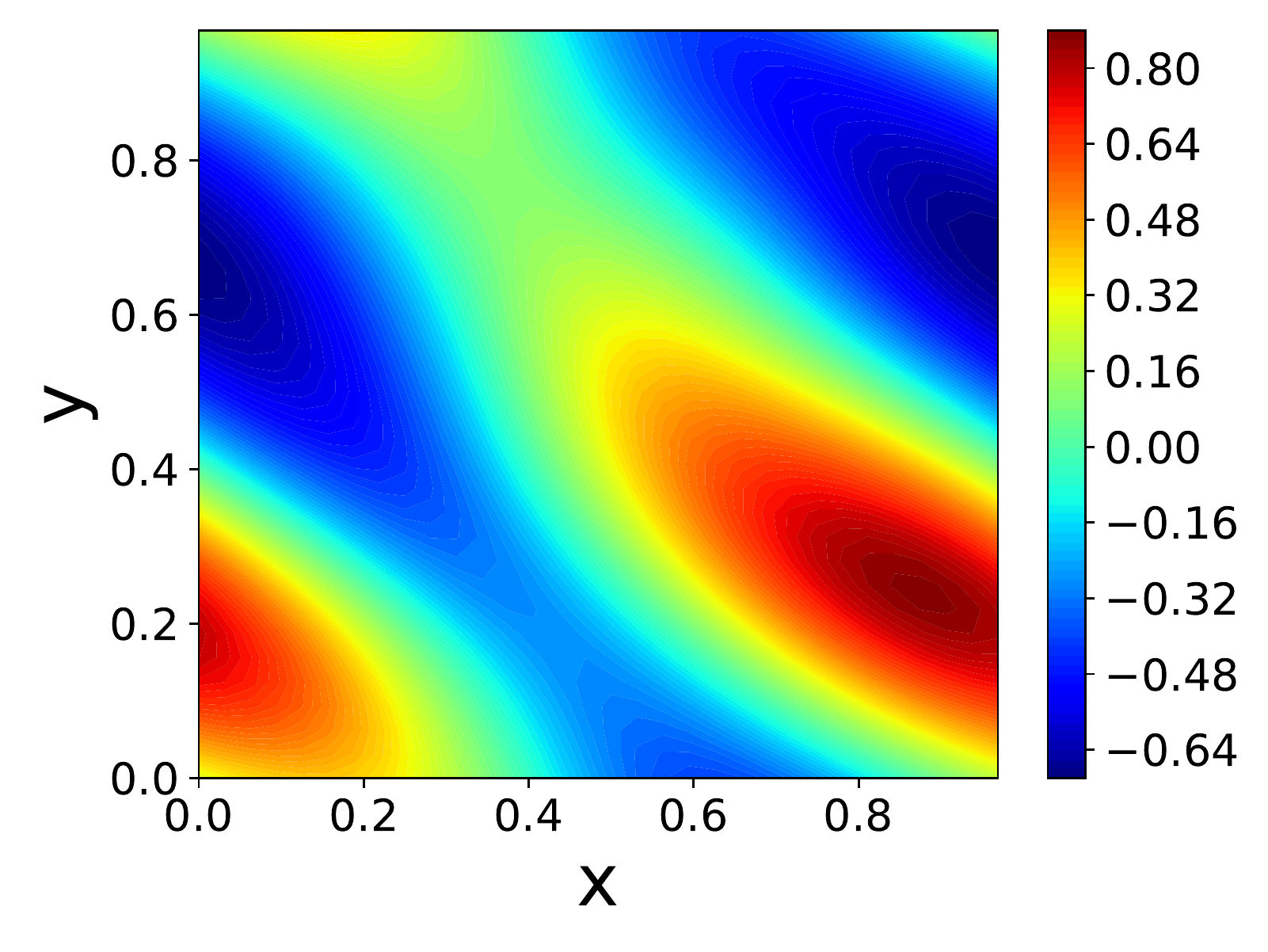}&
\includegraphics[width=0.16\textwidth, height=0.1\textheight]{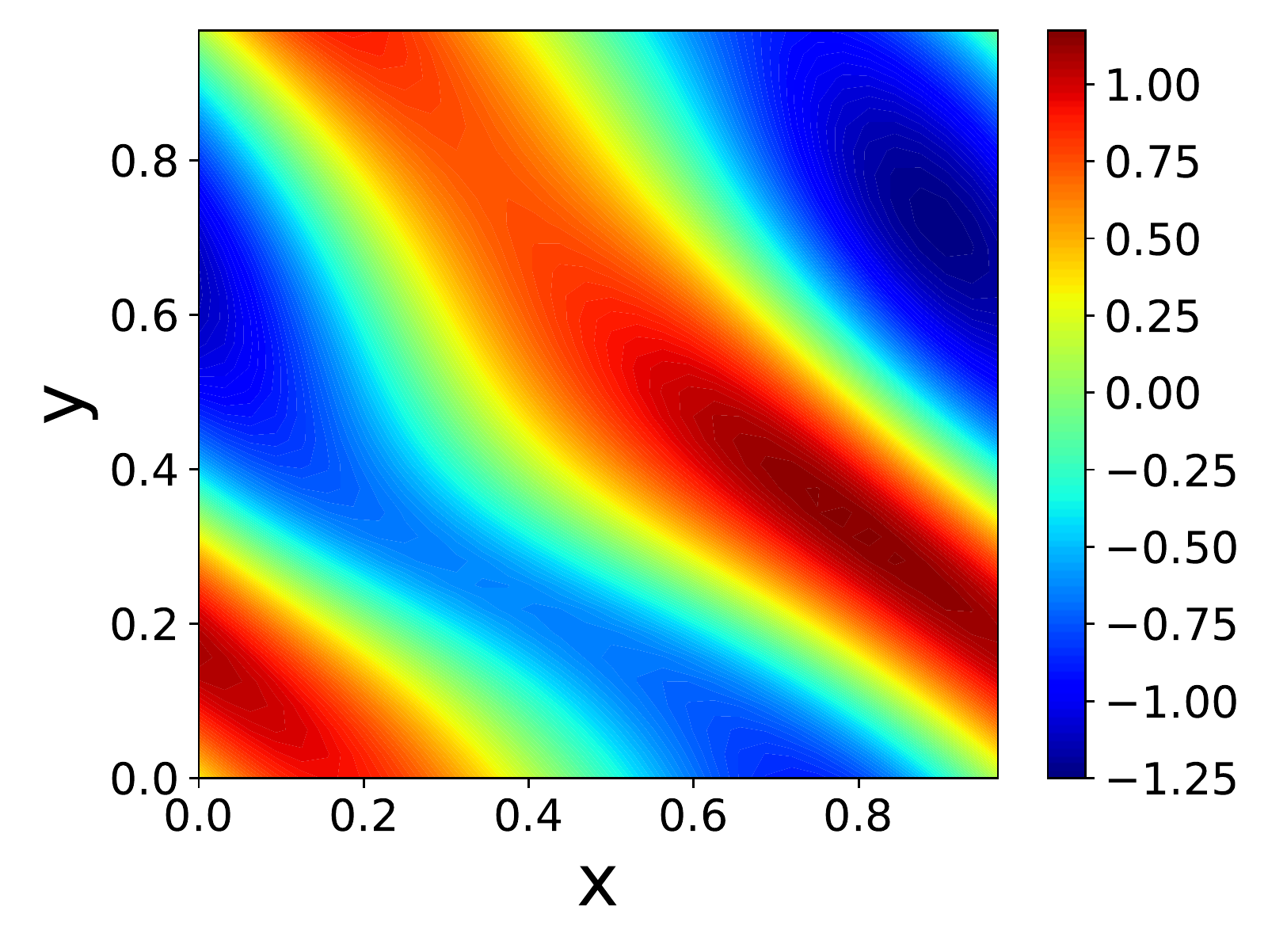}\\
\text{DeepONet}&
\raisebox{-.5\height}{\includegraphics[width=0.16\textwidth, height=0.1\textheight]{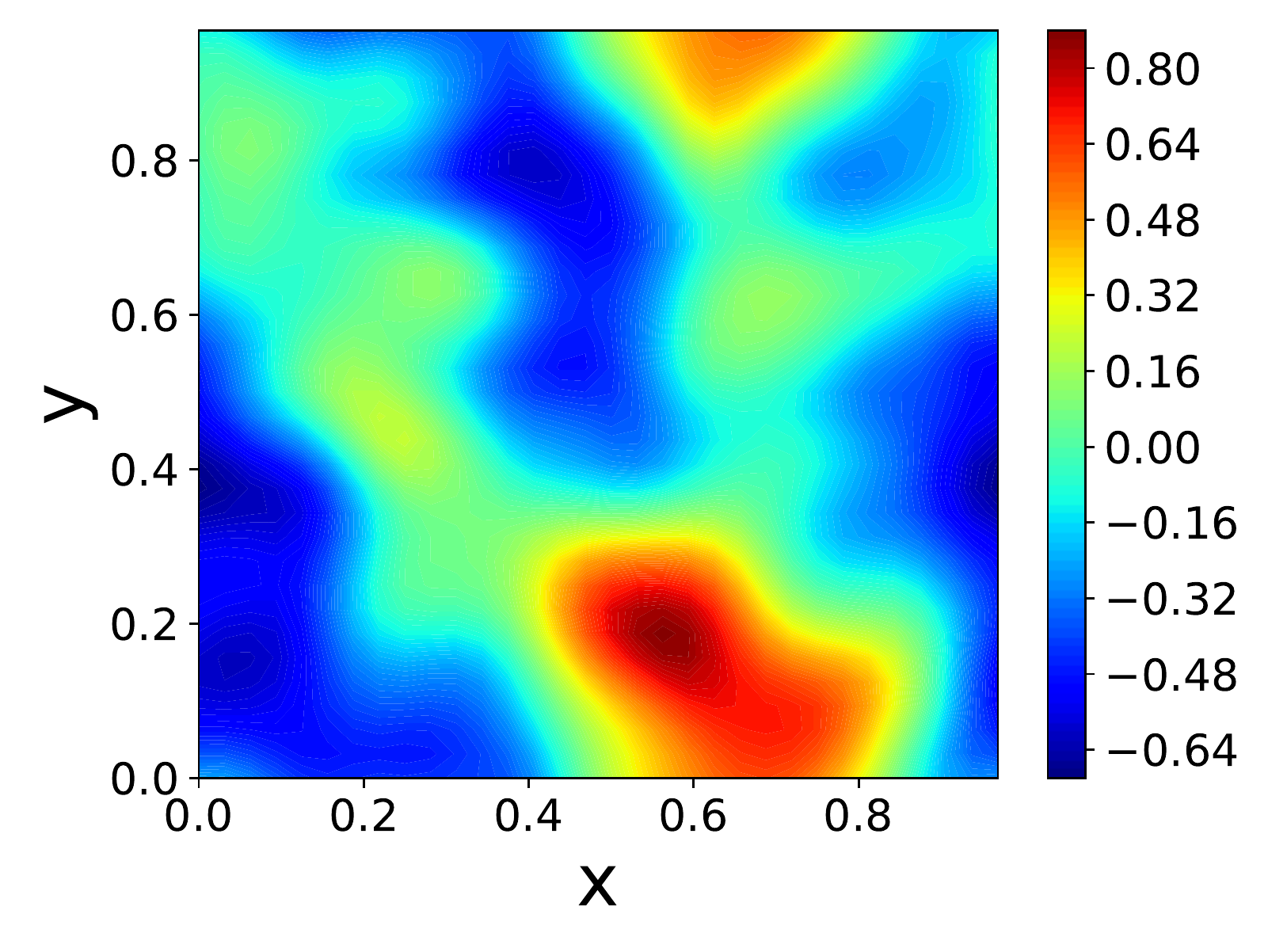}}&
\raisebox{-.5\height}{\includegraphics[width=0.16\textwidth, height=0.1\textheight]{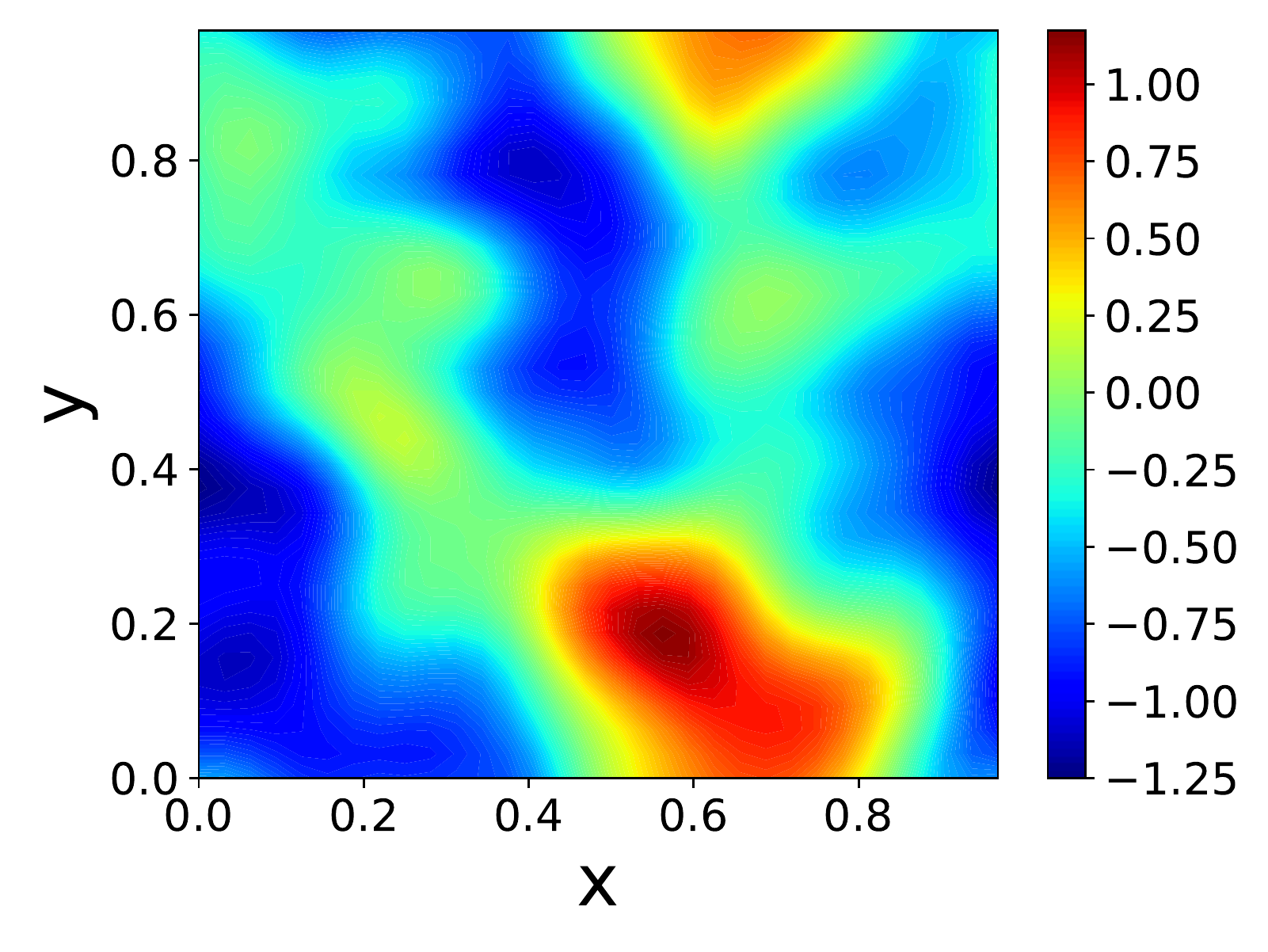}}\\
\text{TL-DeepONet}&
\raisebox{-.5\height}{\includegraphics[width=0.16\textwidth, height=0.1\textheight]{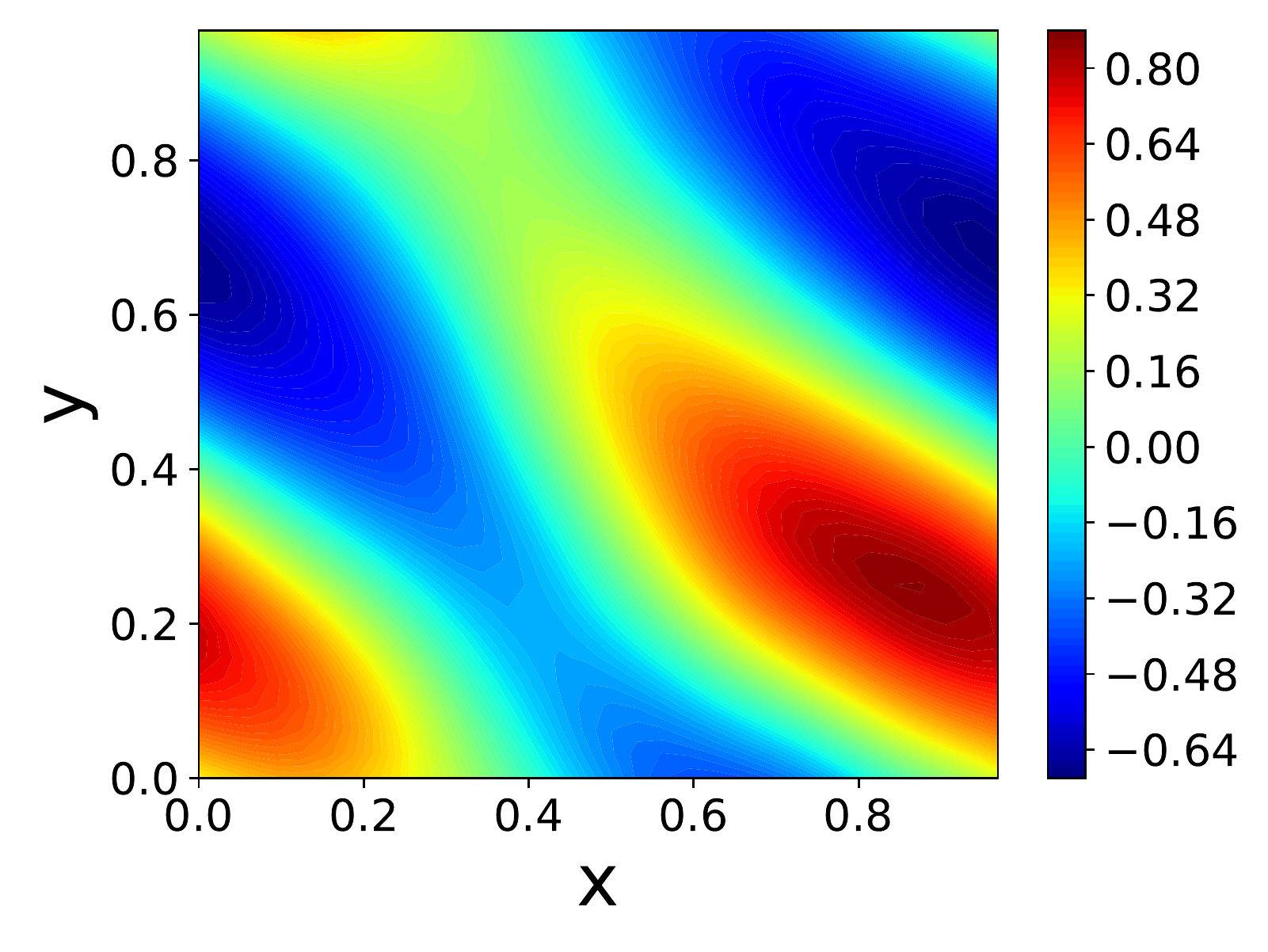}}&
\raisebox{-.5\height}{\includegraphics[width=0.16\textwidth, height=0.1\textheight]{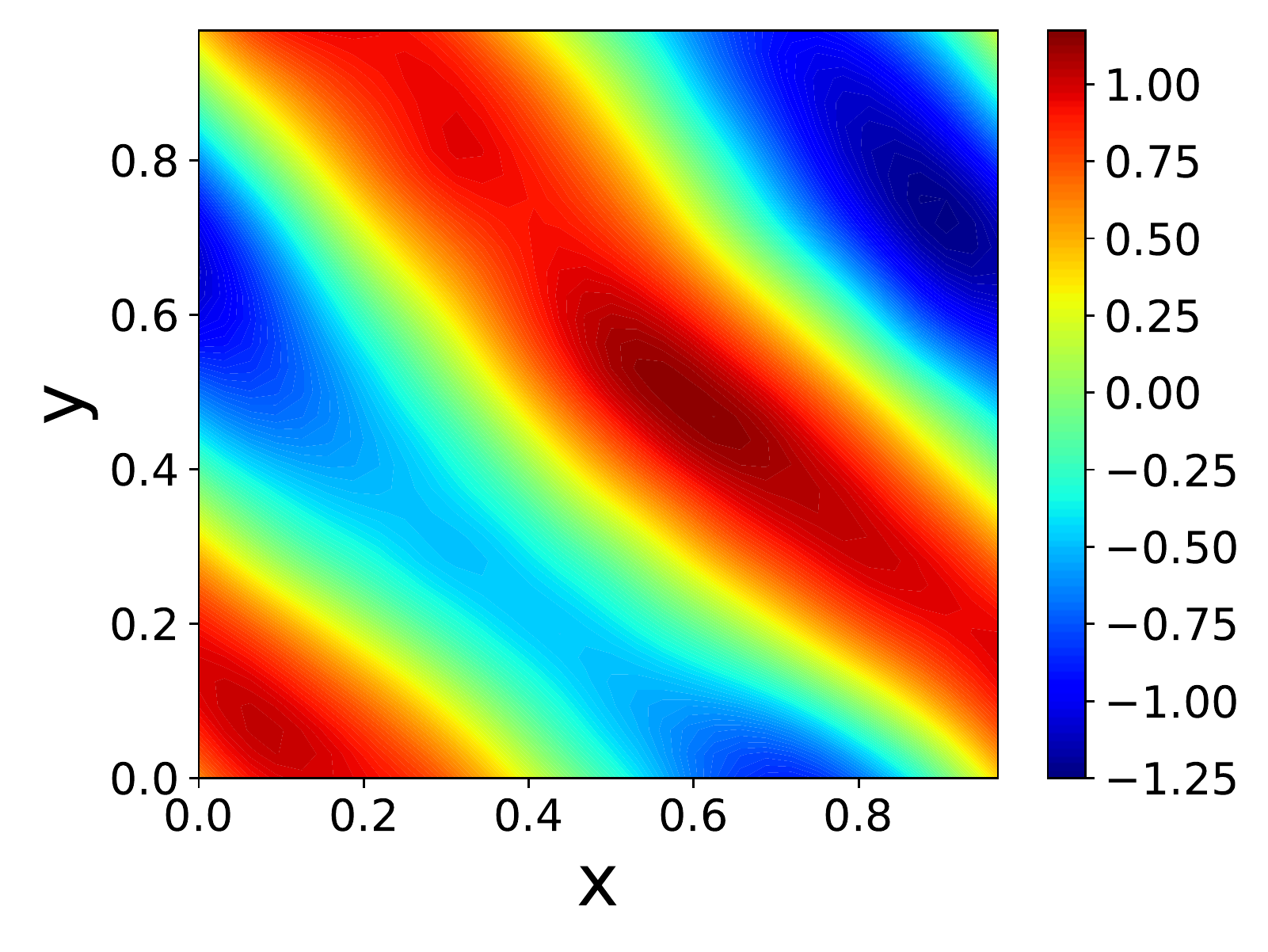}}\\
\end{tabular}
\end{center}
\caption{Results on Navier-Stokes equation with $\nu=0.001$: snapshots of reference solutions (top), and of approximate solutions predicted by DeepONet (middle) and TL-DeepONet (bottom).}
\label{fig:ns_zpzz1} 
\end{figure}

Consider the 2D Navier-Stokes equation in the vorticity form:
\begin{equation} \label{eqn:ns}
\left\{ \begin{array}{l}
\partial_{t} w(\bx, t)+u(\bx, t) \cdot \nabla w(\bx, t)=\nu \Delta w(\bx, t)+  f(\bx) , \\
w(\bx, 0) = w_0(\bx)
\end{array}\right.
\end{equation}
with periodic boundary condition and source $f(\bx) = 0.1 (\sin(2 \pi (x + y)) + \cos(2 \pi (x + y)))$. We would like to learn the propagator $\mP^{\Delta t}$ with $\Delta t = 0.01$ and apply it to predict the solution $w|_{(0,1)^2 \times (\Delta t, T)}$. Table~\ref{table:ns} shows the results with varying values of viscosity $\nu$. Note that the prediction error increases as $\nu$ decreases. TL-DeepONets reduces the errors of DeepONets by two orders of magnitudes although the prediction time of the former increases for less than 4 times. Figure~\ref{fig:ns_zpzz1} shows the snapshots of solutions to \eqref{eqn:ns} with $\nu=0.001$ at two different times.

\begin{table}[h]
\centering
\begin{tabular}{l|llll}
\toprule
\cmidrule(r){1-2}
Neural network     &  $\nu$=1e-1 &  $\nu$=1e-2 &  $\nu$=1e-3 &  $\nu$=1e-4 \\
\midrule
DeepONet   & 9.95e-1    & 1.02e0 &  9.96e-1 &  1.04e0   \\
TL-DeepONet  & \textbf{1.41e-2} &\textbf{1.07e-2} & \textbf{3.35e-2}  & \textbf{9.42e-2}   \\
\bottomrule
\end{tabular}
\caption{Results on 2D Navier-Stokes equation: the relative prediction errors within $[0, 10]$. The average trajectory prediction time is 5.3s for DeepONet and 24.8s for TL-DeepONet.}
\label{table:ns}
\end{table}

\subsection{Multiscale linear radiative transfer equation}
Consider the  linear multiscale radiative transfer equation:
\begin{equation}\label{eq:rte}
\left\{ \begin{array}{l}
\varepsilon \partial_t f + \bv \cdot  \nabla f = \frac{1}{\varepsilon} \Lop f, ~ t \in [0, T], ~ (\bx, \bv) \in \Omega_x \times \mathcal{S}^{d-1}\,, \\
f(t, \bx, \bv) = \phi(\bx) , ~ (\bx, \bv) \in  \Gamma_- \,,\\
f(0, \bx, \bv) = f_0(\bx, \bv).
\end{array}\right.
\end{equation}
Here  $\eps>0$ is the Knudsen number which is a dimensionless parameter that determines the physical regime of the equation, $\displaystyle \Lop(f) = \frac{1}{|\mathcal{S}^{d-1}|} \int_{\mathcal{S}^{d-1}}  f d \bv - f =: \average{f} - f$, and $\Gamma_-=\{(x,v): x\in \partial \Omega_x,~ v\cdot n_x <0\}$ is the inflow part of the boundary. In this example, we aim to learn the propagator $\mP^{\Delta t}$ with $\Delta t =0.01$ and employ it to predict the solution $f(t, \bx, \bv)$ for  $t\in [0,10]$. We mainly consider \eqref{eq:rte} in one and two physical dimensions and refer to Appendix \ref{sec:detailsrte} for a detailed discussion on the experiment set-up and the numerical method. Table~\ref{tab:rte} displays the results corresponding to different Knudsen numbers. The transfer learning enhanced DeepONets reduces the relative error by one or two orders of magnitude although increase the prediction time by around 4 times. Figure~\ref{fig:rte_eps_1}  and Figure~\ref{fig:rte_eps_zpzzz1} in Appendix \ref{sec:detailsrte} show several snapshots of solutions to \eqref{eq:rte} with $\eps=1$ and $\eps=10^{-4}$ respectively.

\begin{table}[ht!]
\centering
\begin{tabular}{l|lll}
\toprule
\cmidrule(r){1-2}
RTE & Neural network & $t_2$ & relative error  \\
\midrule
1D $\eps =$ 1 & DeepONet  & 4.6  & 3.06e-1    \\
& TL-DeepONet  & 22.5 & $\textbf{1.52e-2}$      \\
1D $\eps =$ 1e-4 & DeepONet  &5.1  & 3.74e-1    \\
& TL-DeepONet  & 21.9 & $\textbf{5.52e-3}$      \\
2D $\eps =$ 1 & DeepONet   & 79.8 & 3.58e-1   \\
& TL-DeepONet  &   431.3 & $\textbf{2.19e-2}$   \\
2D $\eps =$ 1e-4 & DeepONet   &83.1 & 2.37e0   \\
& TL-DeepONet  & 379.3 &  $\textbf{8.93e-3}$   \\
\bottomrule
\end{tabular}
\caption{Results on the radiative transfer equation: the relative prediction errors over the time-horizon $[0,10]$. }
\label{tab:rte}
\end{table}

\section{Conclusion}
In this paper, we proposed a new physics-informed DeepONet based on transfer learning for learning evolutionary PDEs. This is achieved in two steps: first learn the propagators and then predict the solutions by  successive actions of propagators on the initial condition. The experimental results demonstrated that the proposed method improves substantially upon the vanilla DeepONet in terms of long-time accuracy and stability while maintains low computational cost. The proposed method also reduced the training sample size needed to achieve the same order of prediction accuracy of the vanilla DeepONets.

\section*{Acknowledgement}
L. Wang is partially supported by
NSF grant DMS-1846854. Y. Lu thanks NSF for the support via the award DMS-2107934.

\bibliography{reference.bib}
\newpage 


\renewcommand{\thesection}{\Alph{section}}
\setcounter{section}{0}

\clearpage


\section{Table of notations}
A table of notations is given in Table~\ref{tab:notation_table}.
\begin{table}
\caption{Table of notations}
\label{tab:notation_table}
\centering
\begin{tabular}{lll}
\toprule
\cmidrule(r){1-2}
Notation & Meaning    \\
\midrule
$\mathcal{P} \text{ or }  \mP^{\Delta t} $     & Target propagator        \\
$\mathcal{P}_{NN}$     &    Neural network approximator   \\
$\mathcal{U}$     &  The Banach space of input functions  \\
$\{ \bx^r_i \}_{i=1}^{N^r_x}$ & The interior sensors \\
$\{ \bx^b_i \}_{i=1}^{N^b_x}$ & The boundary sensors \\
$\{ f_s \}_{s=1}^{N_s}$ & Randomly sampled functions used for training \\
$\theta, \xi$ & Neural network parameters \\
$b^{NN}(\cdot, \theta)$ & The branch net of DeepONet \\
$h^{NN}(\cdot, \theta)$ & The output functions defined by the last \\
& hidden layer in the branch net  \\
$t^{NN}(\cdot, \xi)$ & The trunk net of DeepONet \\
$w = \{ w_j \}_{j=1}^{q}$ &  Weights in the last layer of the branch net \\
$N_p$ & Number of sensors \\
$N_c$ & Number of grid points \\
& in transfer learning step \\
$M$ & Number of training sample pairs \\
$p$ & Number of output features\\
$q$ & \vtop{\hbox{\strut Number of tunable weights at the last hidden}\hbox{\strut layer of the branch net}}\\
\bottomrule
\end{tabular}
\end{table}

\section{Theoretical analysis}

\subsection{Validation of assumption \eqref{assump}} \label{sec:assump}
Here we will give a simple justification for the assumption \eqref{assump}. Consider the $L^2$ norm as an example:
\[
\norm{\mP}_2 = \sup_{f \in L^2(\Omega_x), \norm{f}_2 = 1} \norm{\mP f}_2\,,
\]
then \eqref{assump} is fulfilled if the underlying dynamics is stable under this norm:
\begin{equation*}
\frac{d}{dt}\norm{f}_2^2 (t) \leq 0\,,
\end{equation*}
where $f$ is the solution to \eqref{111}. Indeed, multiplying \eqref{111} by $f$ and integrating in $\bx$, we have 
\begin{align*}
\frac{d}{dt} \frac{1}{2} \norm{f}_2^2 = \int_{\Omega_x} f \mathcal L f  \rd \bx \leq 0\,.
\end{align*}
Then for the semi-discrete in time version 
\begin{equation*}
f^{n+1}- f^n = \Delta t \mathcal L f^{n+1}
\end{equation*}
multiplying it by $f^{n+1}$ and integrating in $\bx$, it becomes 
\begin{align*} \label{0502}
\half ( \norm{f^{n+1}}_2^2 - \norm{f^n}_2^2 &+ \norm{f^{n+1}-f^n}_2^2 ) \\ &=  \int_{\Omega_x} f^{n+1} \mathcal L f^{n+1} \rd x\leq 0\,,
\end{align*}
which readily leads to 
\begin{equation*}
\norm{f^{n+1}}_2 = \norm{\mP f^n}_2 \leq \norm{f^n}_2\,.
\end{equation*}

\subsection{Proof of Theorem~\ref{thm:main}}\label{pf}
\begin{proof}
From \eqref{assump}-\eqref{loss0}, one sees that
\begin{equation}\label{eq:diffP}
\begin{aligned}
\sup_{f\in \mU}\norm{\mP f - \mP_{NN} f}_\mX & = \sup_{f\in \mU}\norm{ \mP (f - \mP^{-1} \mP_{NN} f)}_\mX \\
& \leq \delta \norm{f}_\mX.
\end{aligned}
\end{equation}
Therefore one obtains that 
\[
\sup_{f\in \mU, \norm{f}_\mX = 1}\norm{\mP_{NN} f}_\mX \leq  1 + \delta \,.
\]
Let $f\in \mU$ with $\|f\|_\mX= 1$. It follows from above that 
\begin{align}
& \norm{(\mP)^K f - (\mP_{NN})^K f}_\mX  \nonumber \\
&= \norm{(\mP - \mP_{NN}) (\mP^{K-1} + \mP^{K-2} \mP_{NN} + \cdots + \mP_{NN}^{K-1}) f}_{\mX} \nonumber
\\ & \leq  \sum_{l=0}^{K-1} \norm{(\mP  - \mP_{NN}) \mP^l \mP_{NN}^{K-1-l} f}_{\mX} \label{0501}
\\ & \leq \delta K (1+\delta )^K \nonumber \,,
\end{align}
which proves \eqref{sta1} after taking supreme on $f$. 

Next, if \eqref{assump2} holds, and $\delta$ is chosen to be $\delta \leq \half(1-\eta)$, then we have from \eqref{eq:diffP} that
\[
\sup_{f\in \mU, \norm{f}_\mX = 1} \norm{\mP_{NN} f} \leq \eta + \delta \leq \frac{1+\eta}{2}.
\]
Inserting above into 	\eqref{0501} leads to 
$$\begin{aligned}
\norm{(\mP)^K f - (\mP_{NN})^K f}_\mX & \leq 
\sum_{l=0}^{K-1} \norm{(\mP  - \mP_{NN}) \mP^l \mP_{NN}^{K-1-l} f}_{\mX}\\
& \leq
\delta \sum_{l=0}^{K-1} \eta^l (\eta+\delta)^{K-1-l} \\
& \leq  \delta K \Big(\frac{1+\eta}{2}\Big)^{K-1}.
\end{aligned}
$$
This proves \eqref{est2}. 
\end{proof}

\section{Experiment details}
In this section, we provide the details on the numerical experiments of Section \ref{sec:numerical_exp}.

\subsection{Data generation and configuration of training}\label{sec:data}
\paragraph{Data generation} For all numerical experiments, we use uniform mesh with $N_x^d$ grid points for discretization of the spatial domain $\Omega_x \subseteq \mathbb{R}^d$, and $N_v$ Gaussian quadrature points for discretizing the velocity variable $\bv \in \mathcal{S}^{d-1}$ in the radiative transfer equation only. We generate $N_s$ initial conditions $\{ f_s (\bx) \}_{s=1}^{N_s}$ for training and $N_e$ functions $\{ f_e (\bx) \}_{e=1}^{N_e}$ for testing.  
The training set consists of $N_b$ functions that sampled from a centered Gaussian random field as well as forward passes of those functions through up to $n_t$ times actions of the propagator. This gives $N_s = n_t \times N_b$ training functions. The random functions may be post-processed so that they satisfy the boundary condition of the PDE. Details on the post-processing methods can be found in the subsequent sections. The final training data of size $M$
is constructed as a subset of a larger training set of size $N_s\times N_p$, which consists of pointwise evaluations of $N_s$ randomly sampled functions at $N_p$ physical locations (sensors).
Unless otherwise specified, we set $N_b=100$, $n_t=20$ in 1D test problems and $N_b=100$, $n_t=100$ in 2D test problems. Here a training set of size $M=1000$ may only use 50 functions evaluating at 20 grid points in the domain.

\paragraph{Two error measures.} To quantify the performance of our neural nets, we measure two relative errors of neural operator approaches. The first is the relative error at a single time-step $t^k:= k \Delta t$:
\begin{equation}\label{eqn:avg_l2_loss_t}
\frac{1}{N_e} \sum_{j=1}^{N_e} \sqrt{ \frac{{  \sum_{i=1}^{N_x} \left( \mathcal{P}_{NN}^{k}(f_j)(\bx^r_i) - \mathcal{P}^{k}(f_j)(\bx^r_i)  \right)^2  }}
{{ \sum_{i=1}^{N_x} \left( \mathcal{P}^{k}(f_j)(\bx^r_i)  \right)^2   }}}\,,
\end{equation}
and the second is the relative error over a long time horizon (or equivalently multiple time steps) 
\begin{equation}\label{eqn:avg_l2_loss}
\displaystyle \sqrt{\frac{{  \sum_{i=1}^{N_x}\sum_{j=1}^{N_e}\sum_{n=1}^K \left( \mathcal{P}_{NN}^{n}(f_j)(\bx^r_i) - \mathcal{P}^{n}(f_j)(\bx^r_i)  \right)^2  }}
{{  \sum_{i=1}^{N_x}\sum_{j=1}^{N_e}\sum_{n=1}^K \left( \mathcal{P}^{n}(f_j) (\bx^r_i) \right)^2  }}}\,.
\end{equation}

\paragraph{Neural networks and training parameters} In 1D examples, we use the modified fully connected architecture with depth of 5 layers and width of 100 neurons for both branch and trunk nets, and the design of the optional layers $P$ and $D$ in Figure~\ref{fig:tl_pidon_arch} will be detailed in each of the following examples. The batch size is chosen to be 100 with ADAM optimizer, where the initial learning rate $lr=0.001$ and a 0.95 decay rate in every 5000 steps. Same architecture is used in 2D examples except that a depth of 6 layers is used. In the transfer learning step, to solve the optimization problem \eqref{eqn:opt_w}, we use 
the lstsq function (with rcond=1e-6) from Numpy\cite{harris2020array} for linear operators and the leastsq function in Scipy\cite{2020SciPy-NMeth} (using default setting with $ftol=$1e-5, $xtol=$1e-5) for nonlinear operators. All of the neural networks are trained on a single K40m GPU, and the prediction step is computed on a AMD Ryzen 7 3700x Processor. 


\subsection{Further details on reaction diffusion equation}\label{sec:reaction}
To generate initial conditions that satisfy zero boundary condition, i.e., $f_0(0)=f_0(1)=0$ we first sample $a(x) \sim \mathcal{GP} (0, K_l(x_1, x_2))$ with 
\begin{equation}\label{eqn:kl}
K_l(x_1, x_2) = e^{-\frac{(x_1 - x_2)^2}{2l^2}}, 
\end{equation}
and then let $f_0(x) = a(x)x(1-x)$. Likewise, to enforce the same boundary condition for the output of the neural net, i.e.,  $\mathcal{P}_{NN}(f)(1)=\mathcal{P}_{NN}(f)(1)=0$, we employ an additional layer $D$ in Figure~\ref{fig:tl_pidon_arch} that multiplies the output of trunk nets by $x(1-x)$. 

\subsubsection{Crank-Nicolson scheme for reaction diffusion equation}\label{2nd}
To demonstrate the improvement on the efficiency of using the higher order in time scheme at the transfer learning step, we apply Crank-Nicolson scheme for the nonlinear reaction diffusion equation. As displayed in Table~\ref{table:nrd_2nd}, the second order scheme with $\Delta t = 0.4$ reduces the prediction time compared with first order scheme with $\Delta t = 0.05$ by a factor of $1/6$. 
\begin{table}[h]
\centering
\begin{tabular}{l|l|l|l}
\toprule
\cmidrule(r){1-2}
Neural network    & $\Delta t$    & $t_2$ & Relative error \\
\midrule
TL-DeepONet  & $\Delta t = 0.05$ & 7.1  &  \textbf{1.78e-3}    \\
TL-DeepONet 2nd & $\Delta t = 0.1$ & 4.53   &  \textbf{4.91e-4}   \\
& $\Delta t = 0.2$ & 2.36  & \textbf{2.63e-3}    \\ 
& $\Delta t = 0.4$ & 1.24  & \textbf{9.53e-3}    \\ 
\bottomrule
\end{tabular}
\caption{Comparison of first order and second order in time method for reaction diffusion equation with various $\Delta t$. Here $t_2$ is the averaged  time of predicting the solution trajectories among the time interval $[0, 50]$ based on $30$ test initial conditions.}
\label{table:nrd_2nd}
\end{table}

\subsubsection{Implementation details on Table~\ref{table:nrd} and Table ~\ref{table:nrd_2nd}}
Consider 1D nonlinear reaction diffusion equation \eqref{eqn:nrd} with $d=k=0.001$. We use $N_e=30$ test functions drawn from the Gaussian process defined above with the length scale $l=0.2$ in \eqref{eqn:kl} and we use $N_x = 64$ uniform spatial grids for spatial discretization. The loss functions of CONT DeepONet and CONT TL-DeepONet (c.f. \eqref{eqn:bc_loss}) are calculated using 20 uniform temporal steps on $[0, 1]$. For DeepONet and TL-DeepONet, we set the maximum iteration number $N_{iter} = 100000$ and adopt the stopping criterion that the empirical loss is below 1e-6. For CONT DeepONet and CONT TL-DeepONet, we set maximum iteration number $N_{iter} = 200000$ and use stopping criterion that the empirical loss is below 1e-6. We use $p=100$ features for all four  operator networks. In the transfer learning steps of  CONT TL-DeepONet, we subsample $N_c=400 < 64\times 20$ grid points and update $q = 25$ weights defined in \eqref{eqn:approx_trans}. In the transfer learning steps of TL-DeepONet, we set $N_c = 32 < 64$ and $q = 15$ instead. Additionally, we fix $\Delta t = 0.05$ in Table~\ref{table:nrd} and $M = 3000$ in Table~\ref{table:nrd_2nd}.

\subsection{Further details on Allen-Cahn and Cahn-Hilliard equation}\label{sec:detail_ac_ch}
In all three examples, we consider periodic boundary conditions. To this end, the initial condition is generated from $f_0(\bx) \sim \mathcal{GP} (0, K^p_l(\bx_1, \bx_2))$, where the covariance kernel has the desired periodicity. In particular, the kernel in one dimension reads:
\begin{equation}\label{eqn:kp_1d}
\displaystyle
K^p_l(x_1, x_2) = e^{-\frac{\sin^2(\pi(x_1-x_2))}{2l^2}} \,
\end{equation}
and in two dimension takes the form: 
\begin{equation}\label{eqn:kp_2d}
\displaystyle
K^p_l(\bx_1, \bx_2) = e^{-\frac{\sin^2(\pi(\bx_{1,1}-\bx_{2,1}))+\sin^2(\pi(\bx_{1,2}-\bx_{2,2}))}{2l^2}}.
\end{equation}
To enforce the periodic boundary condition to the output of the trunk net, we employ an additional layer $P$ (see Figure~\ref{fig:tl_pidon_arch}) in the truck net, which upsizes $x$ to $\{ \cos{2 \pi x}, \sin{2 \pi x} \}$.This way, the input of the trunk net already has the desired periodicity and will be maintained throughout. 
Analogously, an additional layer $P$, which plays the role of upsizing $(x,y)$ to 
$\{ \cos{2 \pi x}, \sin{2 \pi x} , \cos{2 \pi y}, \sin{2 \pi y}\}$, is leveraged in the trunk net in 2D case.

\subsubsection{Implementation details on Figure~\ref{fig:failure} and Table~\ref{table:ac}, ~\ref{table:ac_2d}}
Consider Allen-Cahn equation \eqref{eqn:ac} with $d_2=0.1$. In 1D case, we use $N_e=30$ test functions drawn from the Gaussian process defined above with the length scale $l=0.5$ in \eqref{eqn:kp_1d} and use $N_x = 64$ uniform spatial grids for spatial discretization. For DeepONet and TL-DeepONet, we let time step size $\Delta t = 0.05$, set the maximum iteration number $N_{iter} = 100000$ and adopt the stopping criterion that the empirical loss is below 1e-6. For CONT DeepONet and CONT TL-DeepONet, we use additional 20 uniform grids on time span $[0,1]$, set maximum iteration number $N_{iter} = 200000$ and use stopping criterion that the empirical loss is below 1e-6. We use $p=100$ features for all four  operator networks. In the transfer learning steps of CONT TL-DeepONet, we subsample $N_c=400 < 64\times 20$ grid points and update $q = 25$ weights defined in \eqref{eqn:approx_trans}. In the transfer learning steps of TL-DeepONet, we set $N_c = 32 < 64$ and $q = 15$ instead. In 2D case, we use $N_e=30$ test functions drawn from the Gaussian process defined above with the length scale $l=1$ in \eqref{eqn:kp_2d}. We use $\Delta t = 0.01$ for time step size and $N_x = N_y = 20$ uniform spatial grids for spatial discretization. We set the maximum iteration number $N_{iter} = 200000$, adopt the stopping criterion that the empirical loss is below 1e-6 and use number of feature $p=120$. In the transfer learning step, we subsample $N_c = 144 < 20 \times 20$ and update $q = 40$ weights defined in \eqref{eqn:approx_trans}. Additionally, we fix $d_1=0.0005$ and $M=3000$ in Figure~\ref{fig:failure}. For FNO in Figure~\ref{fig:failure}, we choose the  time step size $\Delta t = 0.05$, prepare 50 Input \& Ouput function pairs, set the maximum iteration number $N_{iter} = 100000$ and adopt the stopping criterion that the empirical loss is below 1e-6.

\subsubsection{Implementation details on Table~\ref{table:ch} and Figure~\ref{fig:ch2d}}
Consider Cahn-Hilliard equation \eqref{eqn:ch} with $d_2=0.001$. In 1D case, we use $N_e=30$ test functions drawn from the Gaussian process defined above with the length scale $l=0.5$ in \eqref{eqn:kp_1d}. We use $\Delta t = 0.05$ for time step size and $N_x = 64$ uniform spatial grids for spatial discretization. We set the maximum iteration number $N_{iter} = 100000$, adopt the stopping criterion that the empirical loss is below 1e-6 and use number of feature $p=100$. In the transfer learning step, we subsample $N_c = 32 < 64$ and update $q = 15$ weights defined in \eqref{eqn:approx_trans}. In 2D case, we use $N_e=30$ test functions drawn from the Gaussian process defined above with the length scale $l=1$ in \eqref{eqn:kp_2d}. We use $\Delta t = 0.01$ for time step size and $N_x = N_y = 20$ uniform spatial grids for spatial discretization. We set the maximum iteration number $N_{iter} = 200000$, adopt the stopping criterion that the empirical loss is below 1e-6 and use number of feature $p=100$. In the transfer learning step, we subsample $N_c = 144 < 20 \times 20$ and update $q = 25$ weights defined in \eqref{eqn:approx_trans}. Additionally, we fix $d_1=$2e-6 and $M=30000$ in Figure~\ref{fig:ch2d}.

\subsection{Further details on Navier-Stokes equation}
The data generation of Navier-Stokes equation and the design of optional layer $P$ in Figure~\ref{fig:tl_pidon_arch} are exactly same as the one to Allen-Cahn and Cahn-Hilliard equation in Section~\ref{sec:detail_ac_ch}.

\subsubsection{Calculation of Navie-Stokes equation}
By introducing the stream function $\psi(x,t)$, the velocity field and the vorticity can be found from $u(x,t) = (\frac{\partial \psi}{\partial y}, - \frac{\partial \psi}{\partial x})$ and $w(x,t) = -\Delta \psi(x,t)$. Therefore, we rewirte the equation \eqref{eqn:ns} as
\begin{equation}
\left\{ \begin{array}{l}
w = - (\partial_{xx} \psi +  \partial_{yy} \psi), \\
\partial_{t} w+ \partial_y \psi \partial_x w - \partial_x \psi \partial_y w - \nu (\partial_{xx} w +  \partial_{yy} w) - \nu f = 0, \\
w(x, 0) = w_0(x)\,,
\end{array}\right.
\end{equation}
and use the following semi-discretization scheme in the computation of the numerical solutions:
\begin{equation}\nonumber
\left\{ 
\begin{aligned}
w^{n+1} - &w^n + \Delta t \partial_y \psi^n \partial_x w^{n+1} - \partial_x \psi^n \partial_y w^{n+1} \\ & - \nu (\partial_{xx} w^{n+1} +  \partial_{yy} w^{n+1}) - \nu f = 0, \\
w^{n+1} + &(\partial_{xx} \psi^{n+1} +  \partial_{yy} \psi^{n+1}) = 0.
\end{aligned} 
\right.
\end{equation}

 \subsubsection{Supplementary examples of Navier-Stokes equation}
\begin{figure*}
 \centering
 \begin{center}
 \setlength{\tabcolsep}{-1pt}
 \renewcommand{\arraystretch}{-1}
 \begin{tabular}{cccc}
 \text{Initical condition} \quad & $t=1$ & $t=4$ & $t=10$\\
 \includegraphics[width=0.24\textwidth, height=0.12\textheight]{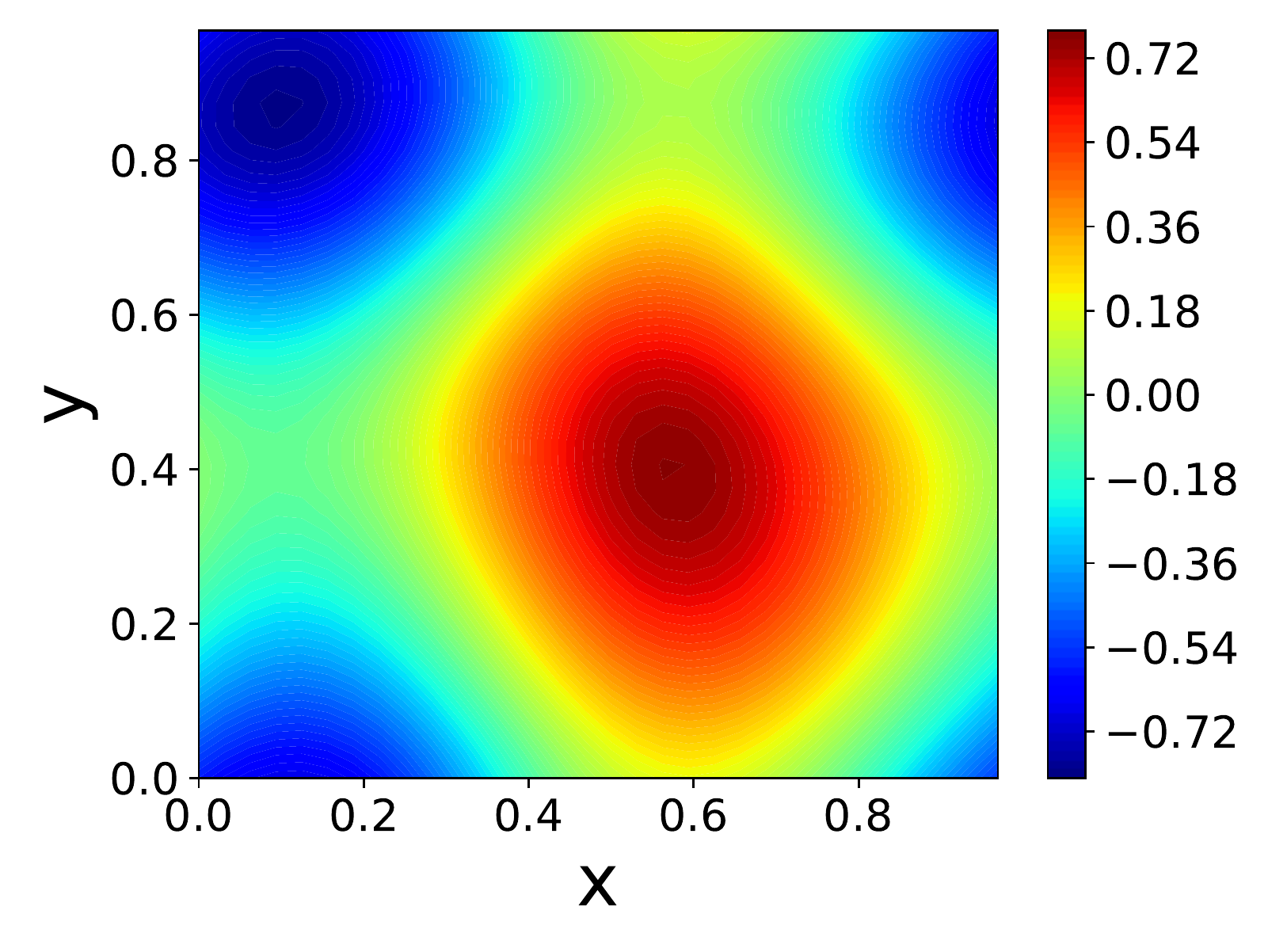}&
 \includegraphics[width=0.24\textwidth, height=0.12\textheight]{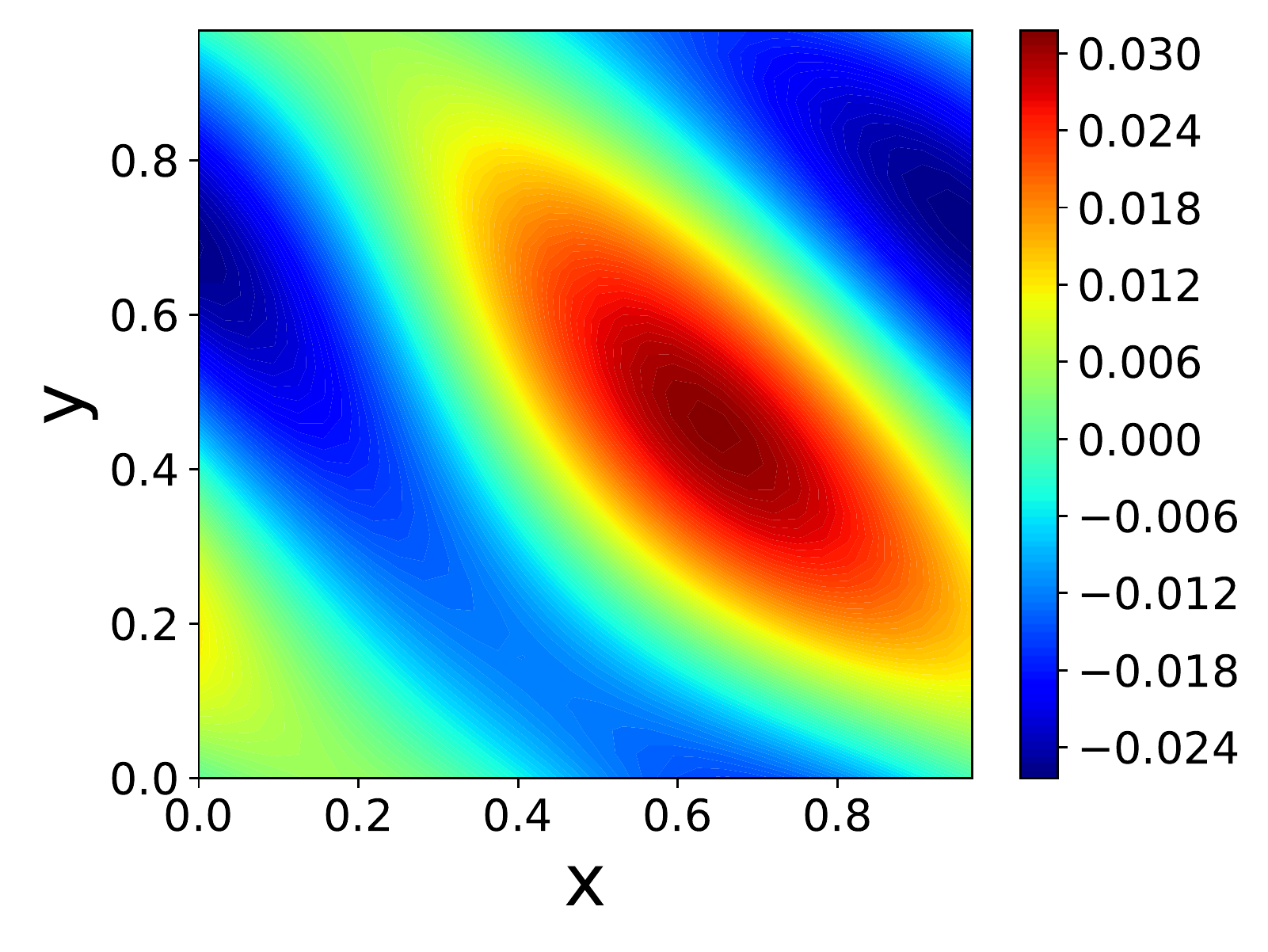}&
 \includegraphics[width=0.24\textwidth, height=0.12\textheight]{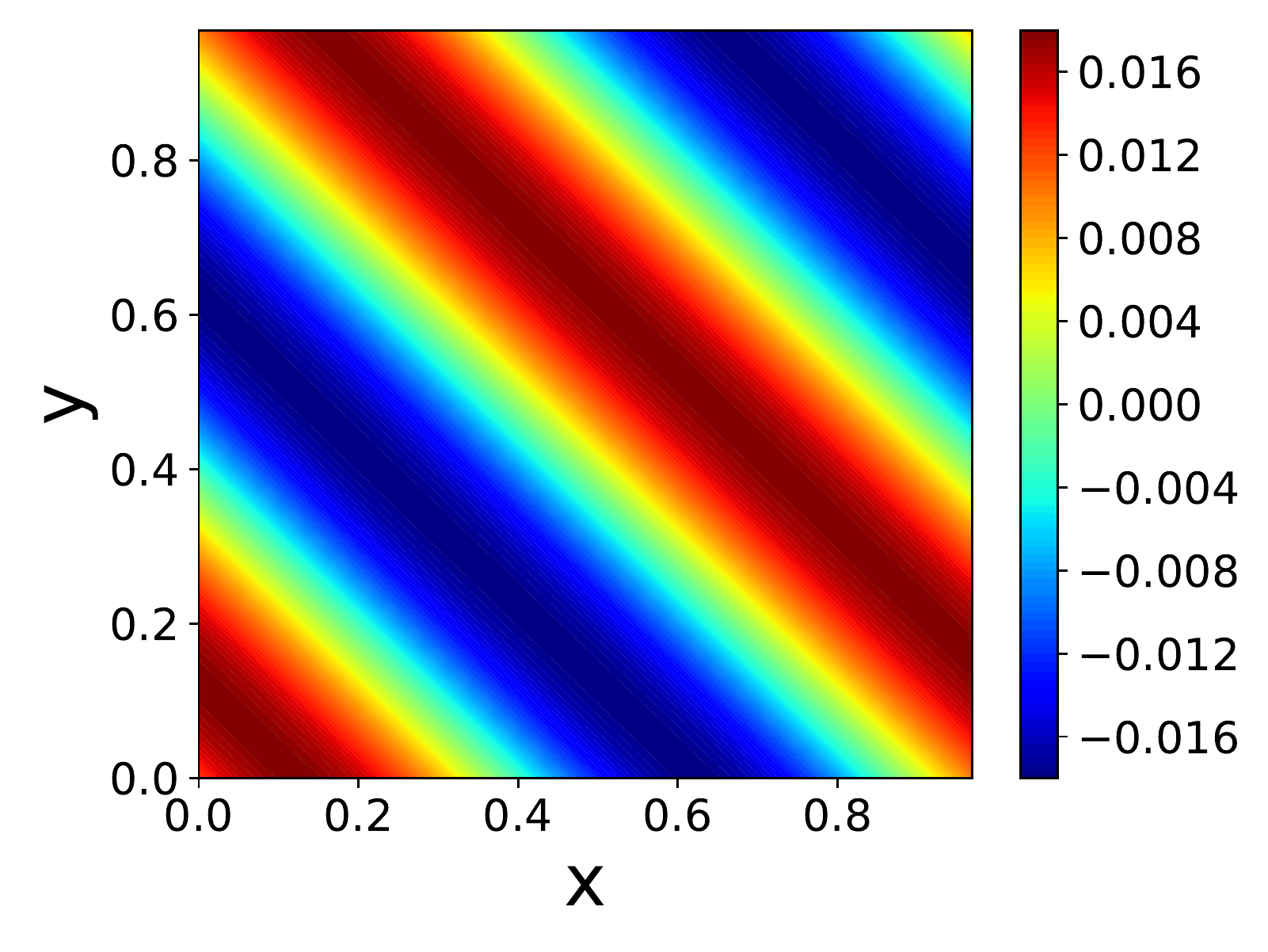}&
 \includegraphics[width=0.24\textwidth, height=0.12\textheight]{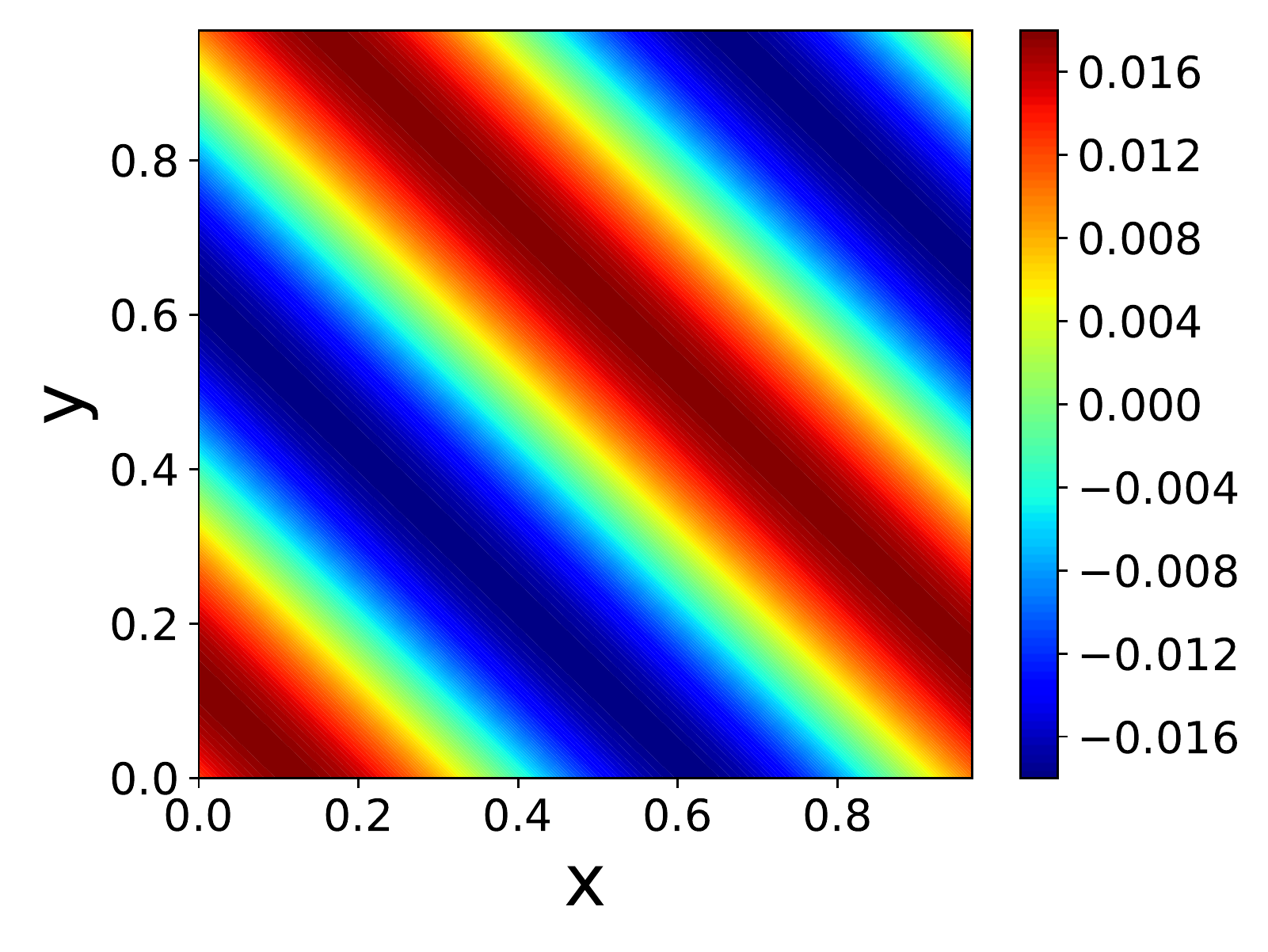}\\
 \text{DeepONet}&
 \raisebox{-.5\height}{\includegraphics[width=0.24\textwidth, height=0.12\textheight]{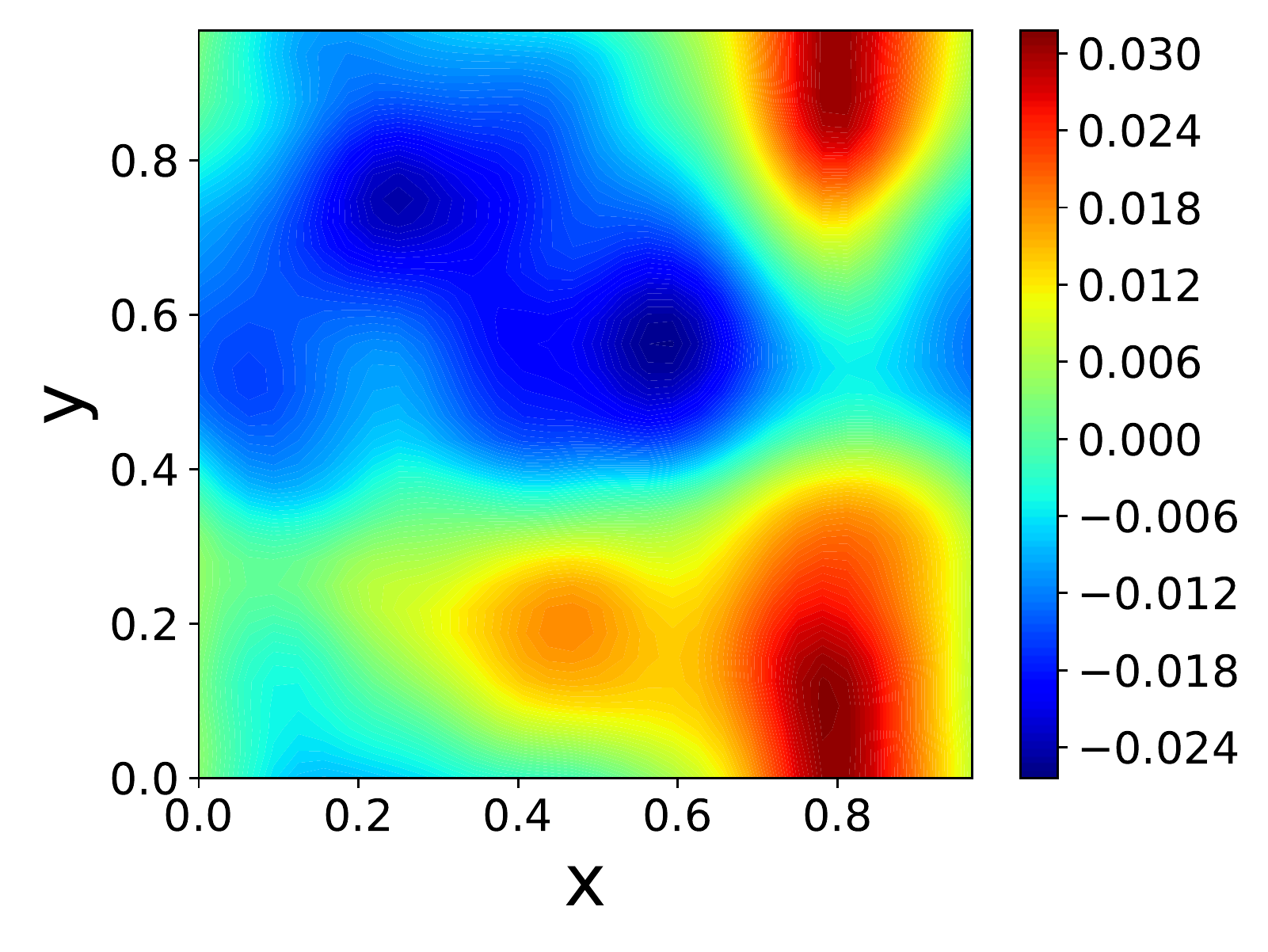}}&
 \raisebox{-.5\height}{\includegraphics[width=0.24\textwidth, height=0.12\textheight]{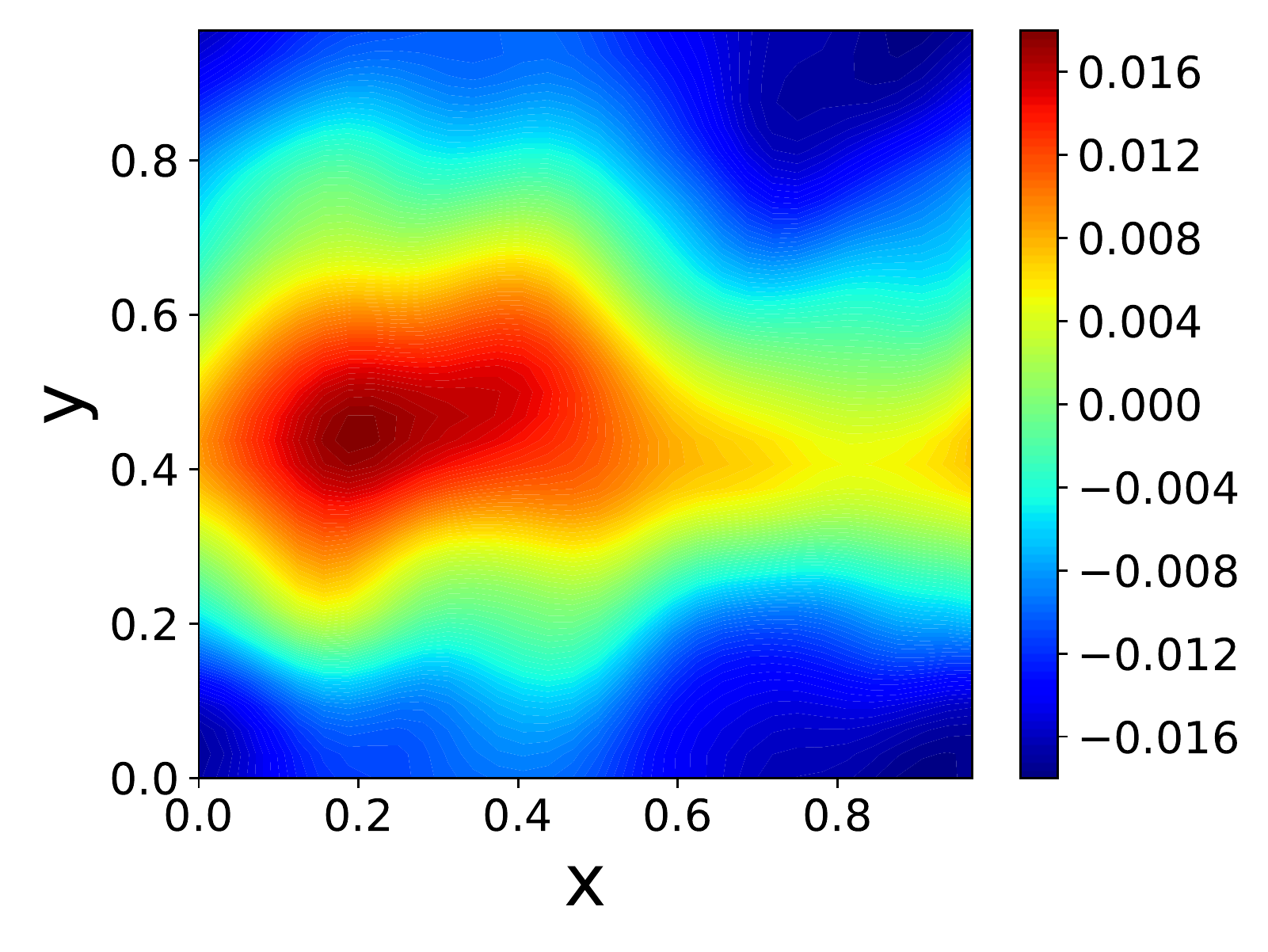}}&
 \raisebox{-.5\height}{\includegraphics[width=0.24\textwidth, height=0.12\textheight]{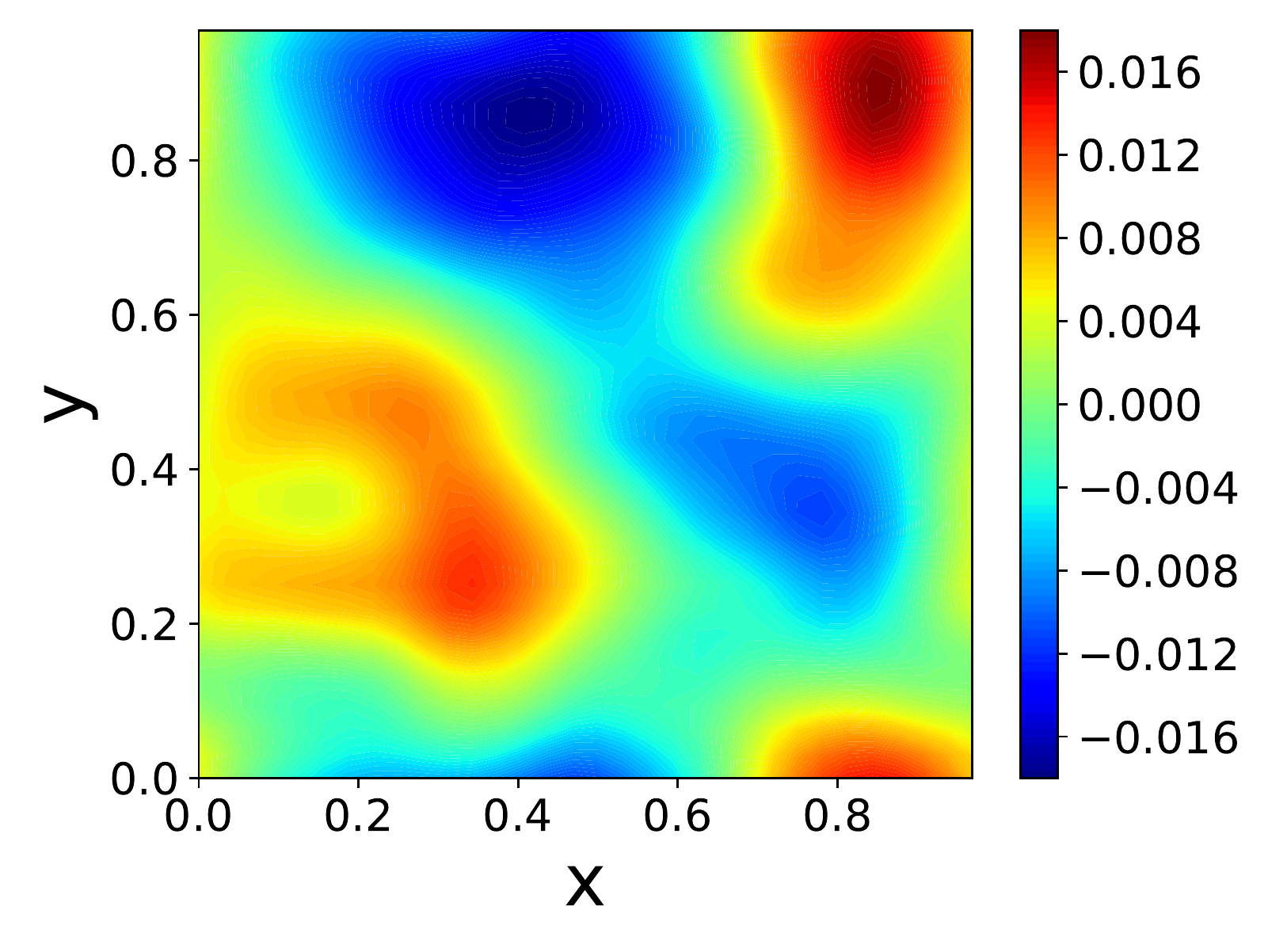}}\\
 \text{TL-DeepONet}&
 \raisebox{-.5\height}{\includegraphics[width=0.24\textwidth, height=0.12\textheight]{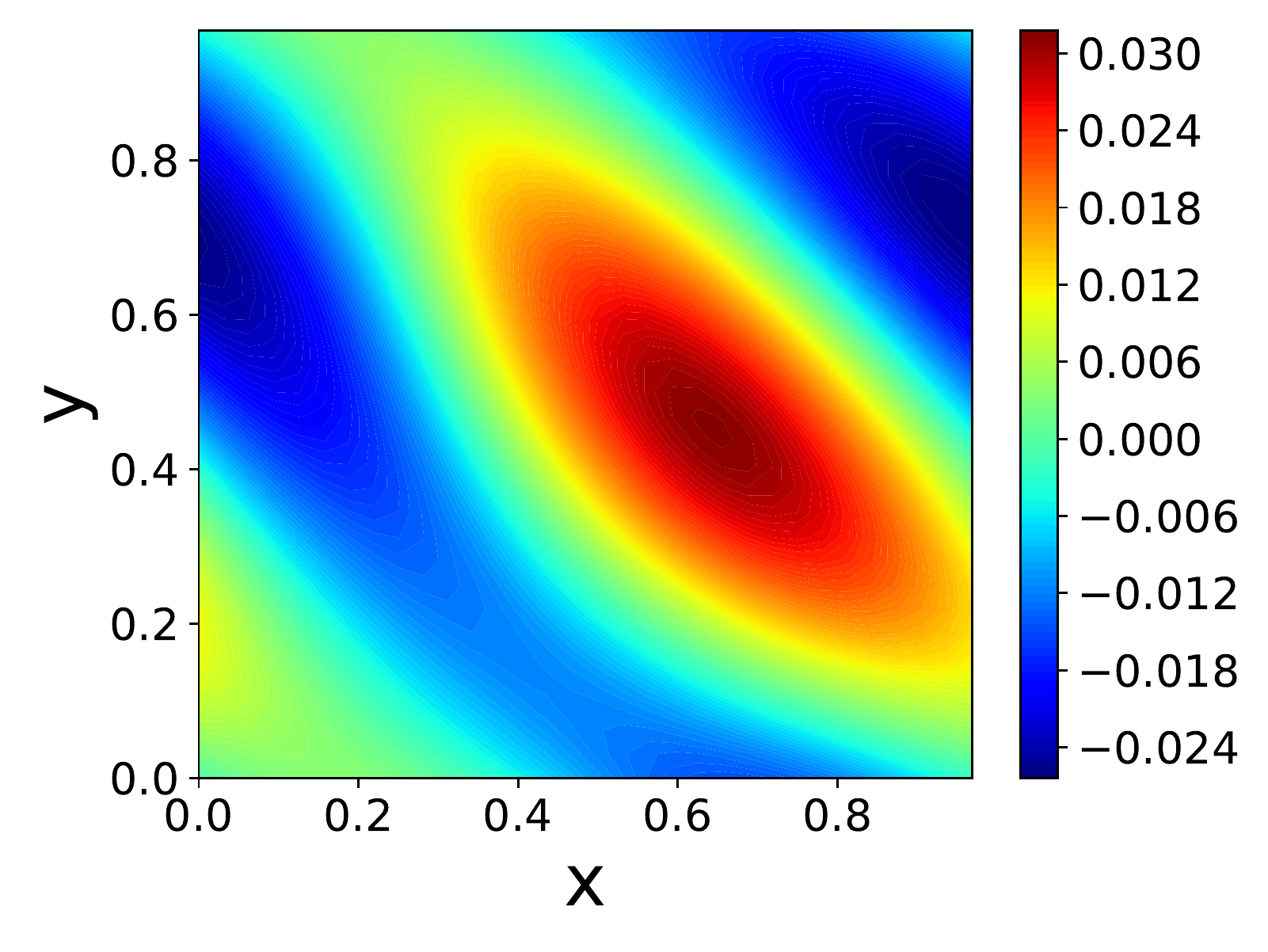}}&
 \raisebox{-.5\height}{\includegraphics[width=0.24\textwidth, height=0.12\textheight]{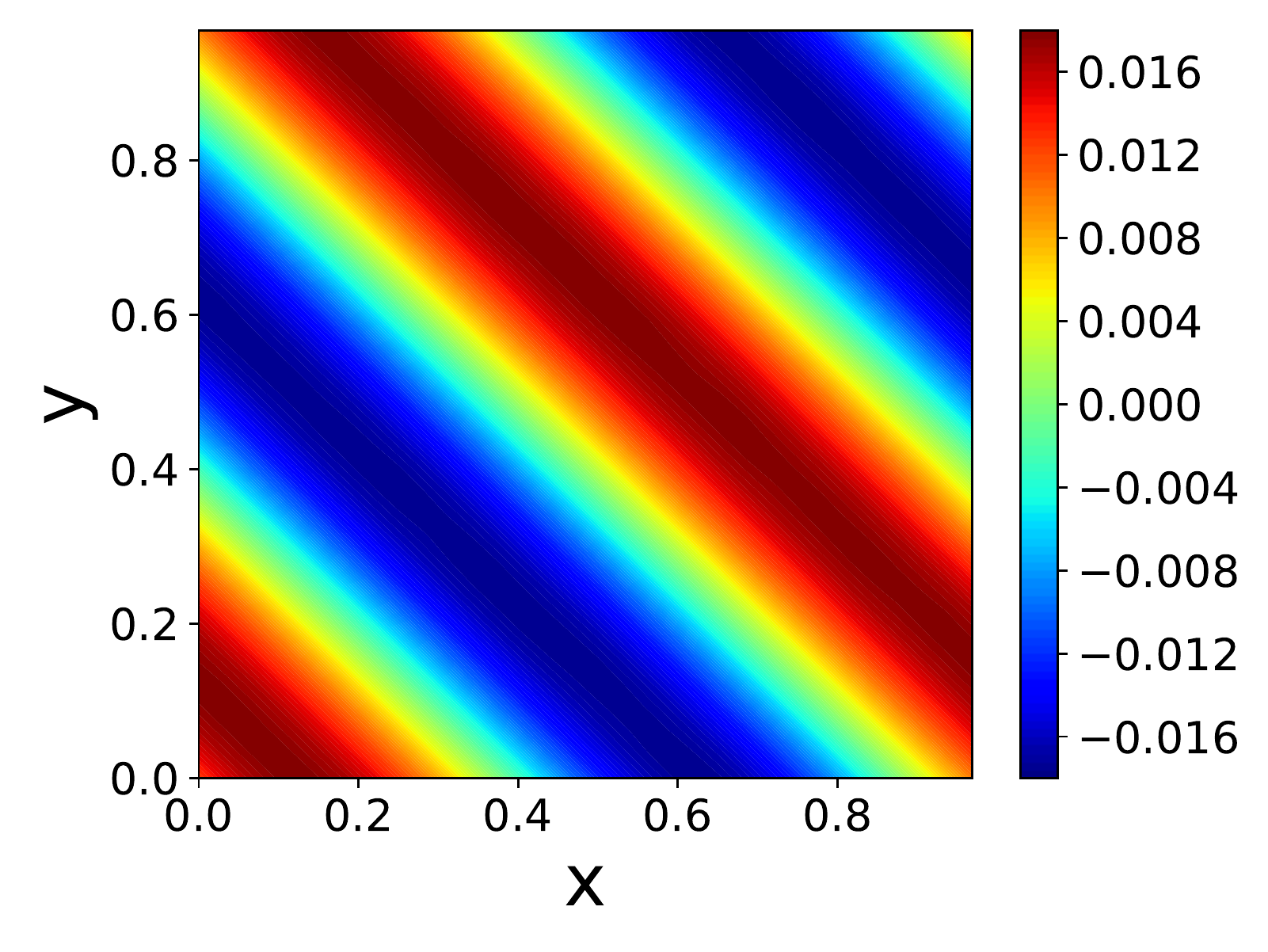}}&
 \raisebox{-.5\height}{\includegraphics[width=0.24\textwidth, height=0.12\textheight]{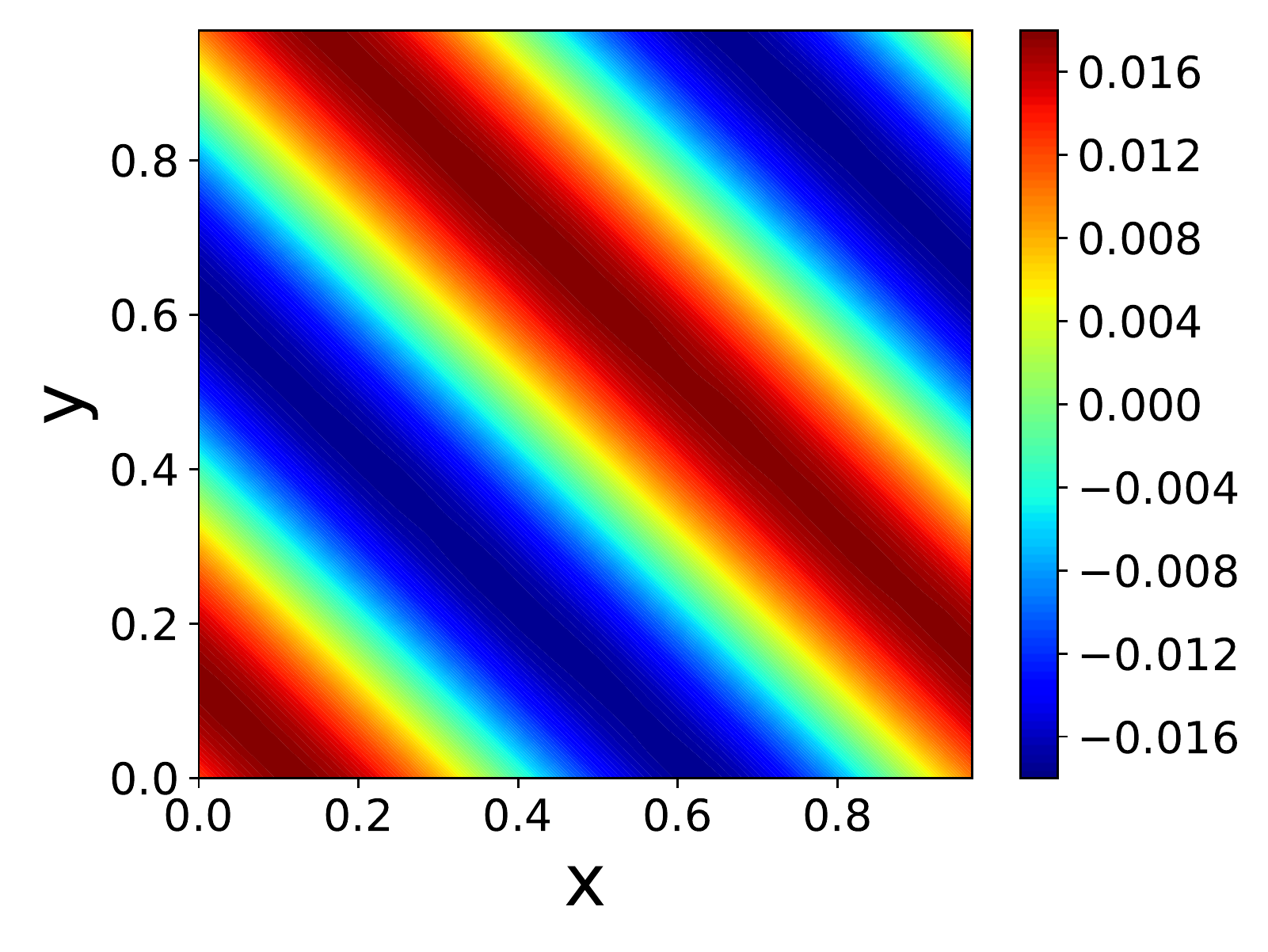}}\\
 \end{tabular}
 \end{center}
 \caption{Results on Navier-Stokes equation with $\nu=0.1$: snapshots of reference solutions (top), and of approximate solutions predicted by DeepONet (middle) and TL-DeepONet (bottom).}
 \label{fig:ns_zp1} 
 \end{figure*}

\begin{figure*}
 \centering
 \begin{center}
 \setlength{\tabcolsep}{-1pt}
 \renewcommand{\arraystretch}{-1}
 \begin{tabular}{cccc}
 \text{Initical condition} \quad & $t=1$ & $t=4$ & $t=10$\\
 \includegraphics[width=0.24\textwidth, height=0.12\textheight]{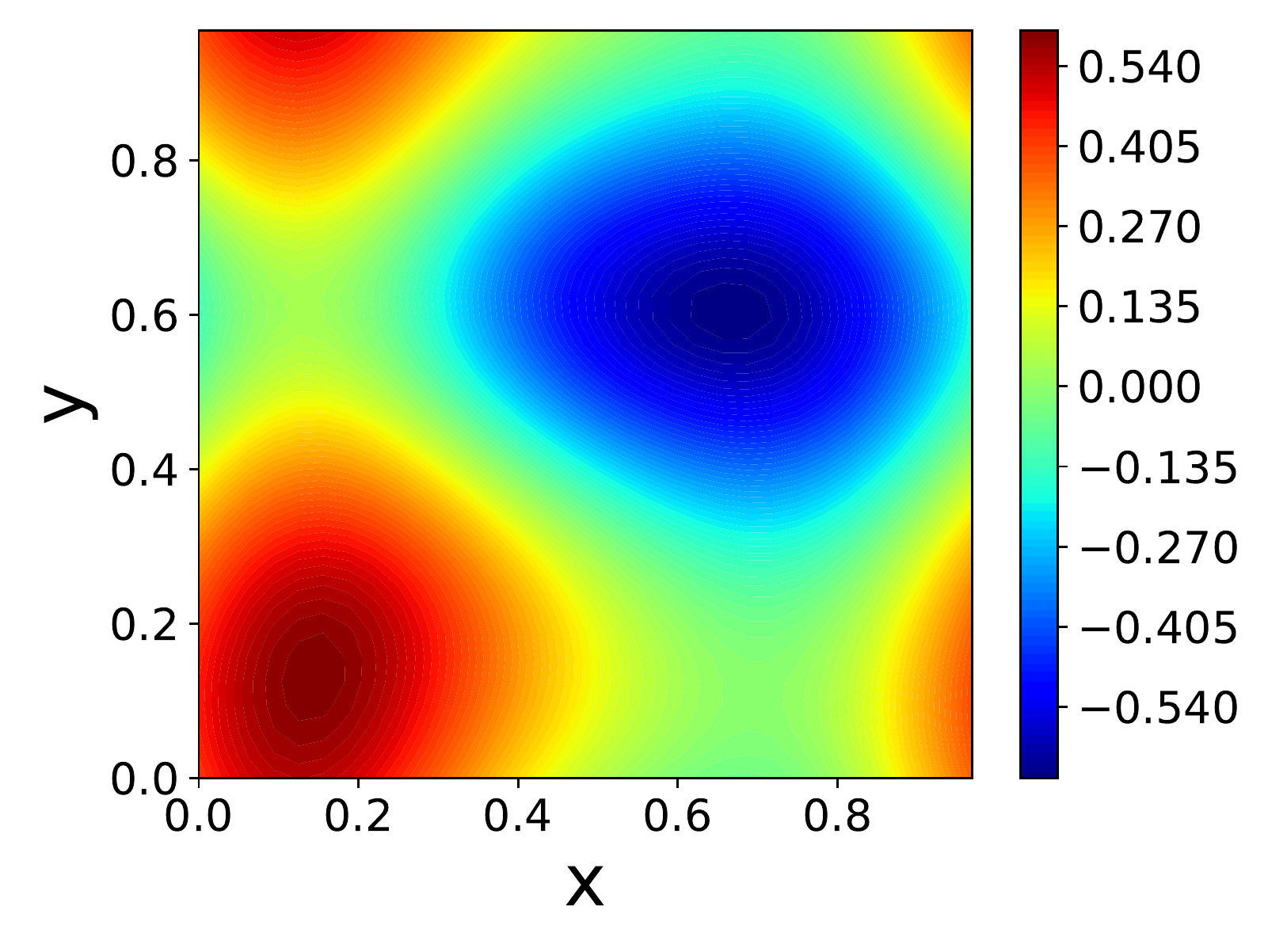}&
 \includegraphics[width=0.24\textwidth, height=0.12\textheight]{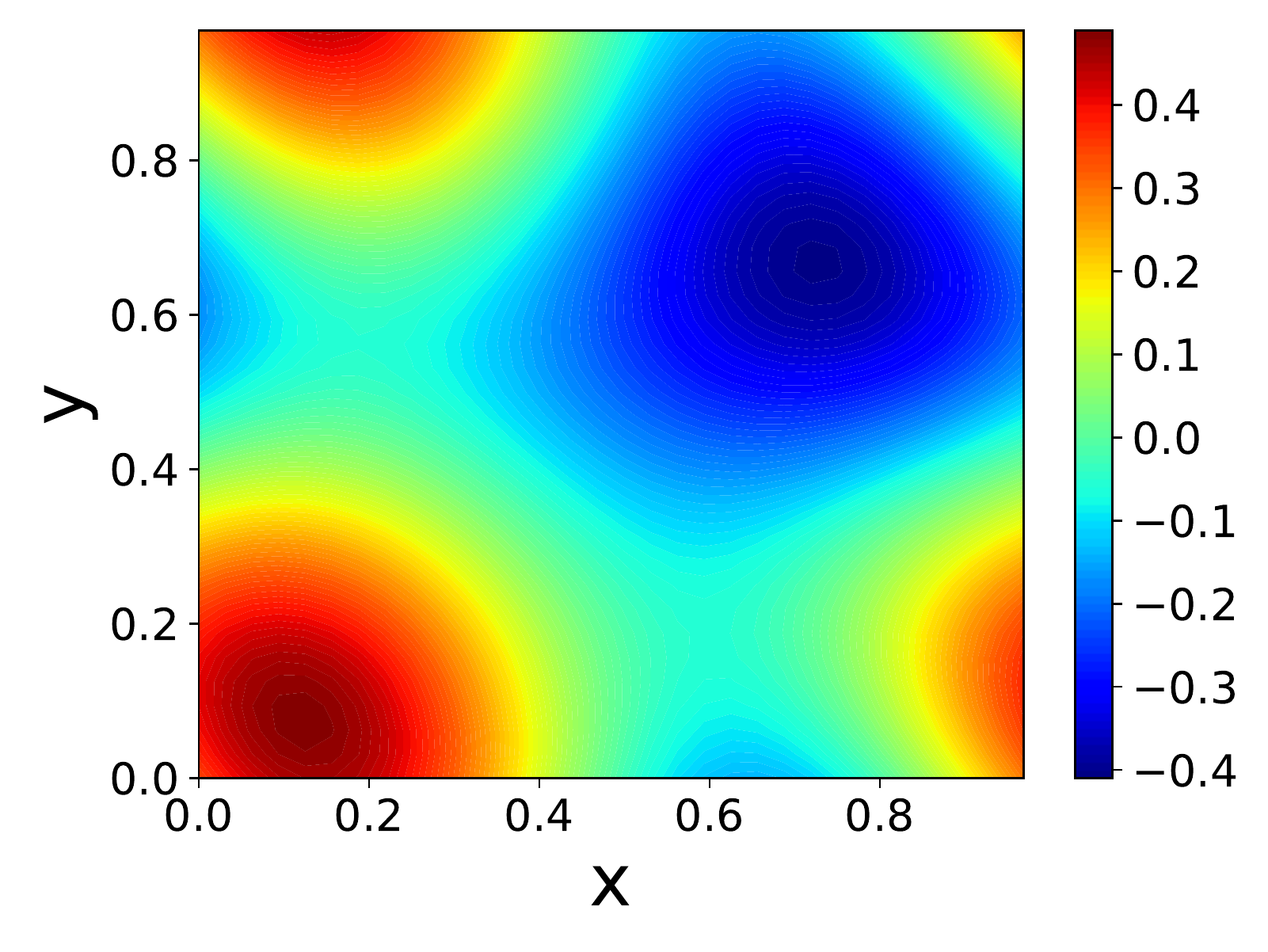}&
 \includegraphics[width=0.24\textwidth, height=0.12\textheight]{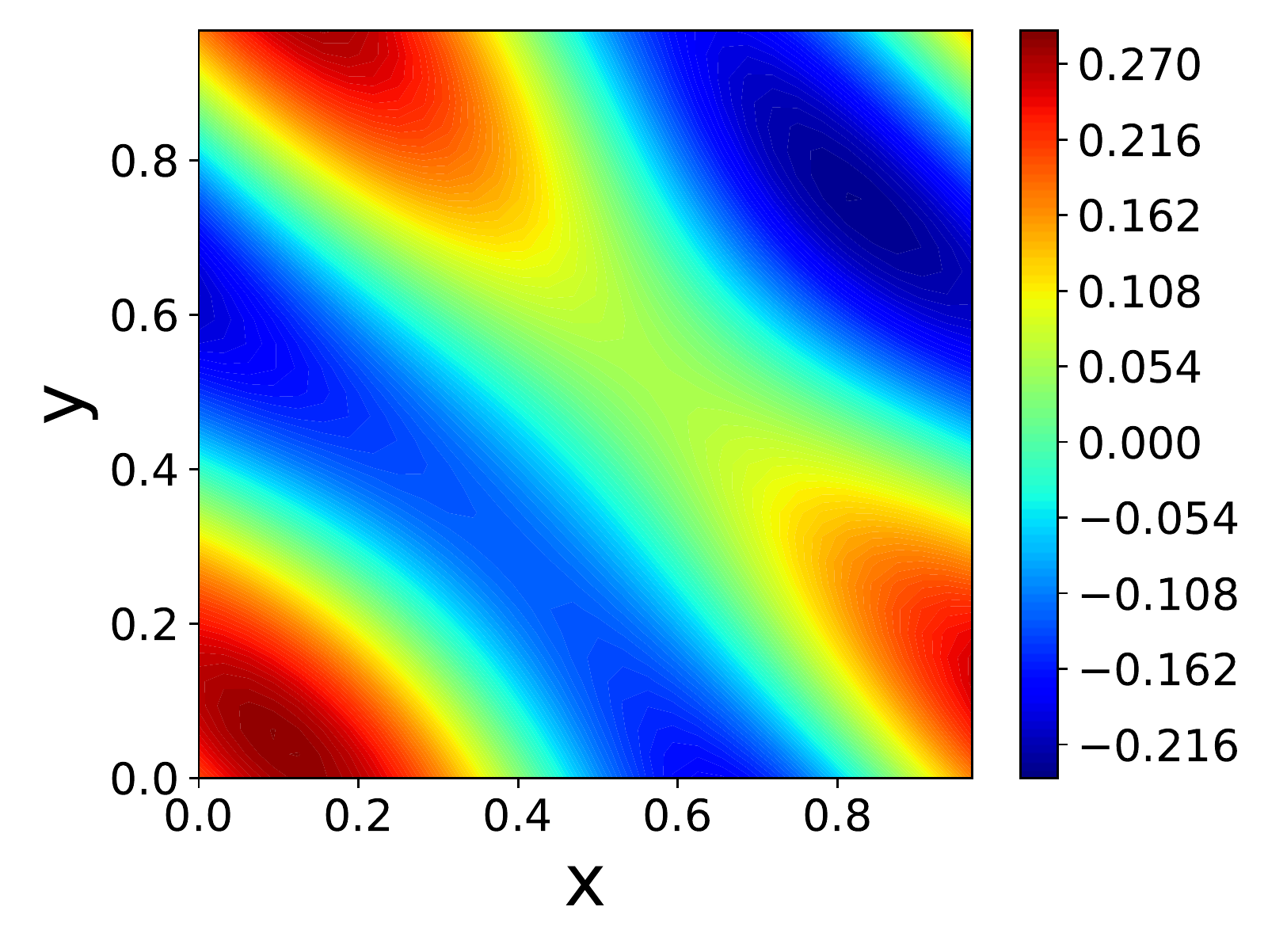}&
 \includegraphics[width=0.24\textwidth, height=0.12\textheight]{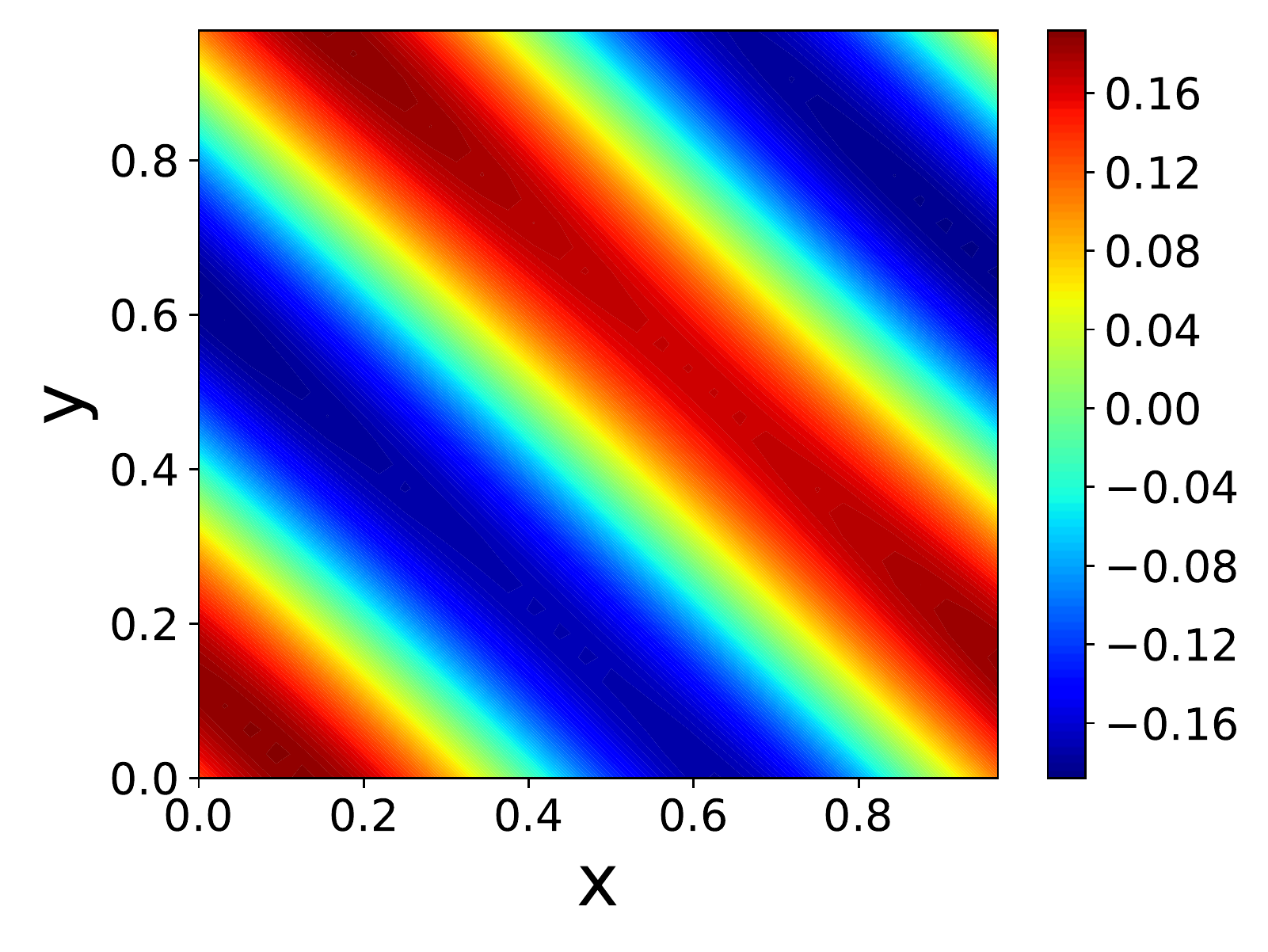}\\
 \text{DeepONet}&
 \raisebox{-.5\height}{\includegraphics[width=0.24\textwidth, height=0.12\textheight]{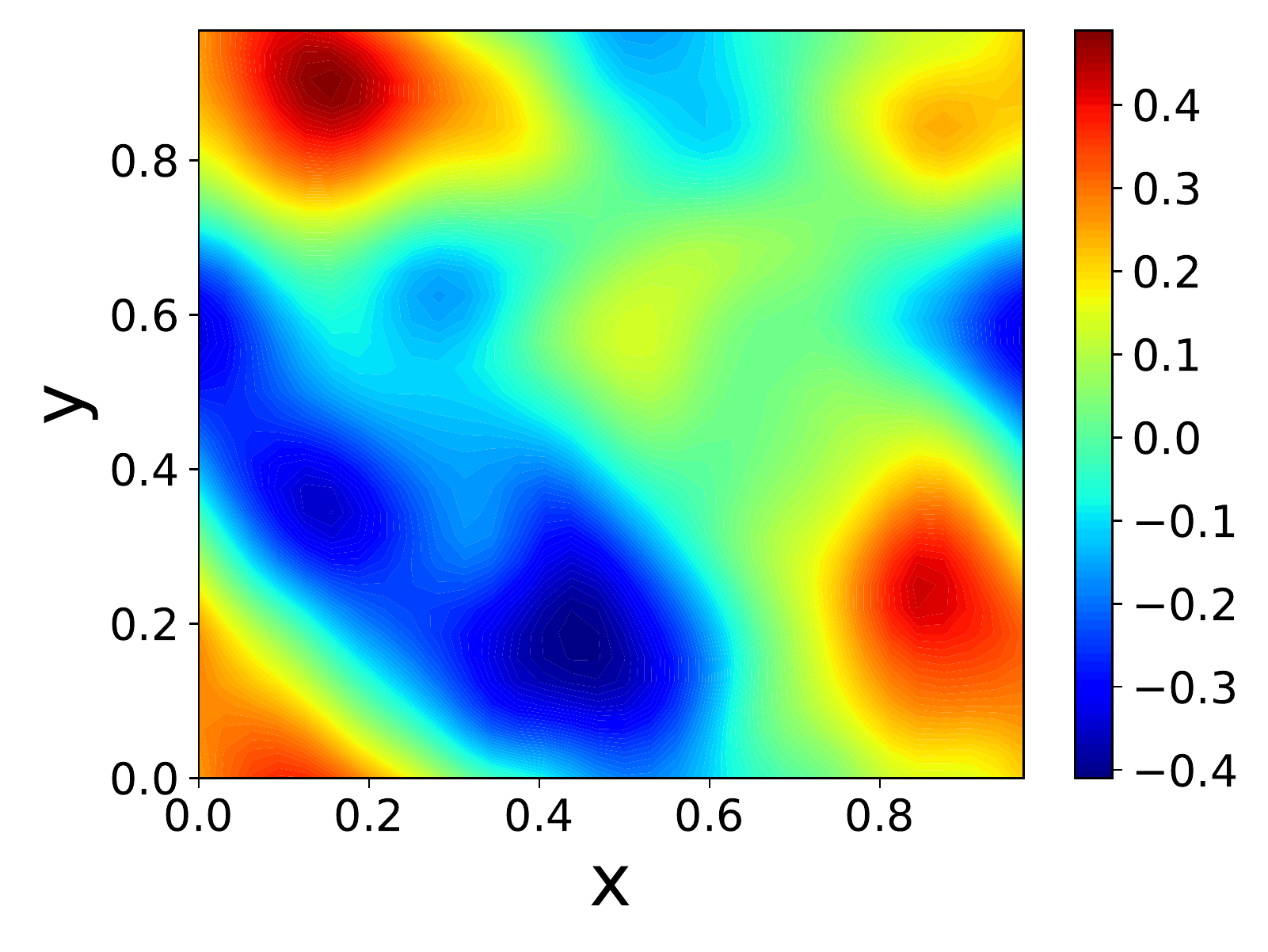}}&
 \raisebox{-.5\height}{\includegraphics[width=0.24\textwidth, height=0.12\textheight]{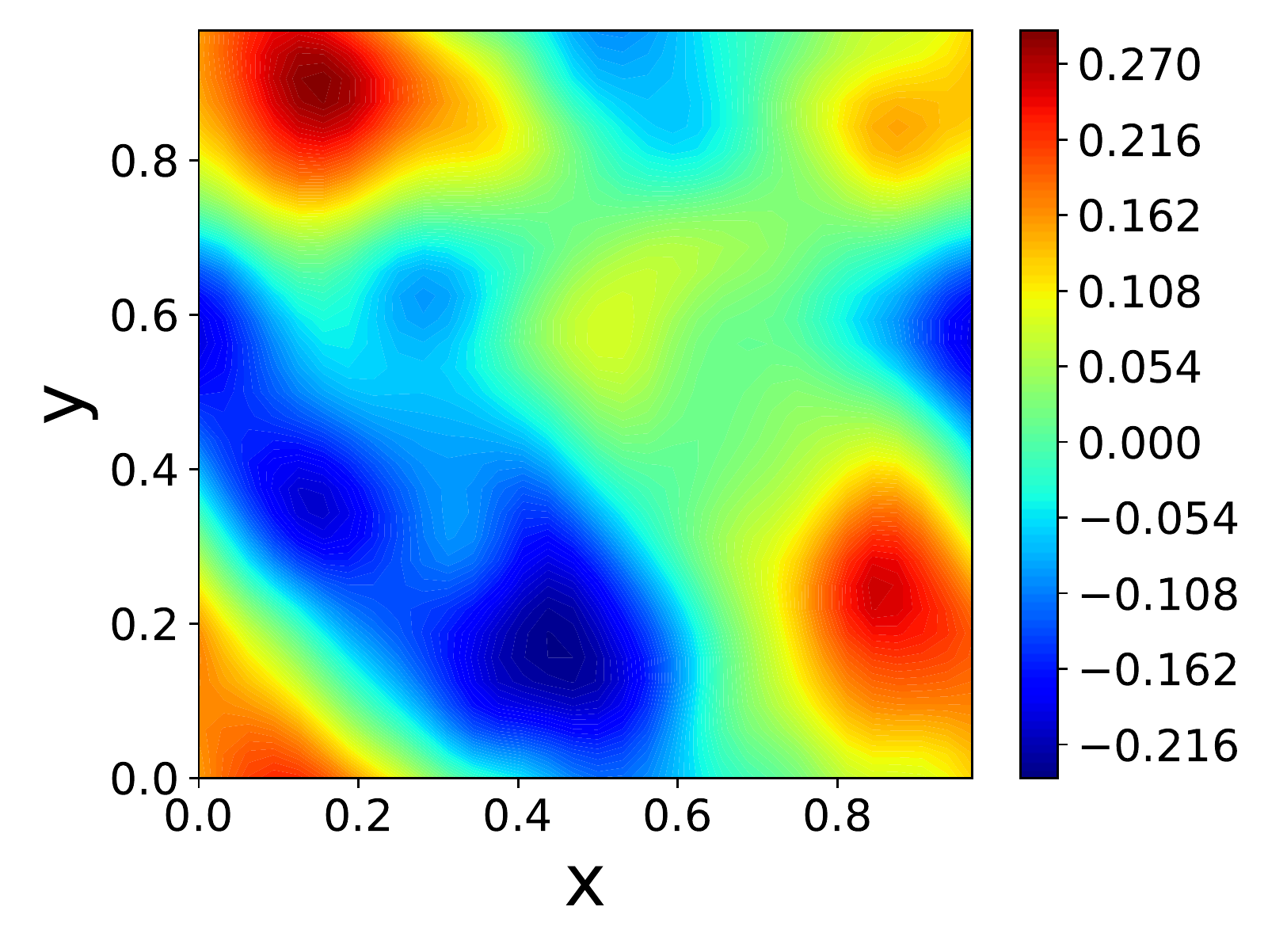}}&
 \raisebox{-.5\height}{\includegraphics[width=0.24\textwidth, height=0.12\textheight]{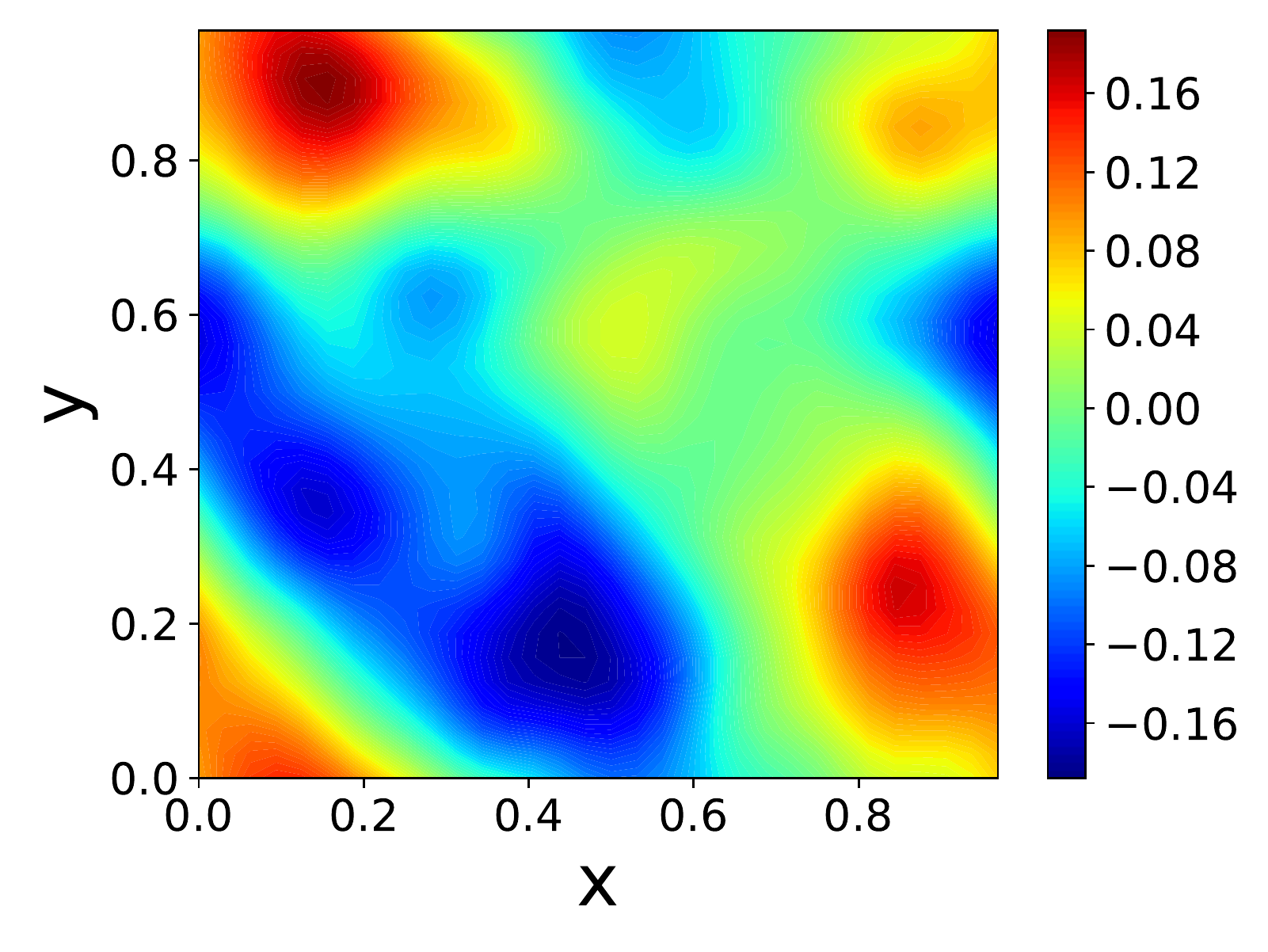}}\\
 \text{TL-DeepONet}&
 \raisebox{-.5\height}{\includegraphics[width=0.24\textwidth, height=0.12\textheight]{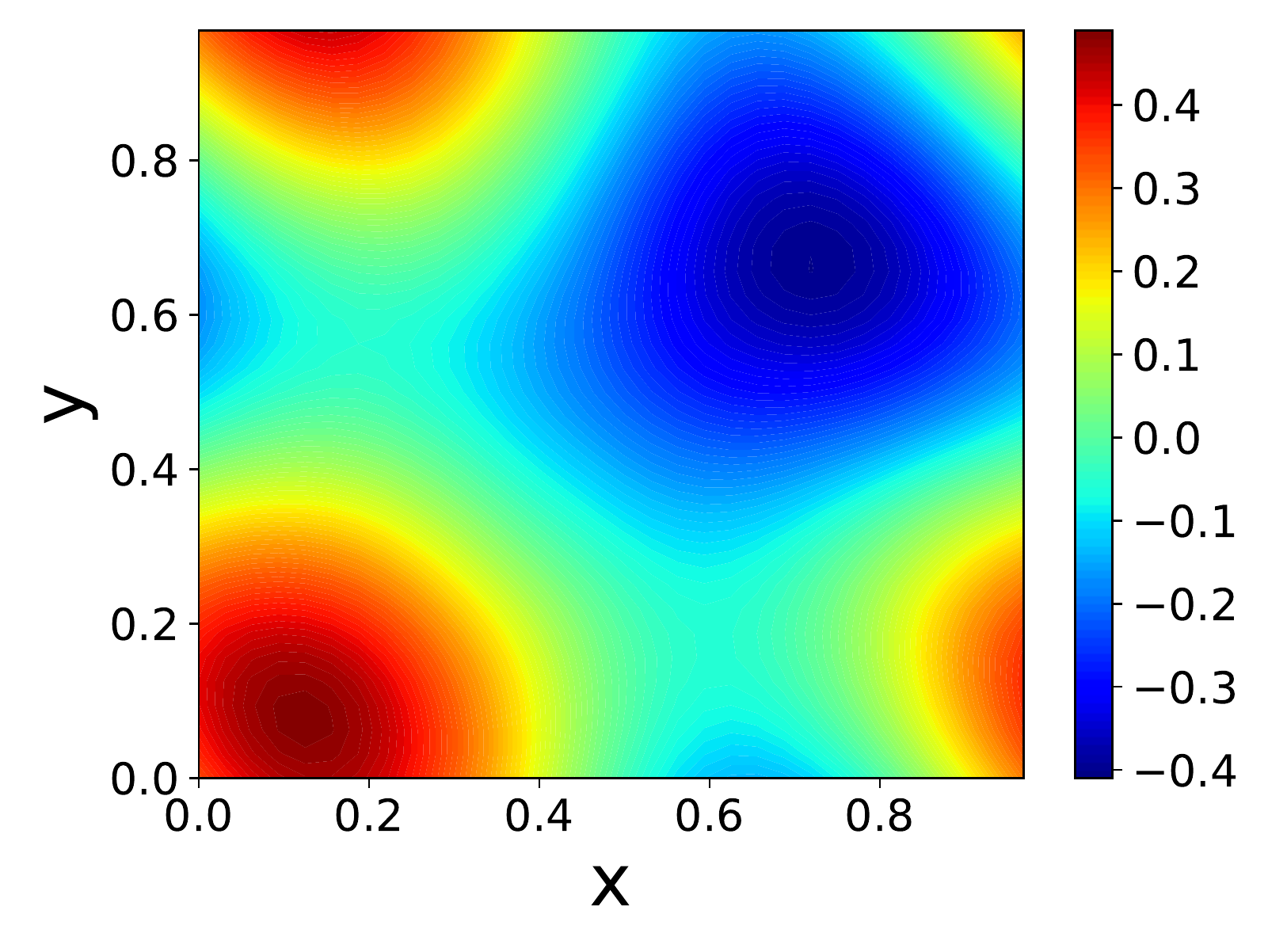}}&
 \raisebox{-.5\height}{\includegraphics[width=0.24\textwidth, height=0.12\textheight]{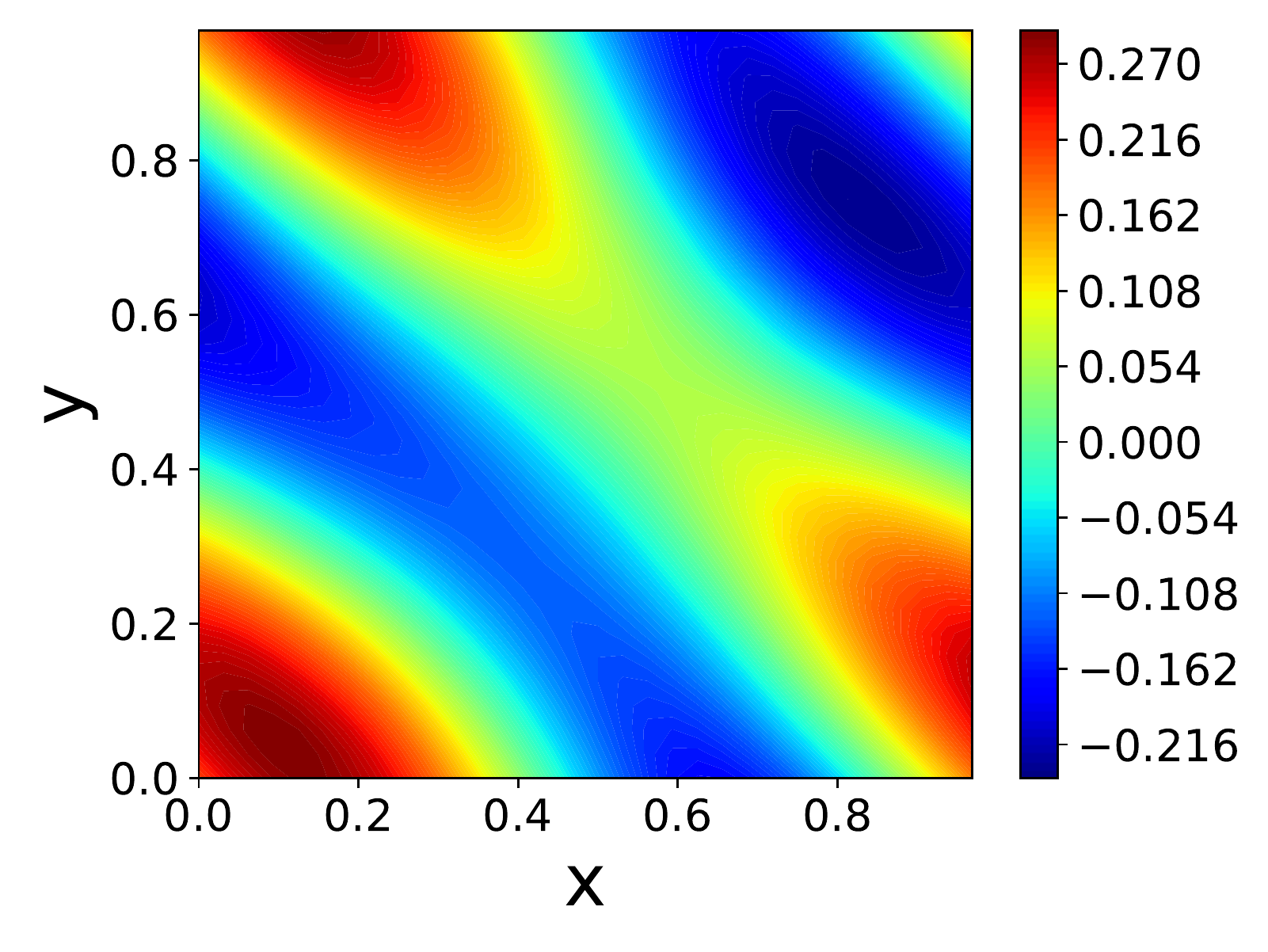}}&
 \raisebox{-.5\height}{\includegraphics[width=0.24\textwidth, height=0.12\textheight]{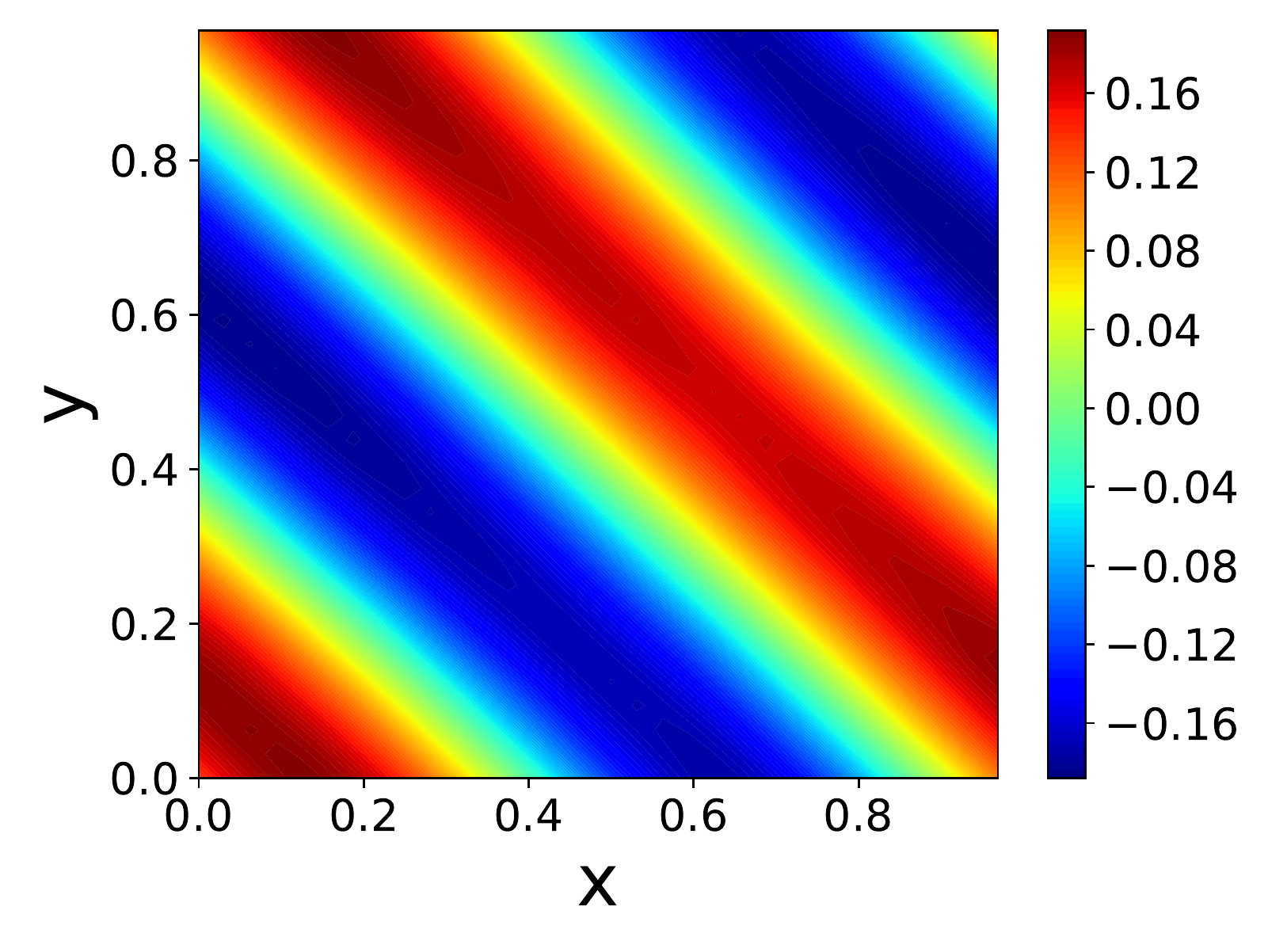}}\\
 \end{tabular}
 \end{center}
 \caption{Results on Navier-Stokes equation with $\nu=0.01$: snapshots of reference solutions (top), and of approximate solutions predicted by DeepONet (middle) and TL-DeepONet (bottom).}
 \label{fig:ns_zpz1} 
 \end{figure*}

\begin{figure*}
 \centering
 \begin{center}
 \setlength{\tabcolsep}{-1pt}
 \renewcommand{\arraystretch}{-1}
 \begin{tabular}{cccc}
 \text{Initical condition} \quad & $t=1$ & $t=4$ & $t=10$\\
 \includegraphics[width=0.24\textwidth, height=0.12\textheight]{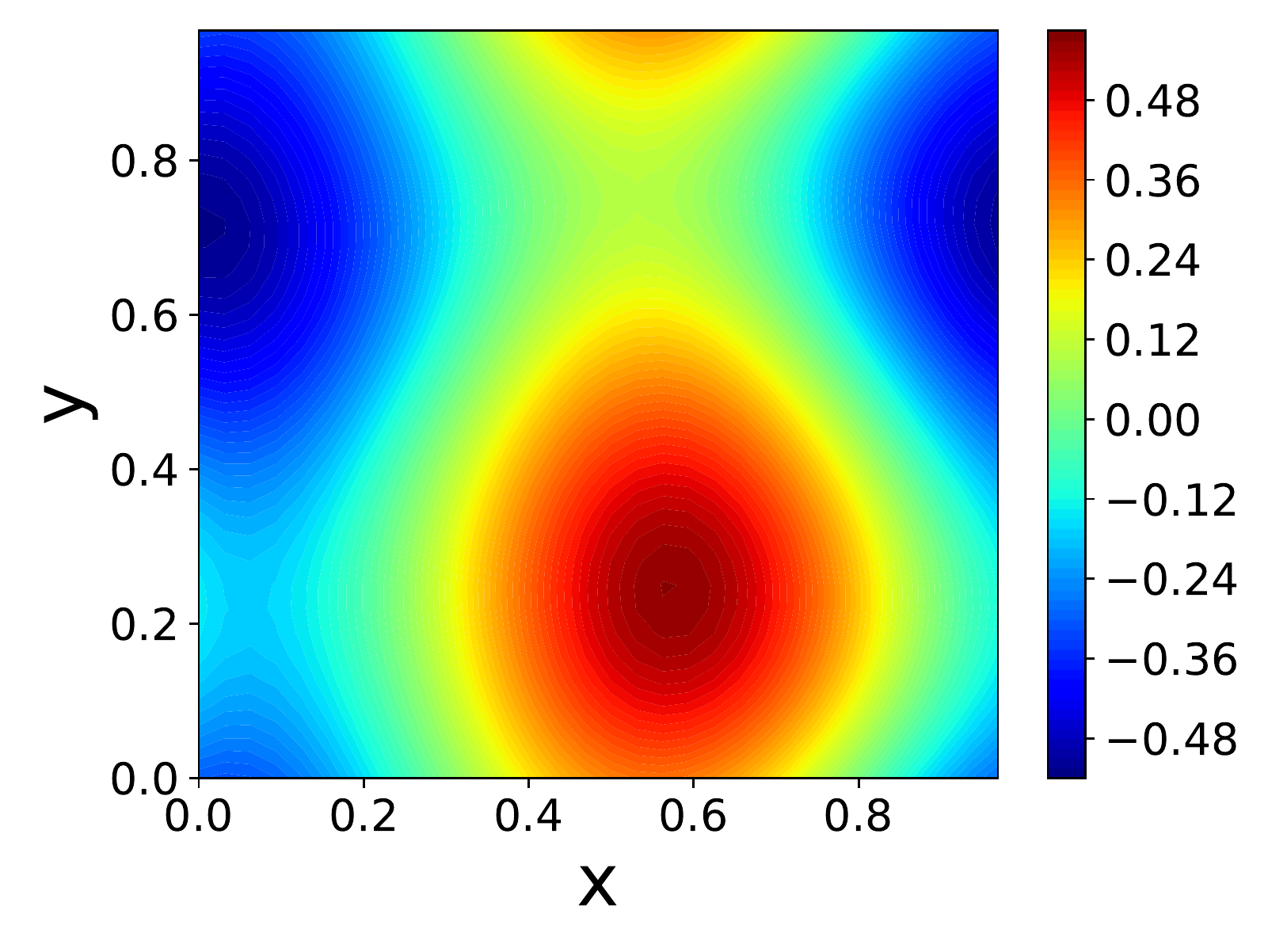}&
 \includegraphics[width=0.24\textwidth, height=0.12\textheight]{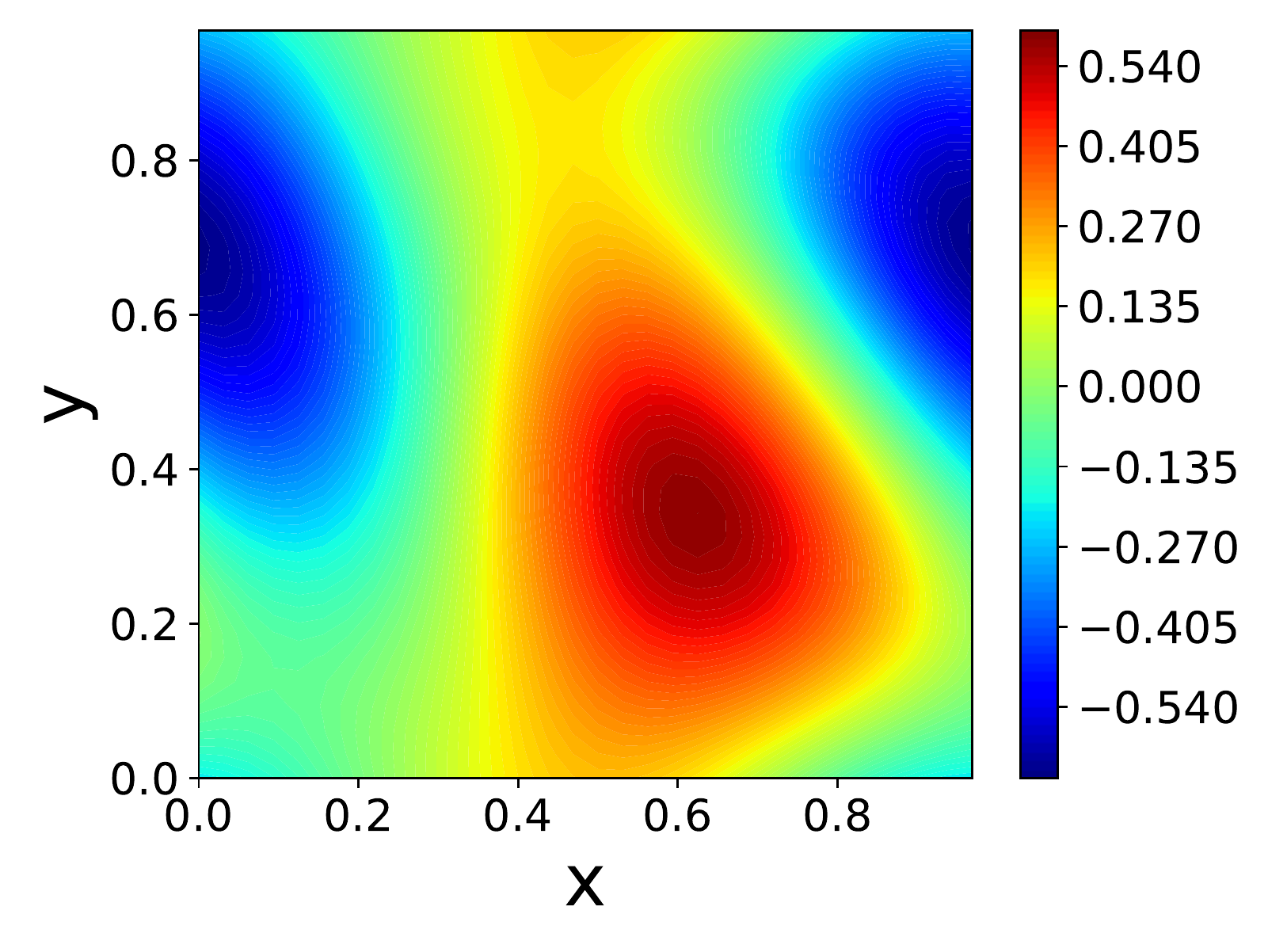}&
 \includegraphics[width=0.24\textwidth, height=0.12\textheight]{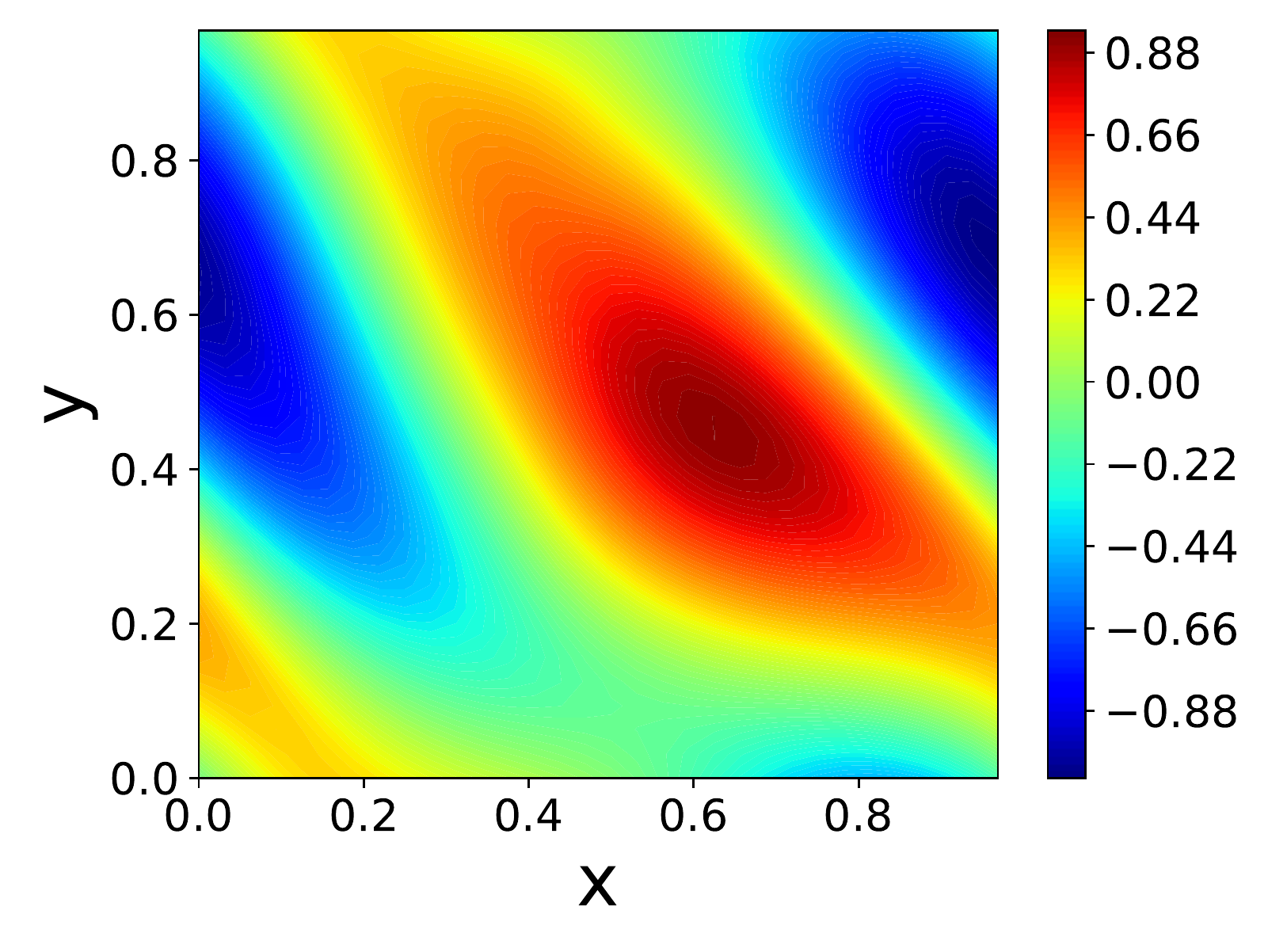}&
 \includegraphics[width=0.24\textwidth, height=0.12\textheight]{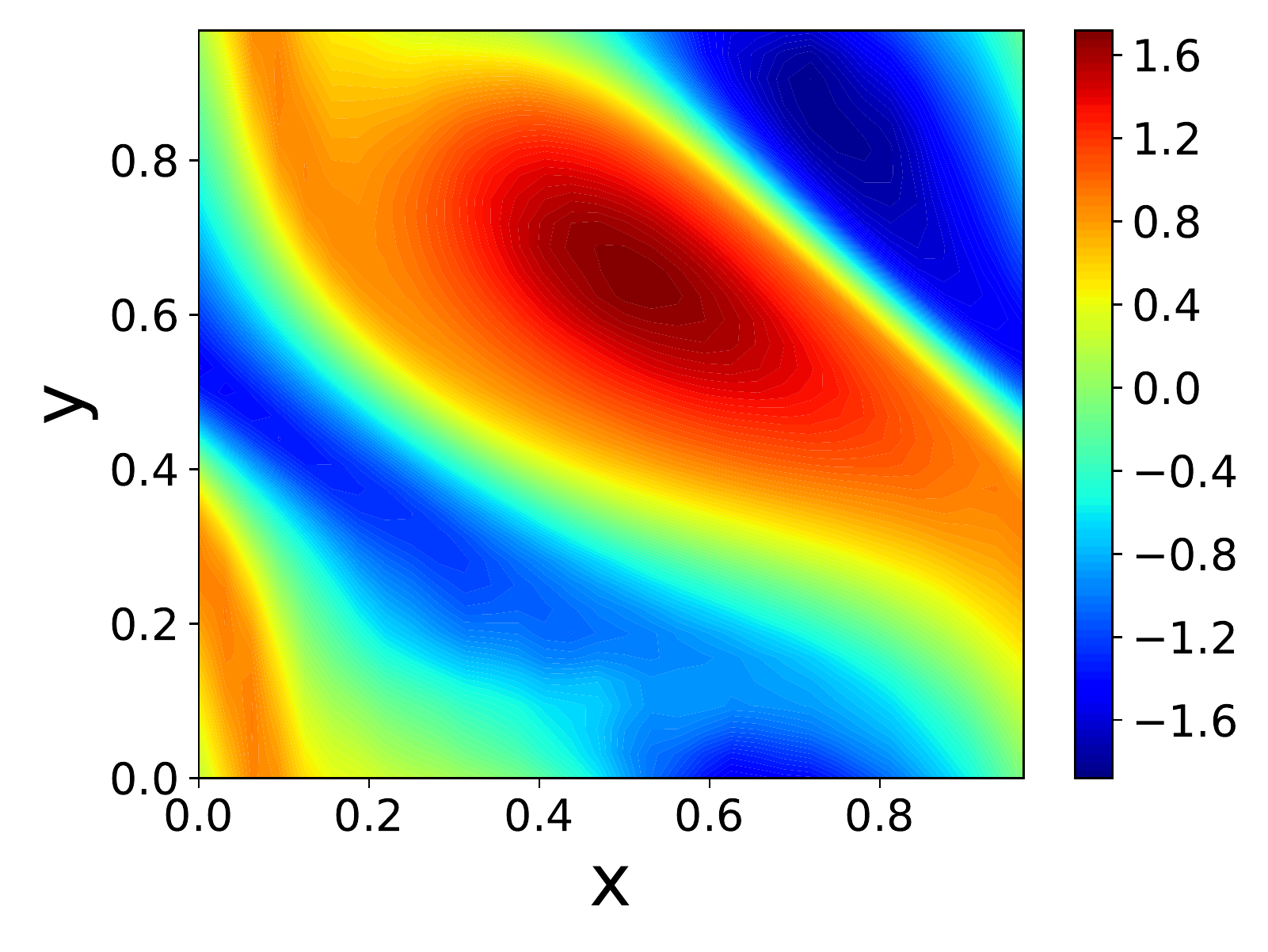}\\
 \text{DeepONet}&
 \raisebox{-.5\height}{\includegraphics[width=0.24\textwidth, height=0.12\textheight]{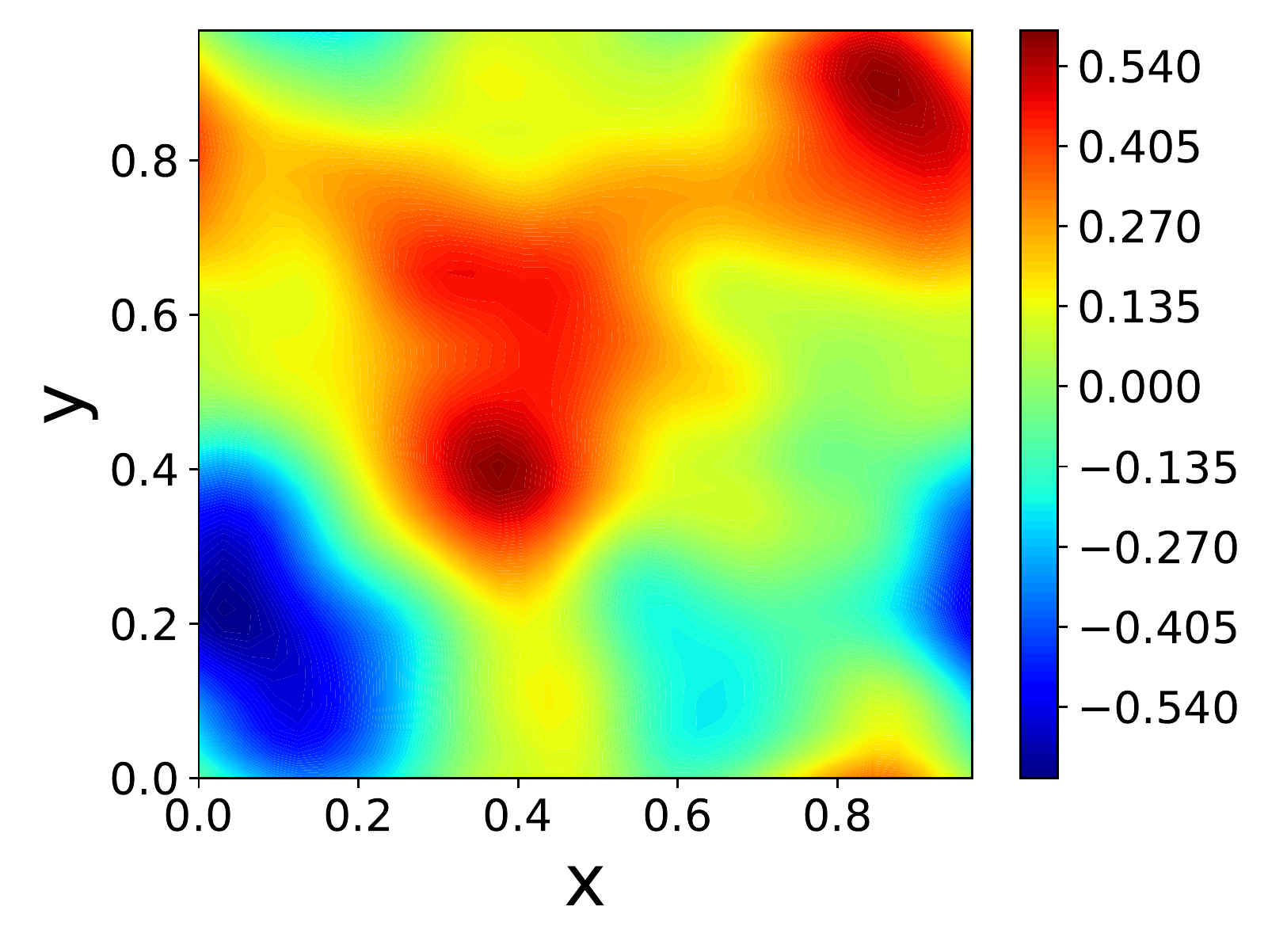}}&
 \raisebox{-.5\height}{\includegraphics[width=0.24\textwidth, height=0.12\textheight]{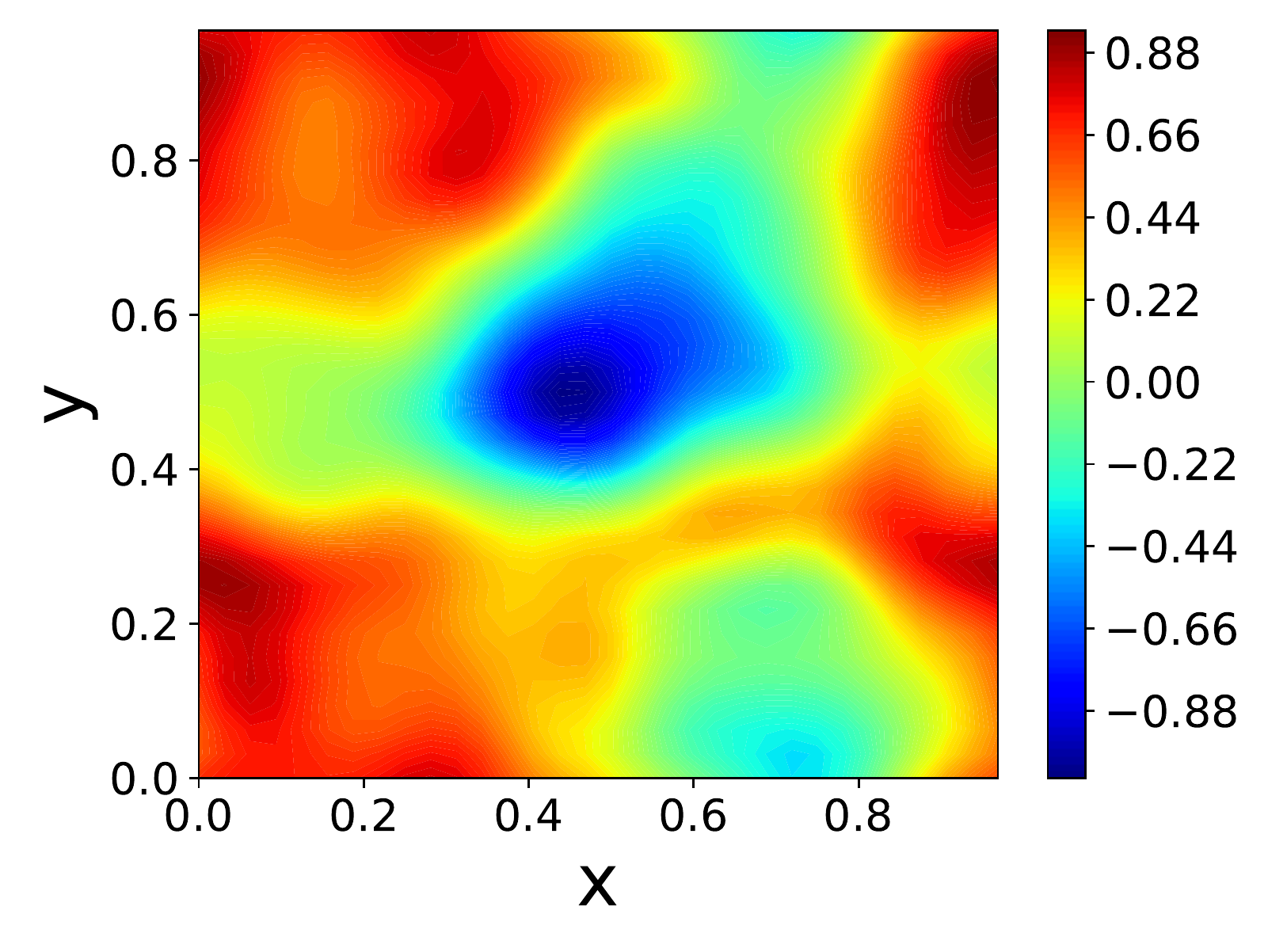}}&
 \raisebox{-.5\height}{\includegraphics[width=0.24\textwidth, height=0.12\textheight]{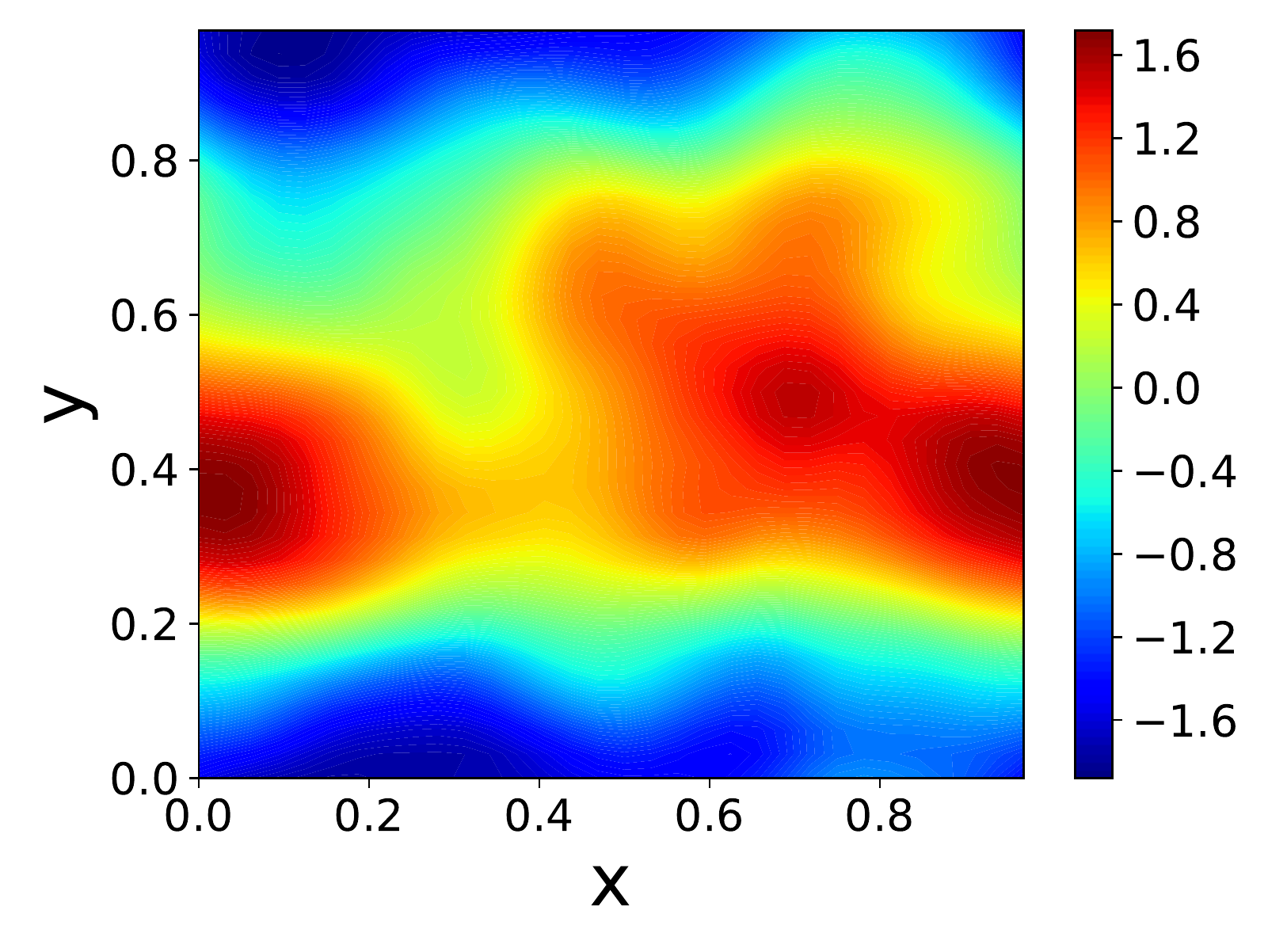}}\\
 \text{TL-DeepONet}&
 \raisebox{-.5\height}{\includegraphics[width=0.24\textwidth, height=0.12\textheight]{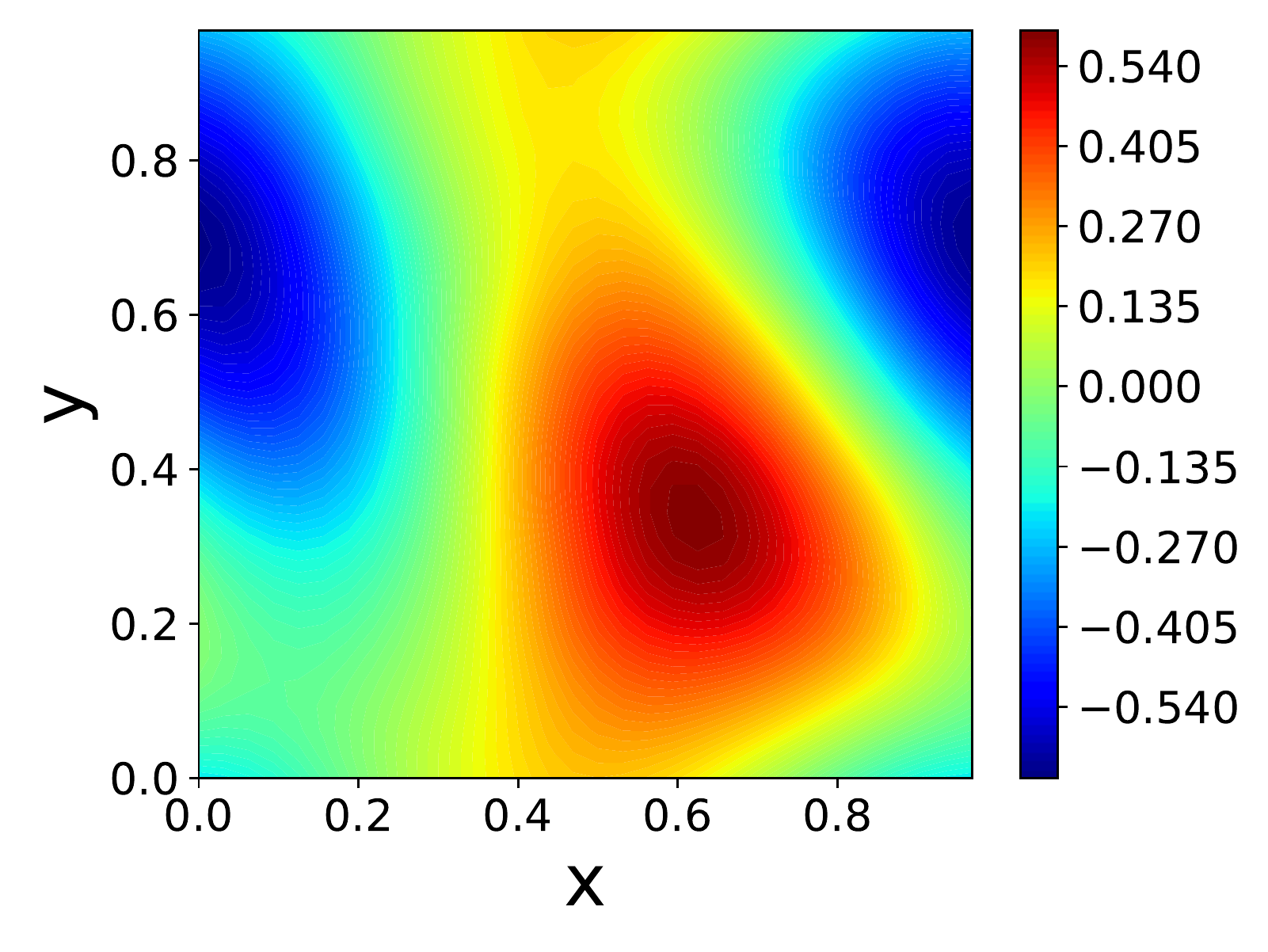}}&
 \raisebox{-.5\height}{\includegraphics[width=0.24\textwidth, height=0.12\textheight]{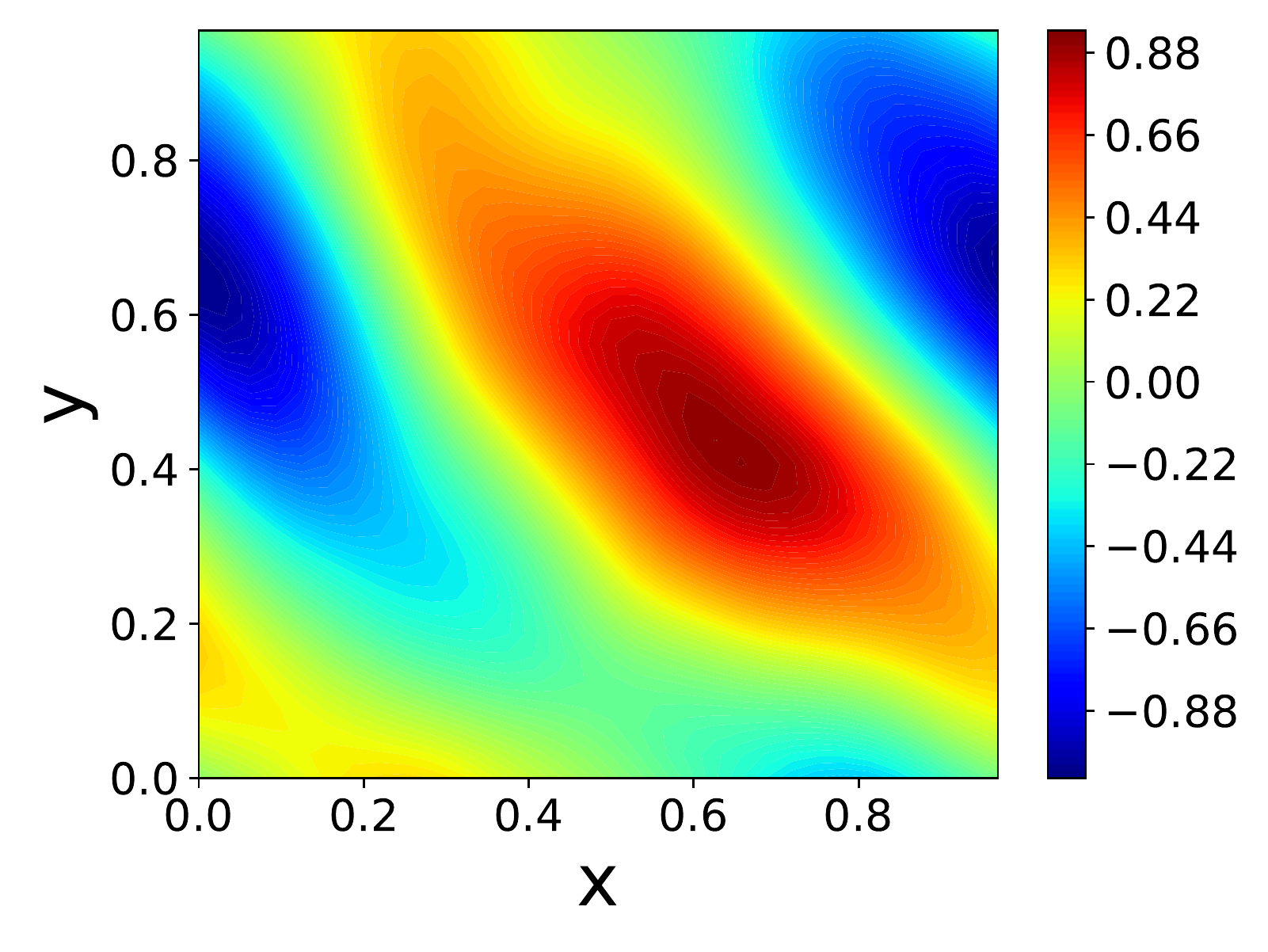}}&
 \raisebox{-.5\height}{\includegraphics[width=0.24\textwidth, height=0.12\textheight]{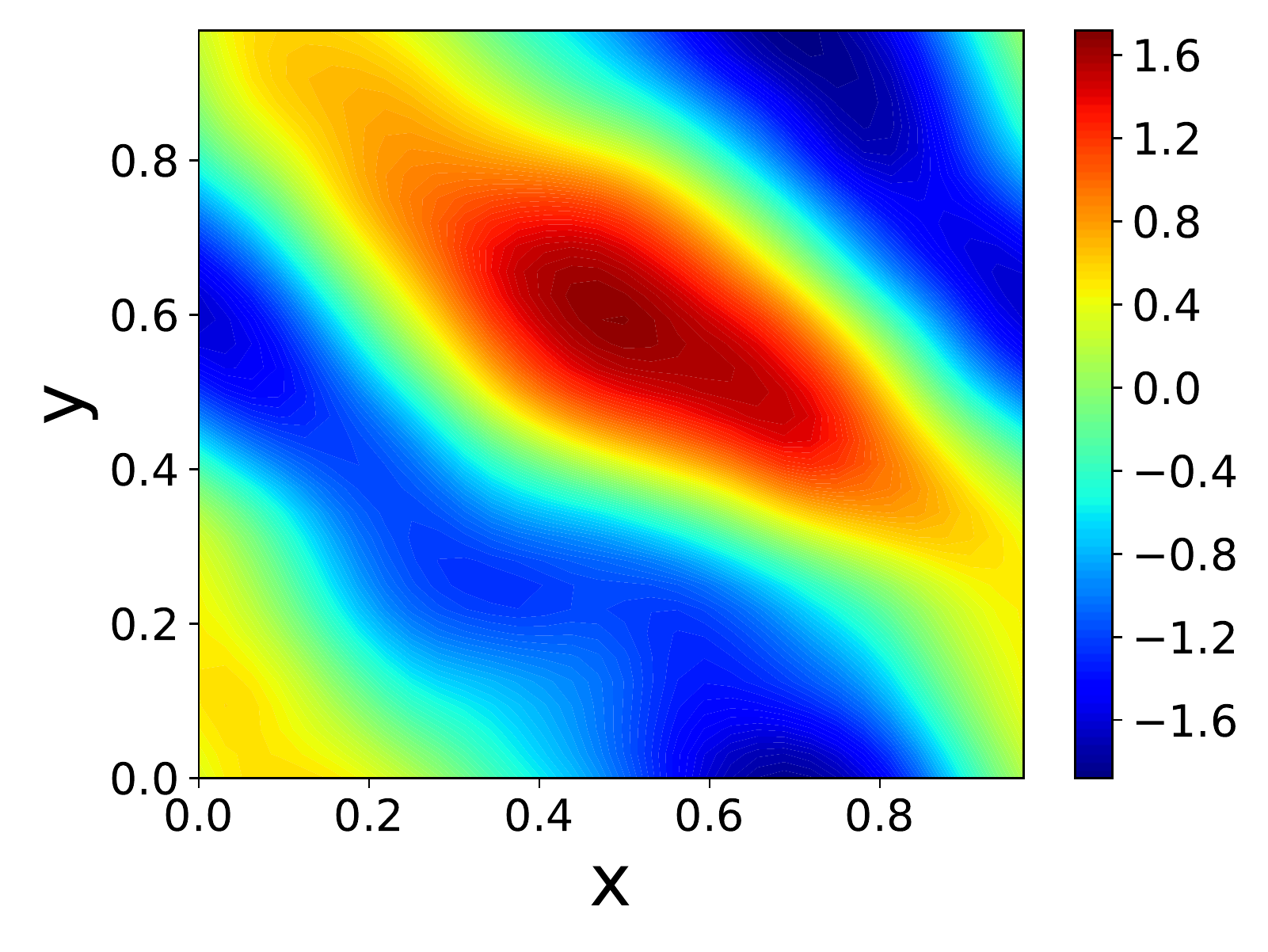}}\\
 \end{tabular}
 \end{center}
 \caption{Results on Navier-Stokes equation with $\nu=0.0001$: snapshots of reference solutions (top), and of approximate solutions predicted by DeepONet (middle) and TL-DeepONet (bottom).}
 \label{fig:ns_zpzzz1} 
 \end{figure*}

 Here we provide the snapshots of solution to \eqref{eqn:ns} with $\nu=0.1, 0.01, 0.0001$ in Figures~\ref{fig:ns_zp1}, \ref{fig:ns_zpz1}, \ref{fig:ns_zpzzz1} respectively.

 \subsubsection{Implementation details on Table~\ref{table:ns} and Figures~\ref{fig:ns_zp1},\ref{fig:ns_zpz1},\ref{fig:ns_zpzz1}, \ref{fig:ns_zpzzz1}}
 Consider 2D Navier-Stokes equation \eqref{eqn:ns}. We use $N_e=30$ test functions drawn from the Gaussian process defined above with the length scale $l=1$ in \eqref{eqn:kp_2d}. We use $\Delta t = 0.01$ for time step size and $N_x = N_y = 20$ uniform spatial grids for spatial discretization. We set the maximum iteration number $N_{iter} = 200000$, adopt the stopping criterion that the empirical loss is below 1e-6 and use number of feature $p=120$. In the transfer learning step, we subsample $N_c = 144 < 20 \times 20$ and update $q = 40$ weights defined in \eqref{eqn:approx_trans}.

 \subsection{Details on Multiscale linear radiative transfer equation}\label{sec:detailsrte}

 \subsubsection{Computation of multiscale radiative transfer equation}
 Following \cite{lu2022solving}, instead of discretizing the original radiative transfer equation \eqref{eq:rte}, we consider a new system of equations based on its micro-macro decomposition. More concretely, let $f = \rho + \eps g, ~\rho = \average{f}$. We consider 
 \begin{equation}\label{eq:decomp}
 \left\{ \begin{array}{l}
 \partial_t \rho + \average{\bv \cdot \nabla g} = 0 \\
 \eps^2 g_t + \eps \bv \cdot \nabla g - \eps \average{\bv \cdot \nabla g} + \bv \cdot \nabla \rho =  \Lop g,  \\
 \rho(t, \bx) + \eps g(t, \bx, \bv) = \phi(\bx) , ~ (\bx, \bv) \in  \Gamma_-\\
 f(0, \bx, \bv) = f_0(\bx, \bv) .
 \end{array}\right.
 \end{equation}
 It was shown in \cite{lu2022solving} that the PINN loss based on \eqref{eq:rte} suffers from the instability issue when  the small Knudsen number $\eps$ is small while  the PINN loss based on the system \eqref{eq:decomp} above  is uniformly stable with respect to the small Knudsen number  in the sense that the $L^2$-error of the neural network solution is uniformly controlled by the loss. We consider \eqref{eq:rte} and its equivalent system  \eqref{eq:decomp} in one and two dimensions. To enforce the inflow boundary condition 
 \[
 f(t, \bx, \bv) = \phi(\bx) , ~ (\bx, \bv) \in  \Gamma_-\,,
 \]
 we first parameterize $f$ as follows: 
 \begin{equation}\label{eqn:f_decomp_abc}
 f(t,\bx,\bv) = \mathcal{N}_1(t,\bx)A(\bx) + C(\bx) + \eps \mathcal{N}_2(t, \bx, \bv)B(\bx,\bv)\,,
 \end{equation}
 where $A(\bx)$, $B(\bx,\bv)$ and $C(\bx)$ are determined according to the  specific boundary conditions in each example, whilst $\mathcal{N}_1(t, \bx)$ and $\mathcal{N}_2(t, \bx, \bv)$ are to-be approximated by the neural nets. In other words, instead of using two neural nets to approximate  $\rho(t,\bx) = \average{f}$ and $g(t,\bx, \bv)= f - \rho$, we approximate $\mathcal{N}_1$ and $\mathcal{N}_2$, and recover $\rho$ and $g$ via 
 \begin{equation*} 
 \rho(t, \bx) = \mathcal{N}_1(t,\bx)A(\bx) + C(\bx) + \eps \average{\mathcal{N}_2(t, \bx, \bv)B(\bx,\bv)},
 \end{equation*}
 and 
 \begin{equation*} 
 g(t, \bx, \bv) = \mathcal{N}_2(t, \bx, \bv)B(\bx,\bv) - \average{\mathcal{N}_2(t, \bx, \bv)B(\bx,\bv)}\,.
 \end{equation*}

 \subsubsection{1D Example}
 Here the computational domain is $(x,v) \in [0,1] \times [-1, 1]$ and the inflow boundary condition takes the form:
 \[
 f(0,v>0) = 1, ~ f(1, v<0)=\frac{1}{2}\,.
 \]
 Then in \eqref{eqn:f_decomp_abc} we use 
 \begin{align*}
 A(x) &= x(1-x)\,,
 \\ B(x,v) &= R(v)x + R(-v)(1-x) \,,
 \\ C(x) &= (1-\frac{1}{2}x)\,,
 \end{align*}
 where $R(v)$ is the ReLU function. In the same vein, we generate the initial data by first sampling $a(x,v) \sim \mathcal{GP} (0, K_l(x_1, v_1),(x_2, v_2))$ with
 \begin{equation}\label{eqn:kl_xv_1d}
 K_l((x_1, v_1),(x_2, v_2)) = e^{-\frac{(x_1-x_2)^2 + (v_1-v_2)^2}{2l^2}}\,,
 \end{equation}
 and then selecting those that are strictly positive to construct 
 \[
 f_0(x,v) = a(x,v)B(x,v) + C(x)\,.
 \]

 \subsubsection{2D Example} Consider the spatial domain as $(x, y) \in [0, 1]^2$, and velocity variable is $(v_x, v_y) = (\cos \alpha, \sin \alpha)$ with $\alpha \in [0, 2 \pi]$. The inflow boundary condition reads:
 \begin{align*}
 &f(t, 0, y, v_x >0) = \frac{1}{2} - (y - \frac{1}{2})^2, \\ & f(t, 1, y, v_x <0) = f(t, x, 0, v_y >0) = f(t, x, 1, v_y <0) = \frac{1}{2}.
 \end{align*}
 Then to guarantee that $f$ in \eqref{eqn:f_decomp_abc} satisfies this condition, we choose 
 \begin{align*}
 A(x,y) &= x(1-x)y(1-y)\,,
 \\  B(x,y,v_x, v_y) & = R(v_x)xy(1-y) + R(-v_x)(1-x)y(1-y)\nonumber \\ &+ R(v_y)yx(1-x) + R(-v_y)(1-y)x(1-x)\nonumber \\
 & + R(v_x)R(v_y)xy + R(v_x)R(-v_y)x(1-y) \nonumber \\ & + R(-v_x)R(v_y)(1-x)y \nonumber \\ &+ R(-v_x)R(-v_y)(1-x)(1-y)\,,\nonumber
 \\ C(x,y) &=(\frac{1}{4} - (y - \frac{1}{2})^2)(1-x) + \frac{1}{4}.
 \end{align*}
 Similarly, the initial conditions are generated by first sampling  $a(x,y,\alpha) \sim \mathcal{GP} (0, K_l(x_1, y_1, \alpha_1),(x_2, y_2, \alpha_2))$ with
 \begin{equation}\label{eqn:kl_xv_2d}
 K_l((x_1, y_1, \alpha_1),(x_2, y_2, \alpha_2)) = e^{-\frac{(x_1-x_2)^2 + (y_1-y_2)^2 + (\alpha_1-\alpha_2)^2}{2l^2}}.
 \end{equation}
 and then retaining only the positive samples to construct $f_0$ via:
 \[
 f_0(x,y,\alpha) = a(x,y,\alpha)B(x,y,\cos(\alpha), \sin(\alpha)) + C(x, y).
 \]

 \subsubsection{Supplementary examples of radiative transfer equation}
\begin{figure*}
 \centering
 \setlength{\tabcolsep}{0pt}
 \renewcommand{\arraystretch}{-1}
 \begin{tabular}{cccc}
 \text{Initical condition} \quad & $t=0.1$ & $t=0.5$ & $t=10$\\
 \raisebox{-.5\height}{\includegraphics[width=0.24\textwidth, height=0.12\textheight]{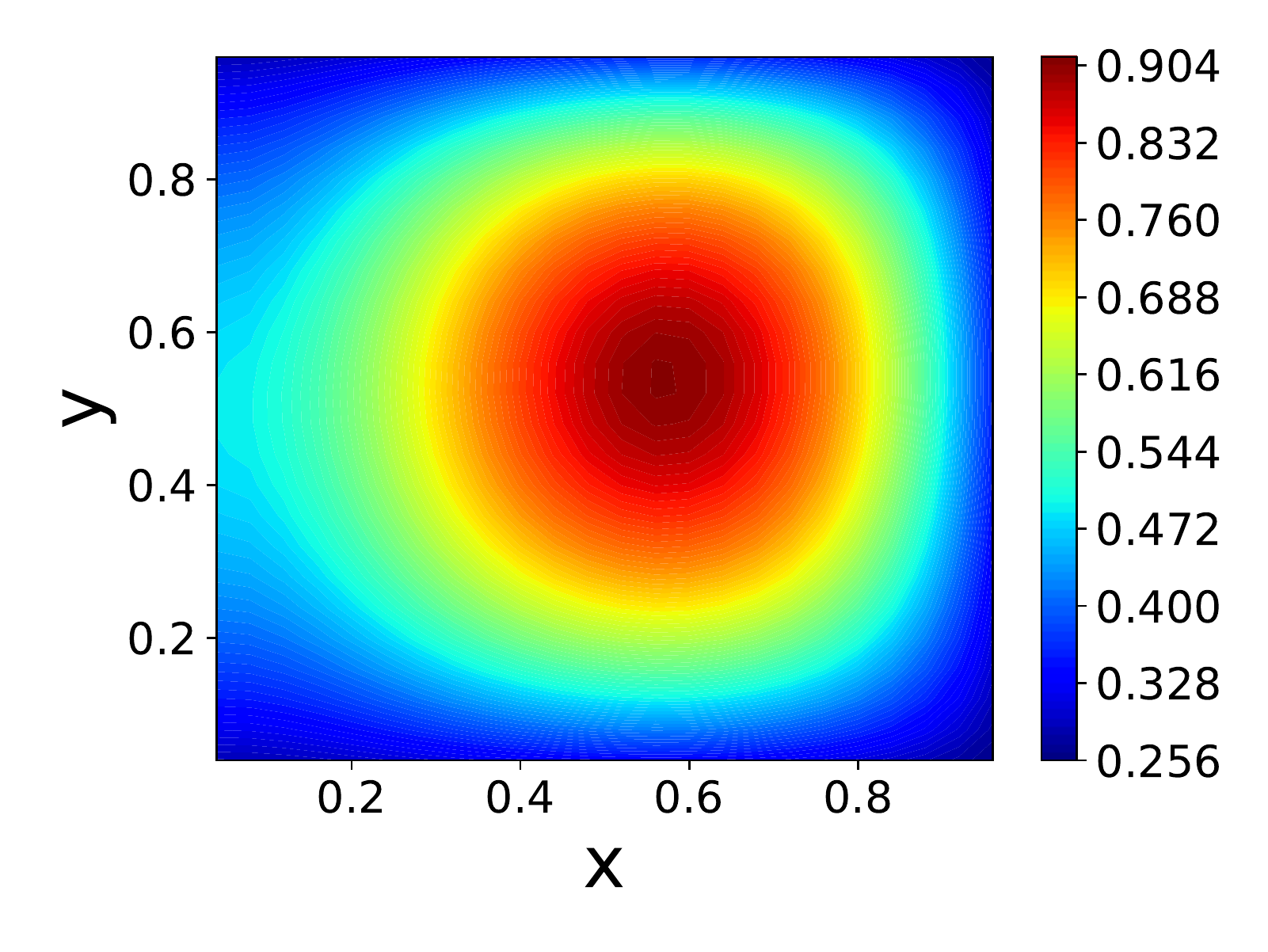}}&
 \raisebox{-.5\height}{\includegraphics[width=0.24\textwidth, height=0.12\textheight]{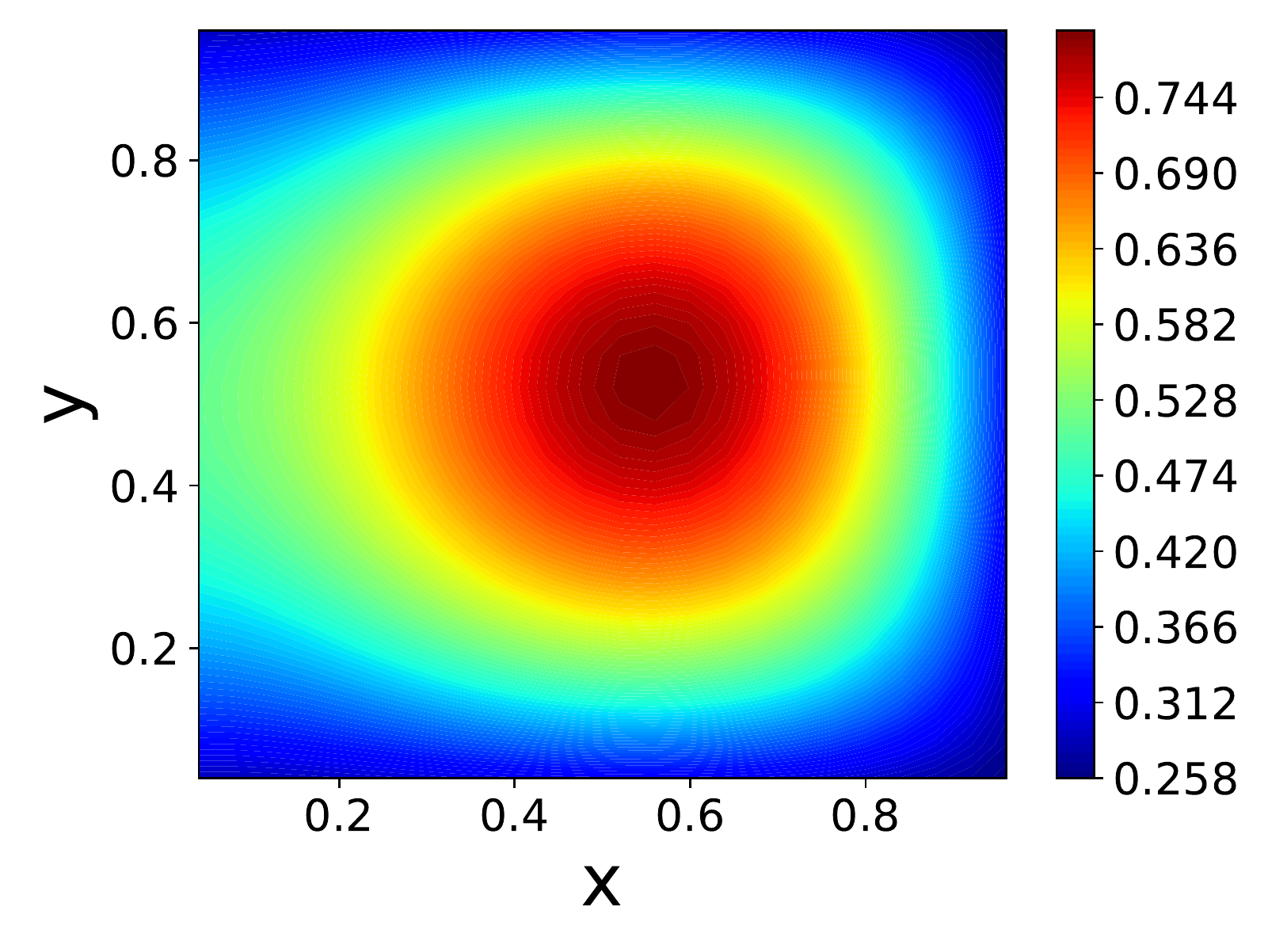}}&
 \raisebox{-.5\height}{\includegraphics[width=0.24\textwidth, height=0.12\textheight]{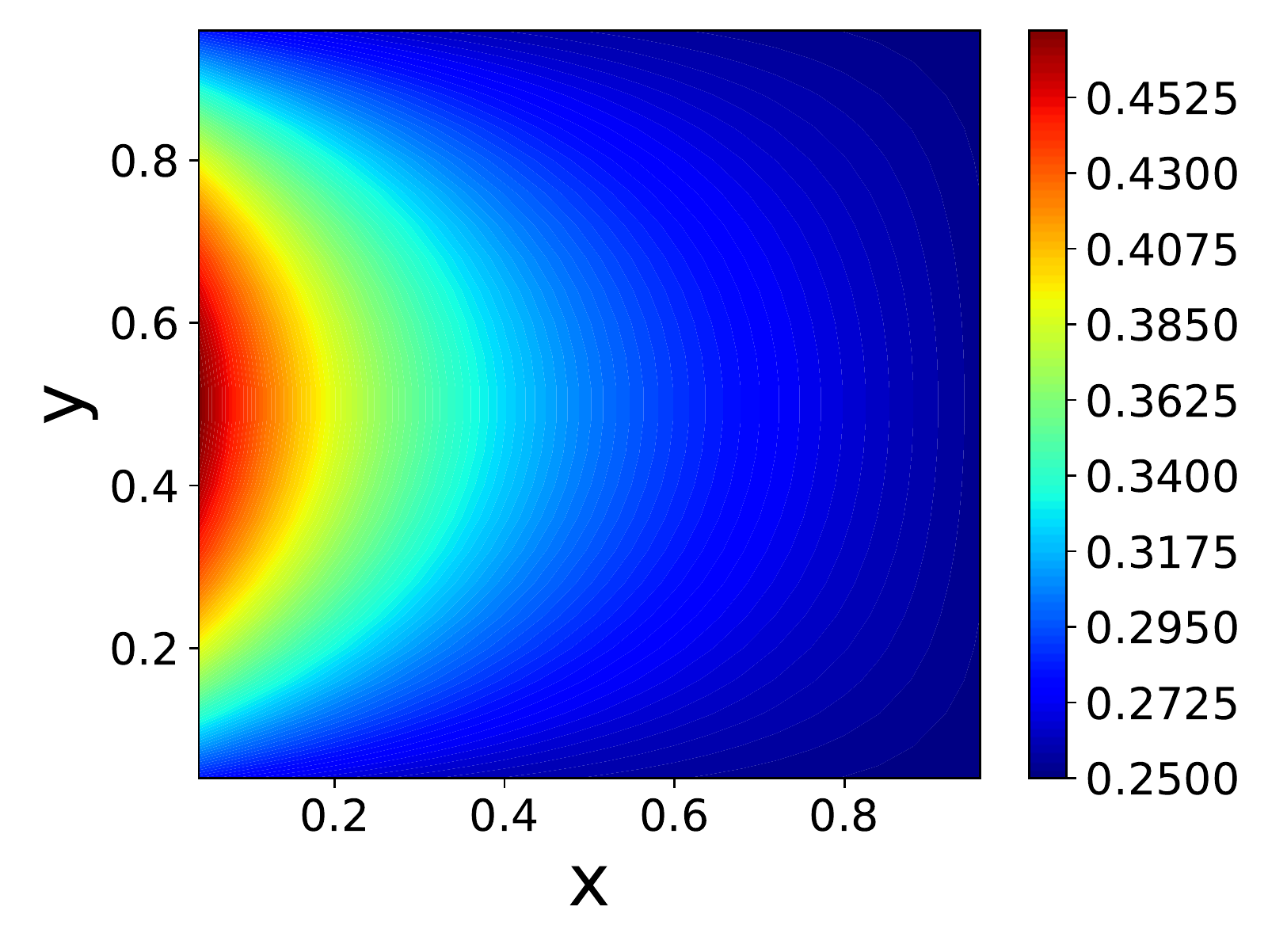}}&
 \raisebox{-.5\height}{\includegraphics[width=0.24\textwidth, height=0.12\textheight]{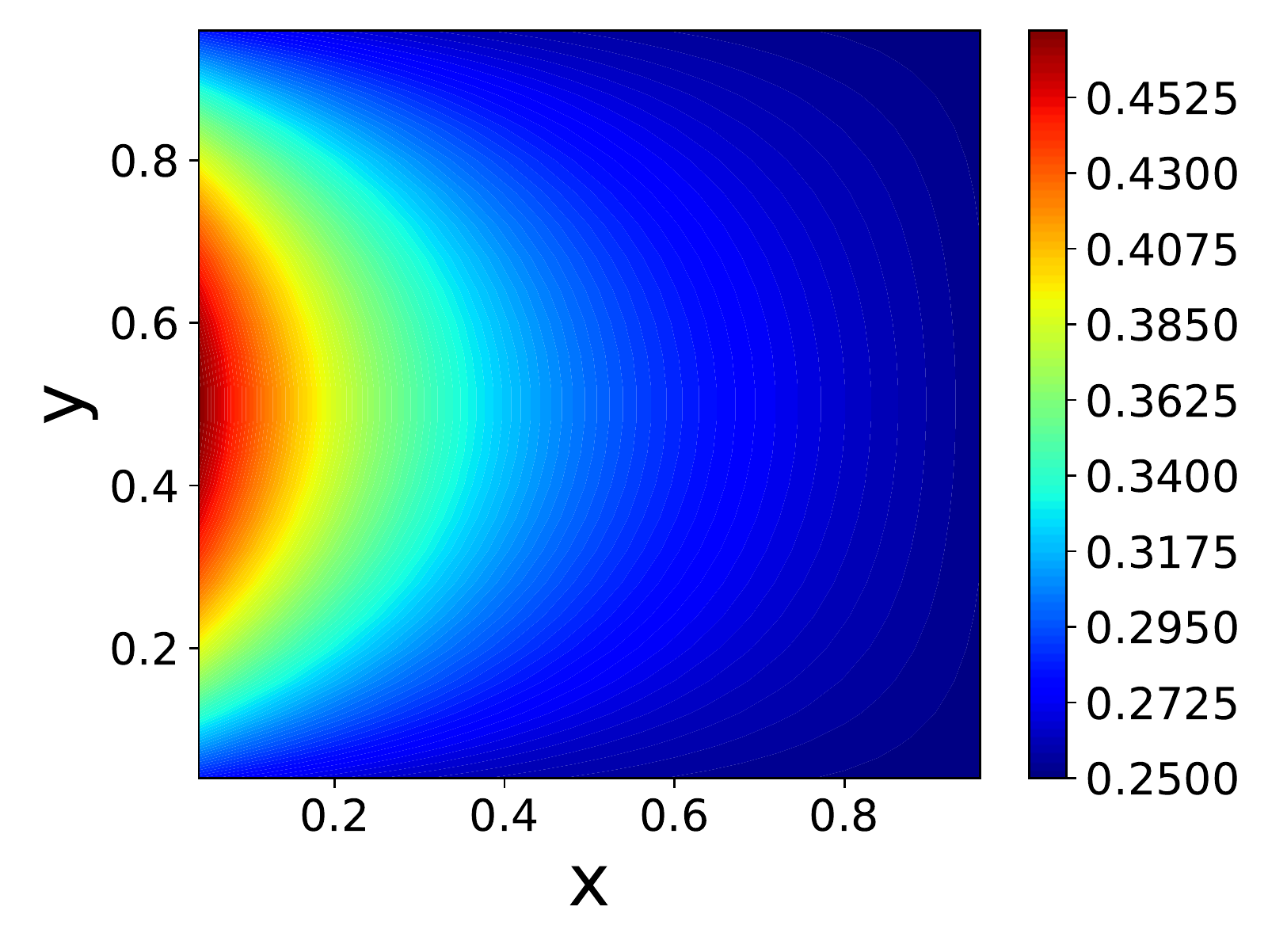}}\\
 \text{DeepONet}&
 \raisebox{-.5\height}{\includegraphics[width=0.24\textwidth, height=0.12\textheight]{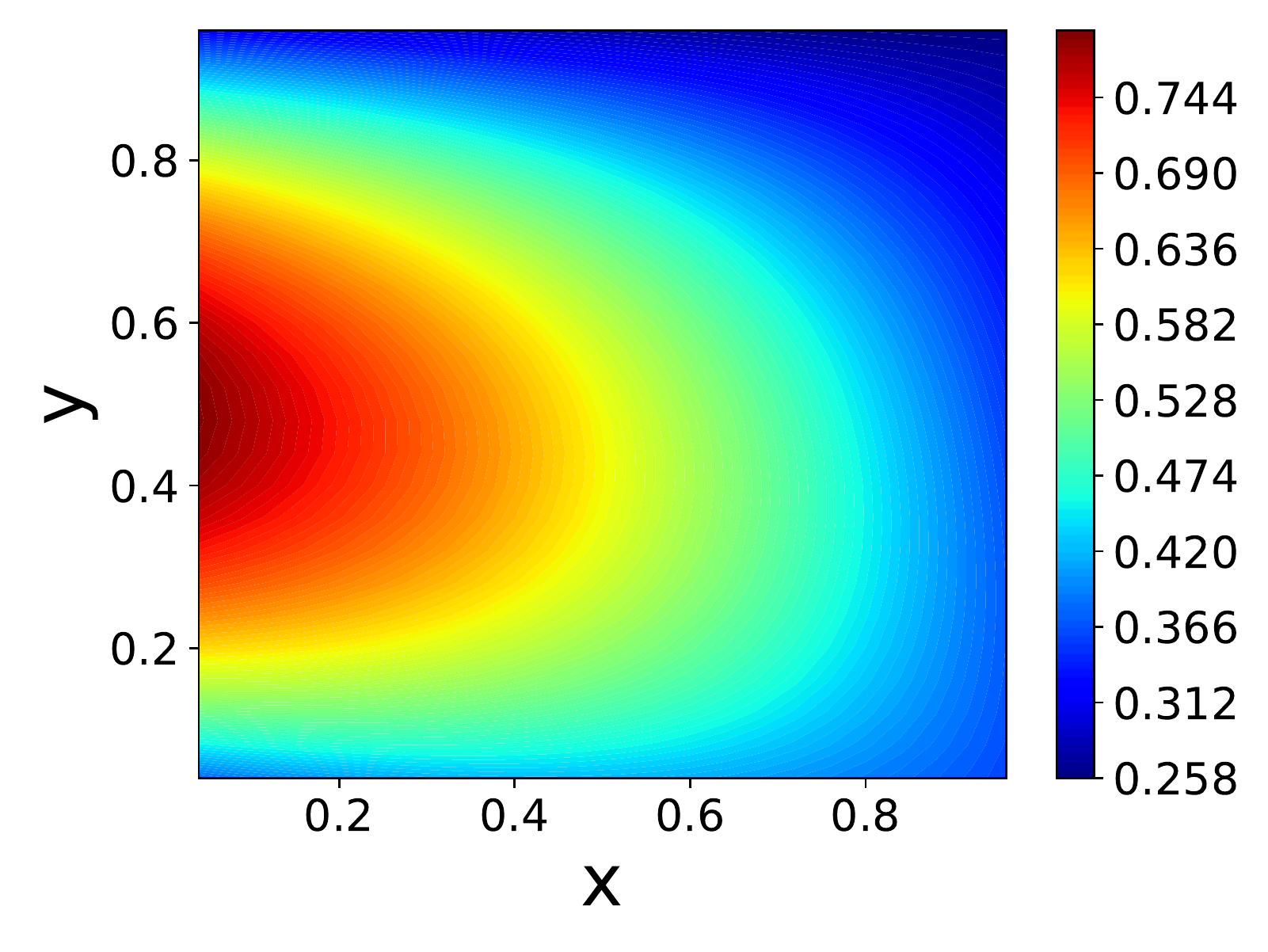}}&
 \raisebox{-.5\height}{\includegraphics[width=0.24\textwidth, height=0.12\textheight]{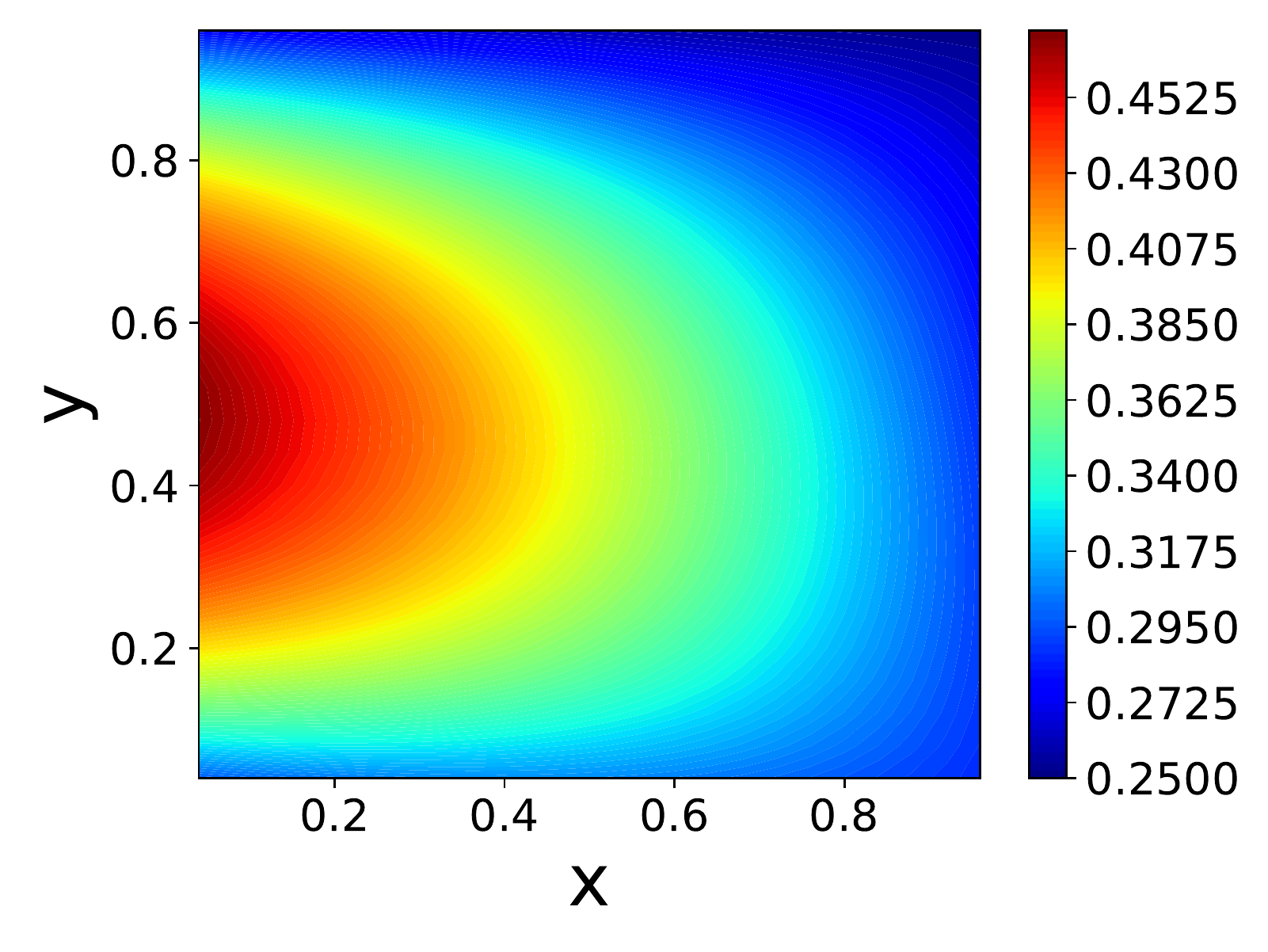}}&
 \raisebox{-.5\height}{\includegraphics[width=0.24\textwidth, height=0.12\textheight]{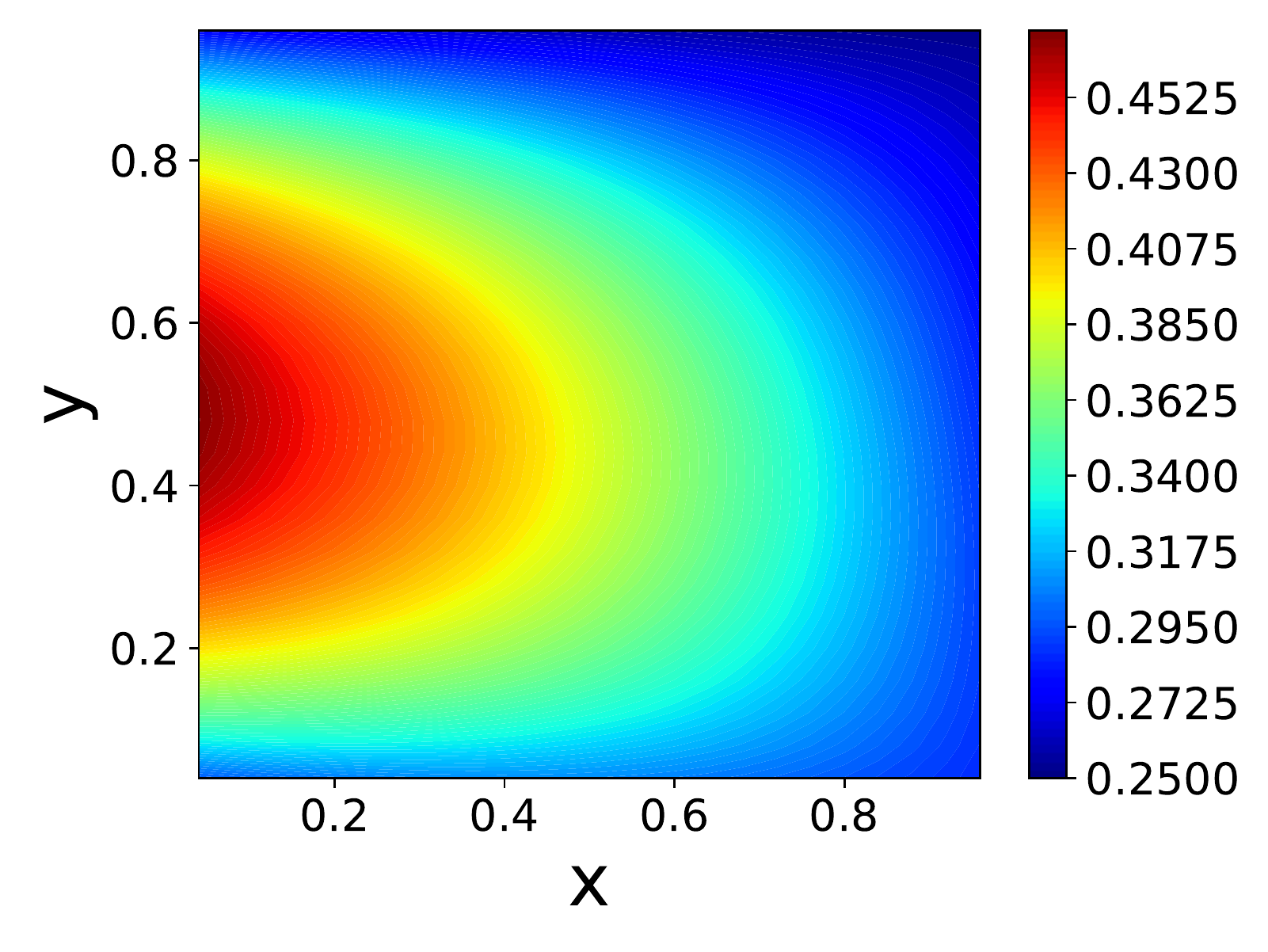}}\\
 \text{TL-DeepONet}&
 \raisebox{-.5\height}{\includegraphics[width=0.24\textwidth, height=0.12\textheight]{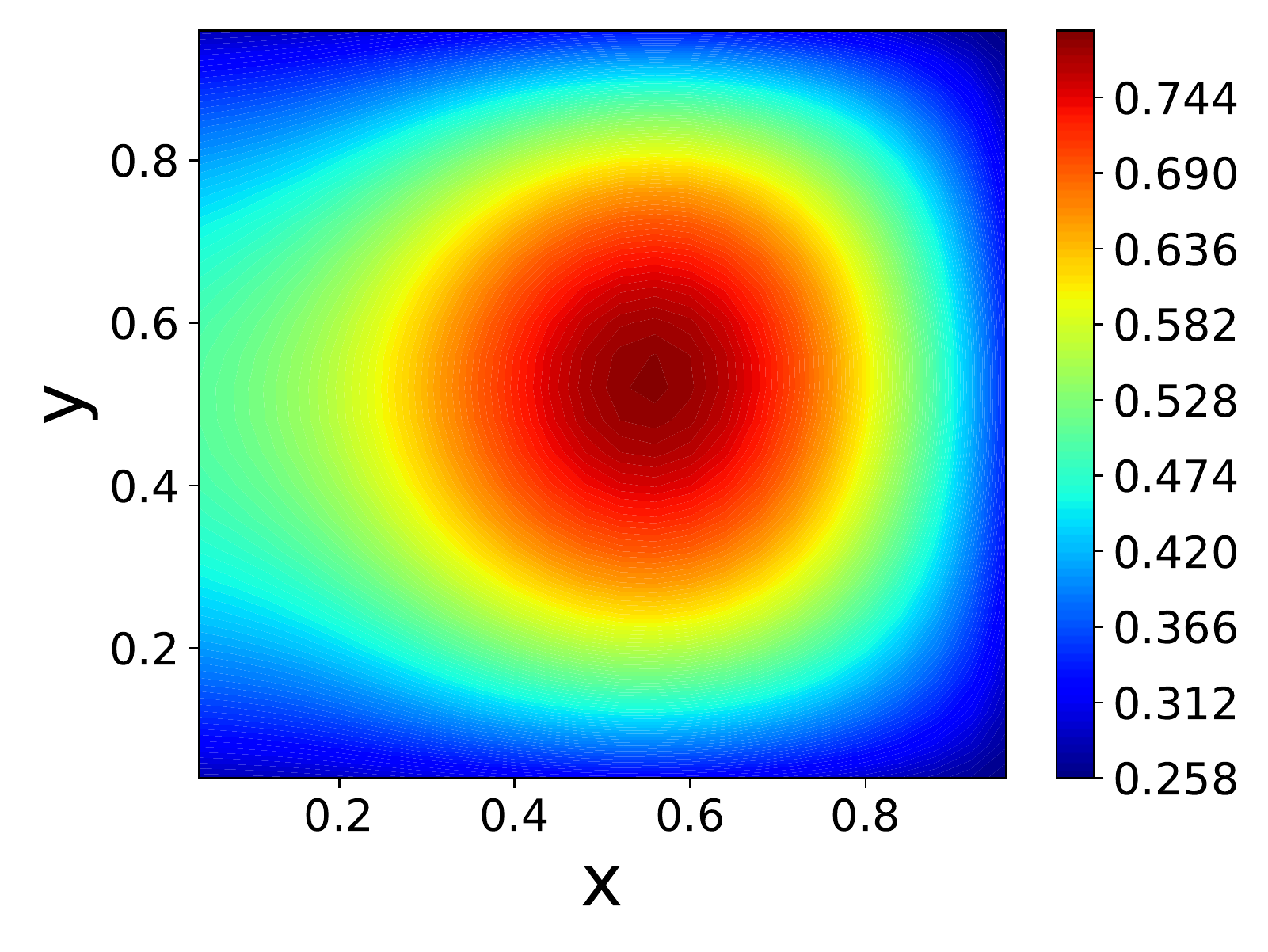}}&
 \raisebox{-.5\height}{\includegraphics[width=0.24\textwidth, height=0.12\textheight]{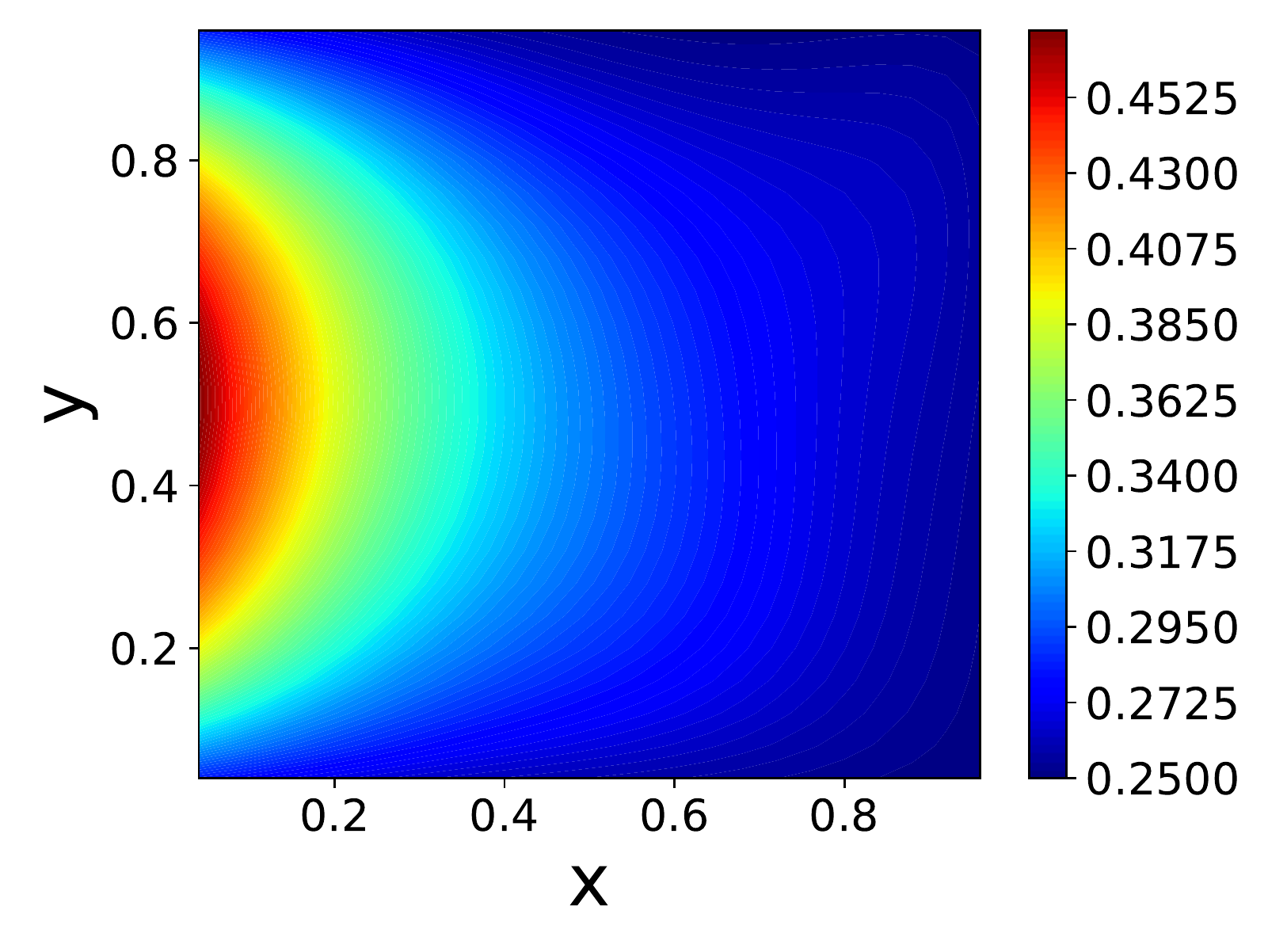}}&
 \raisebox{-.5\height}{\includegraphics[width=0.24\textwidth, height=0.12\textheight]{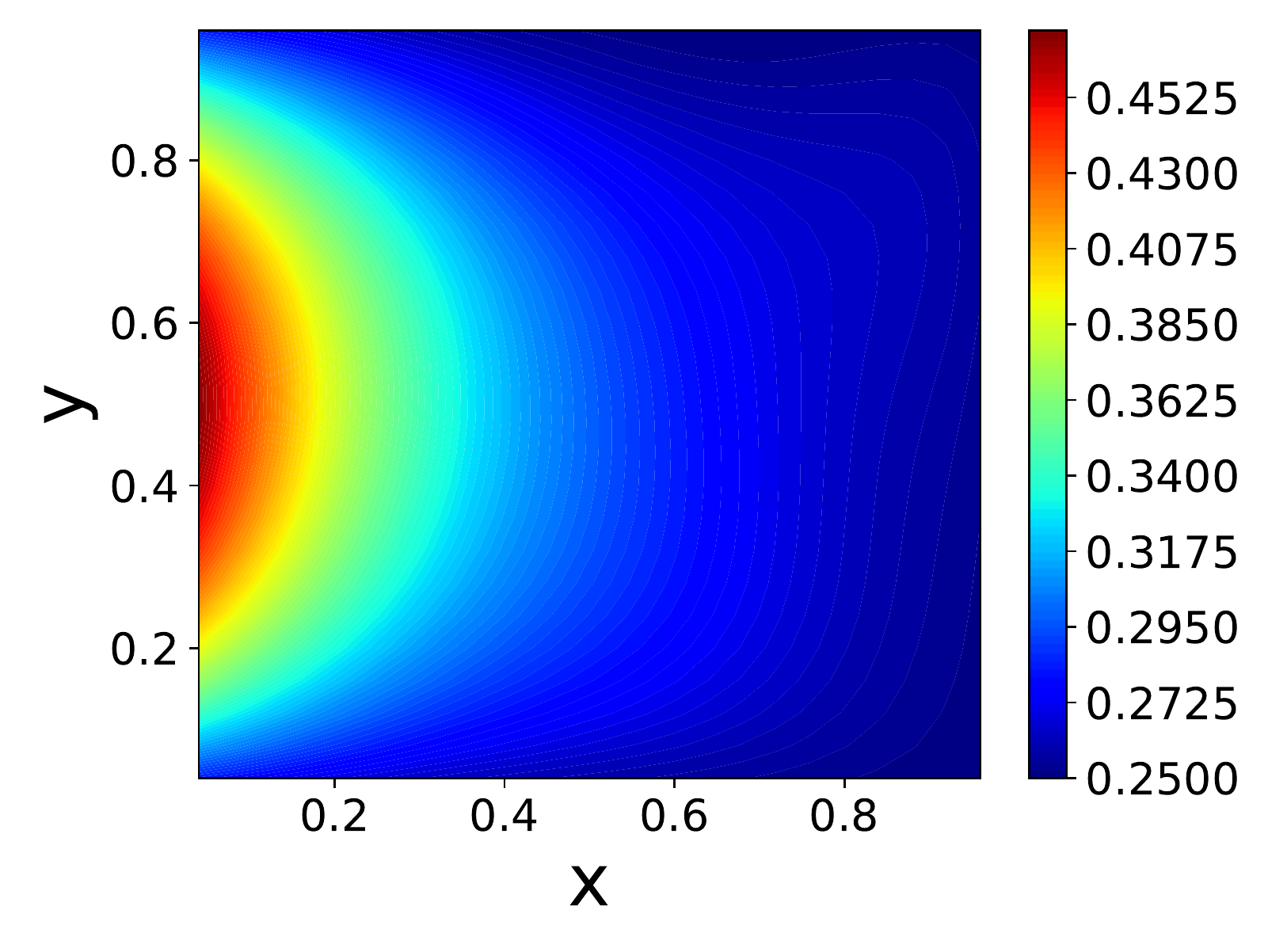}}\\
 \end{tabular}

 \begin{tabular}{cccc}
 \text{Initical condition} \quad & $t=0.1$ & $t=0.5$ & $t=10$\\
 \raisebox{-.5\height}{\includegraphics[width=0.24\textwidth, height=0.12\textheight]{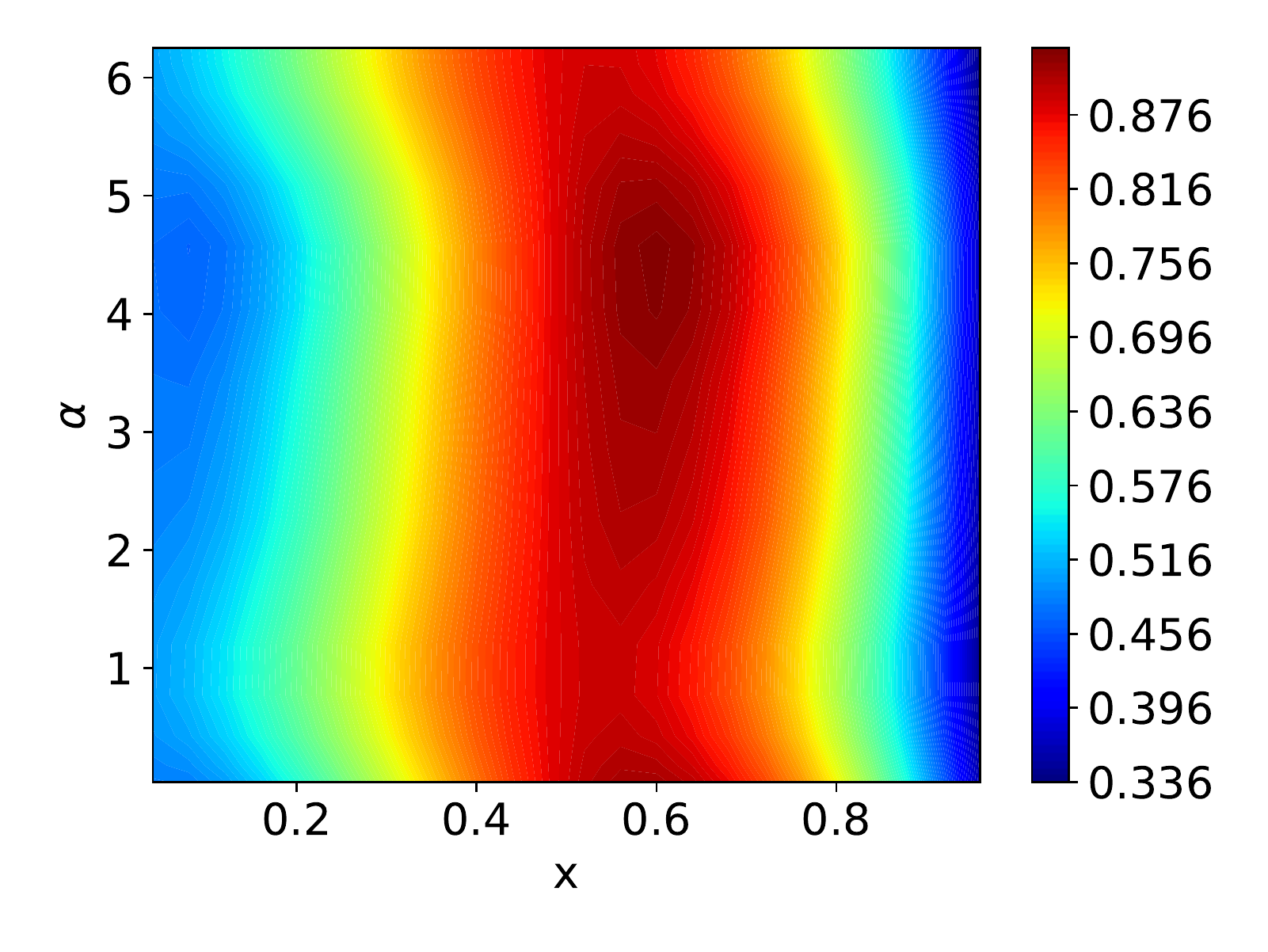}}&
 \raisebox{-.5\height}{\includegraphics[width=0.24\textwidth, height=0.12\textheight]{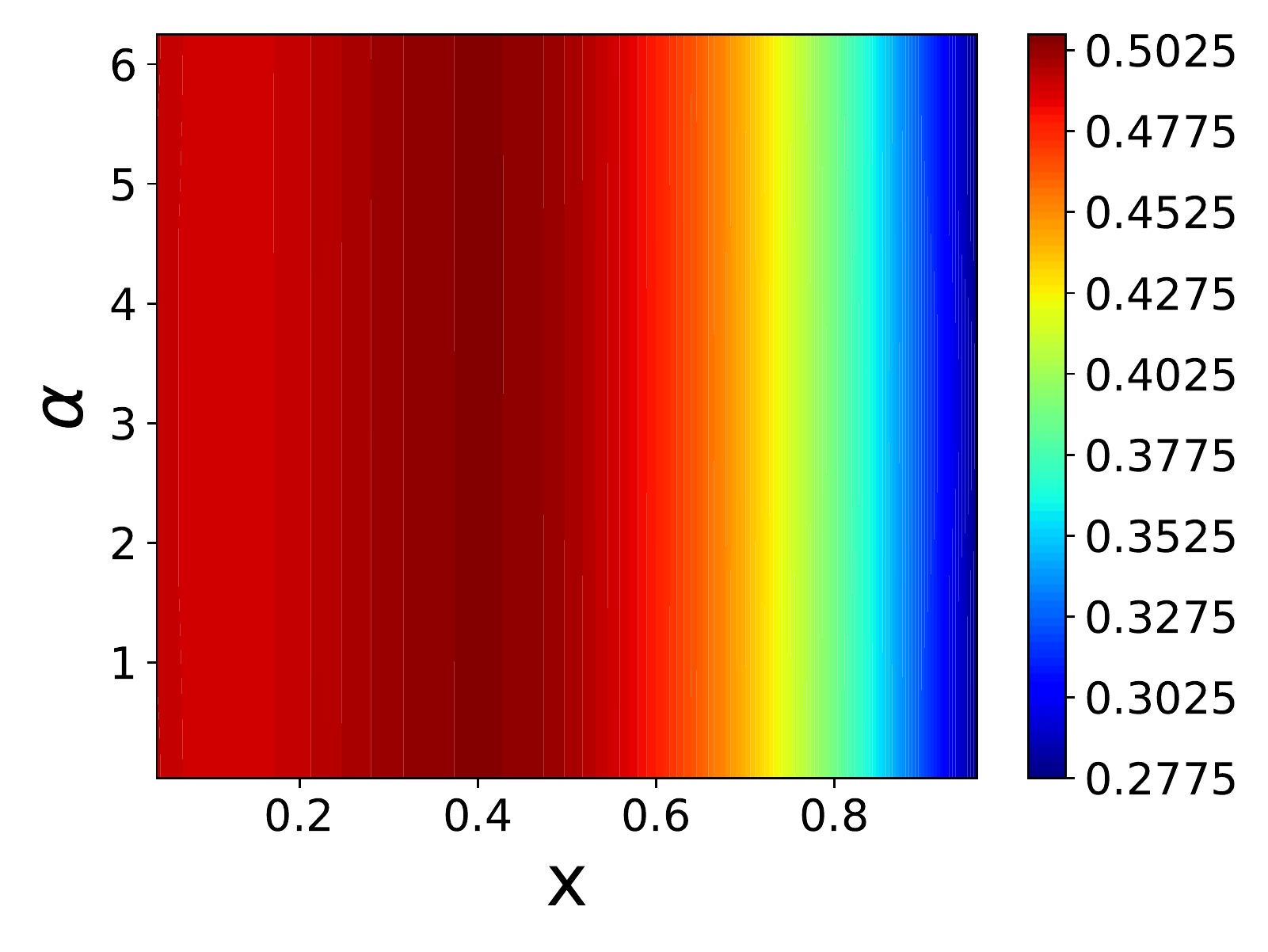}}&
 \raisebox{-.5\height}{\includegraphics[width=0.24\textwidth, height=0.12\textheight]{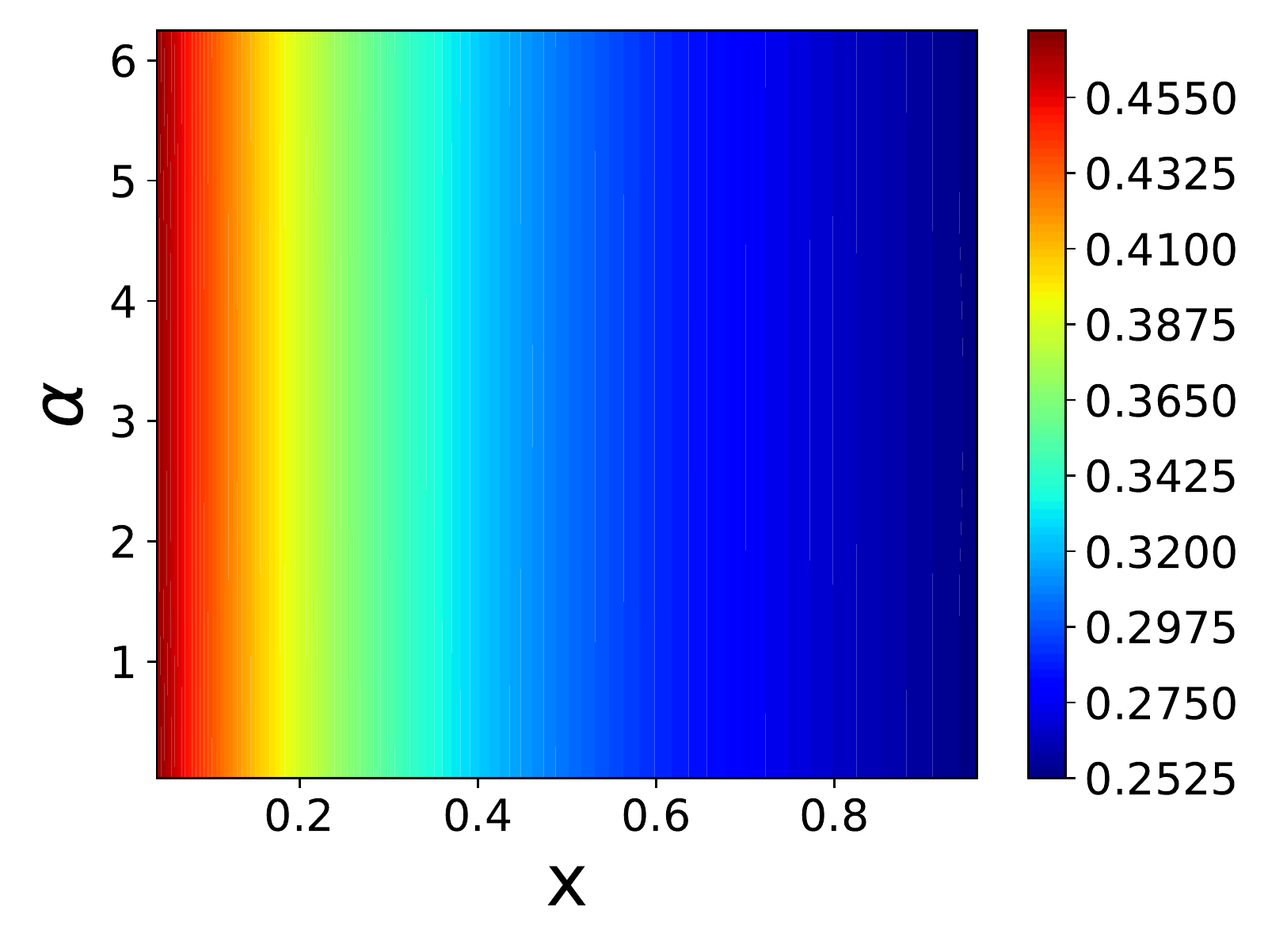}}&
 \raisebox{-.5\height}{\includegraphics[width=0.24\textwidth, height=0.12\textheight]{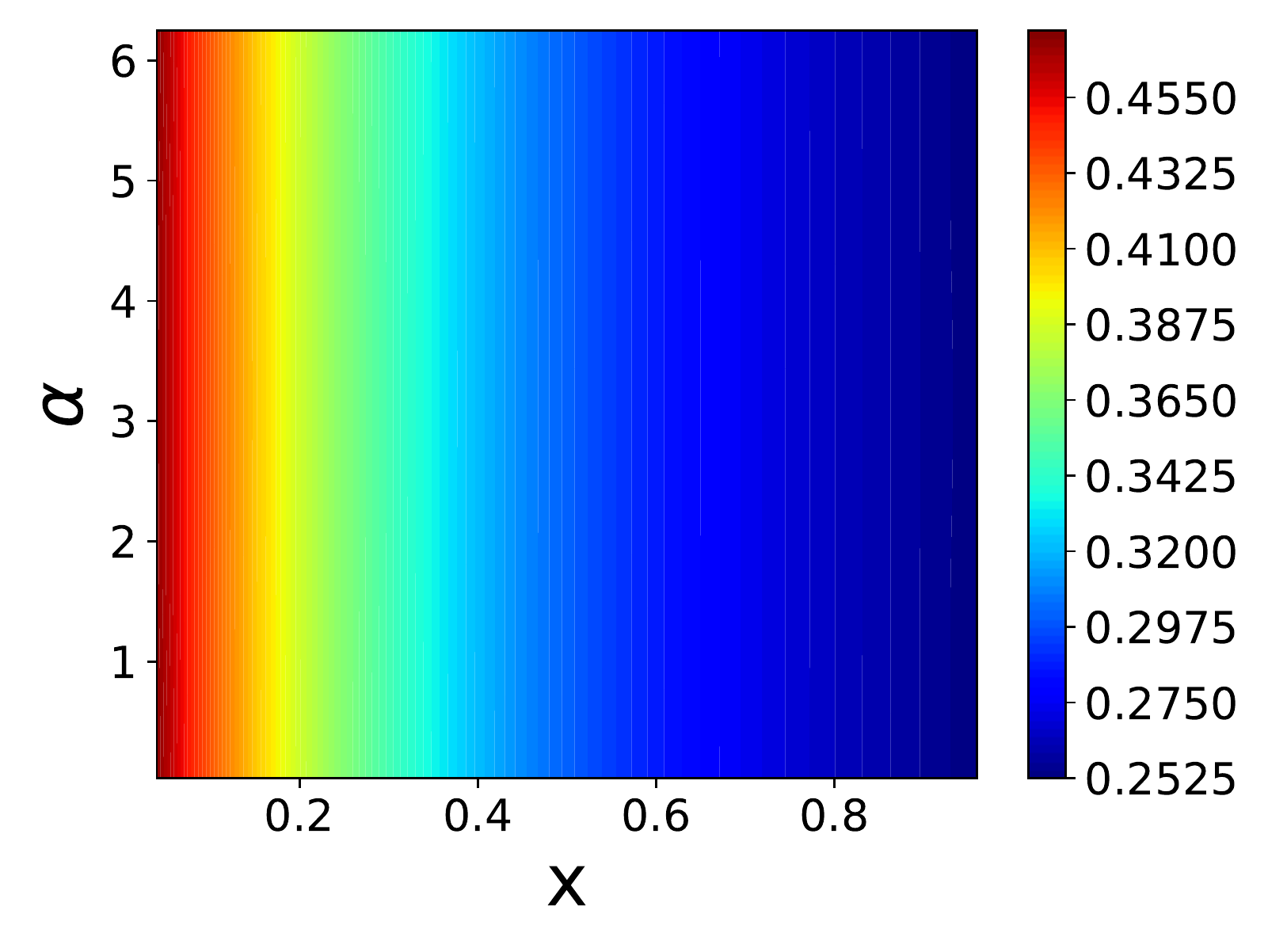}}\\
 \text{DeepONet}&
 \raisebox{-.5\height}{\includegraphics[width=0.24\textwidth, height=0.12\textheight]{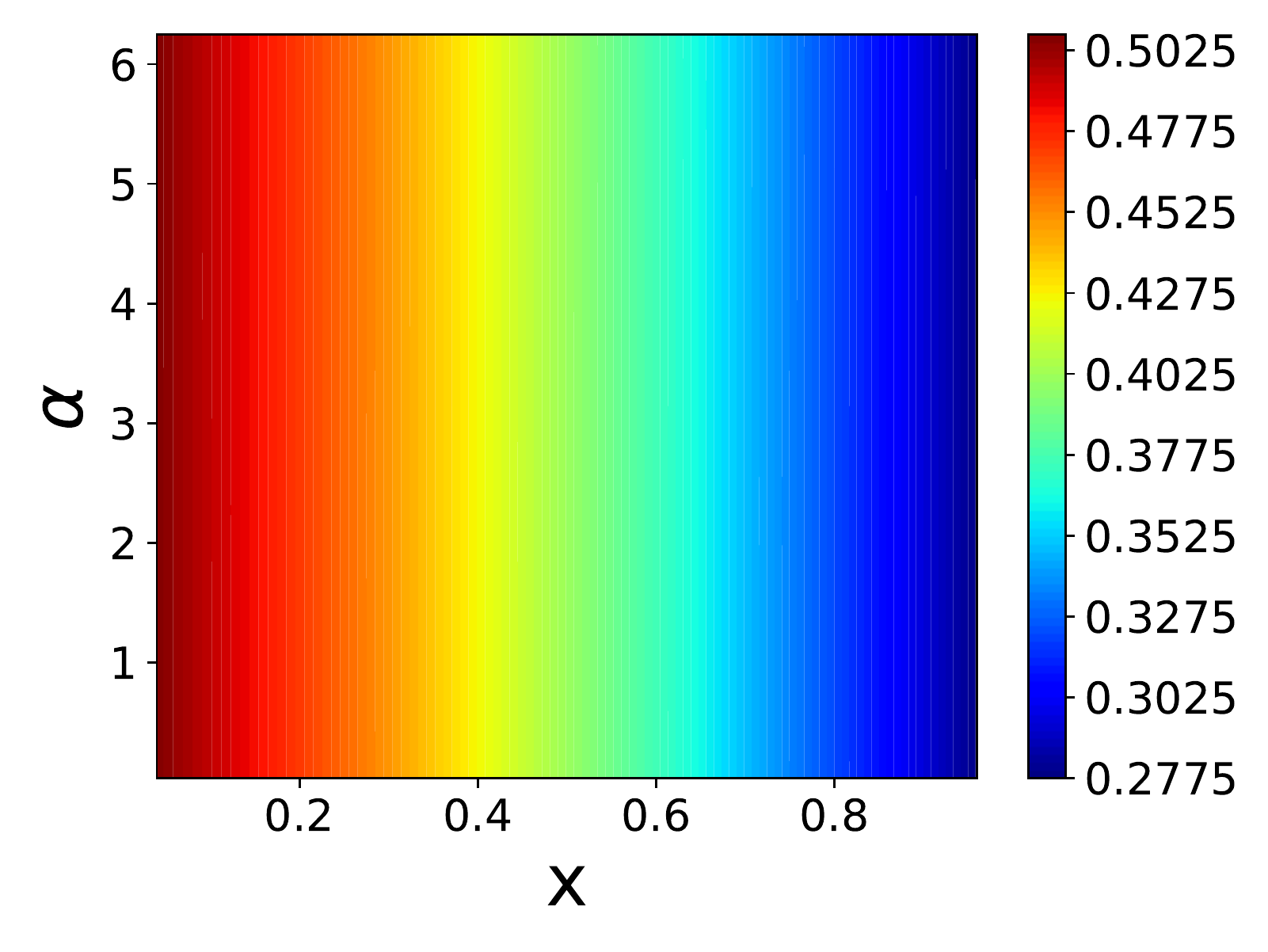}}&
 \raisebox{-.5\height}{\includegraphics[width=0.24\textwidth, height=0.12\textheight]{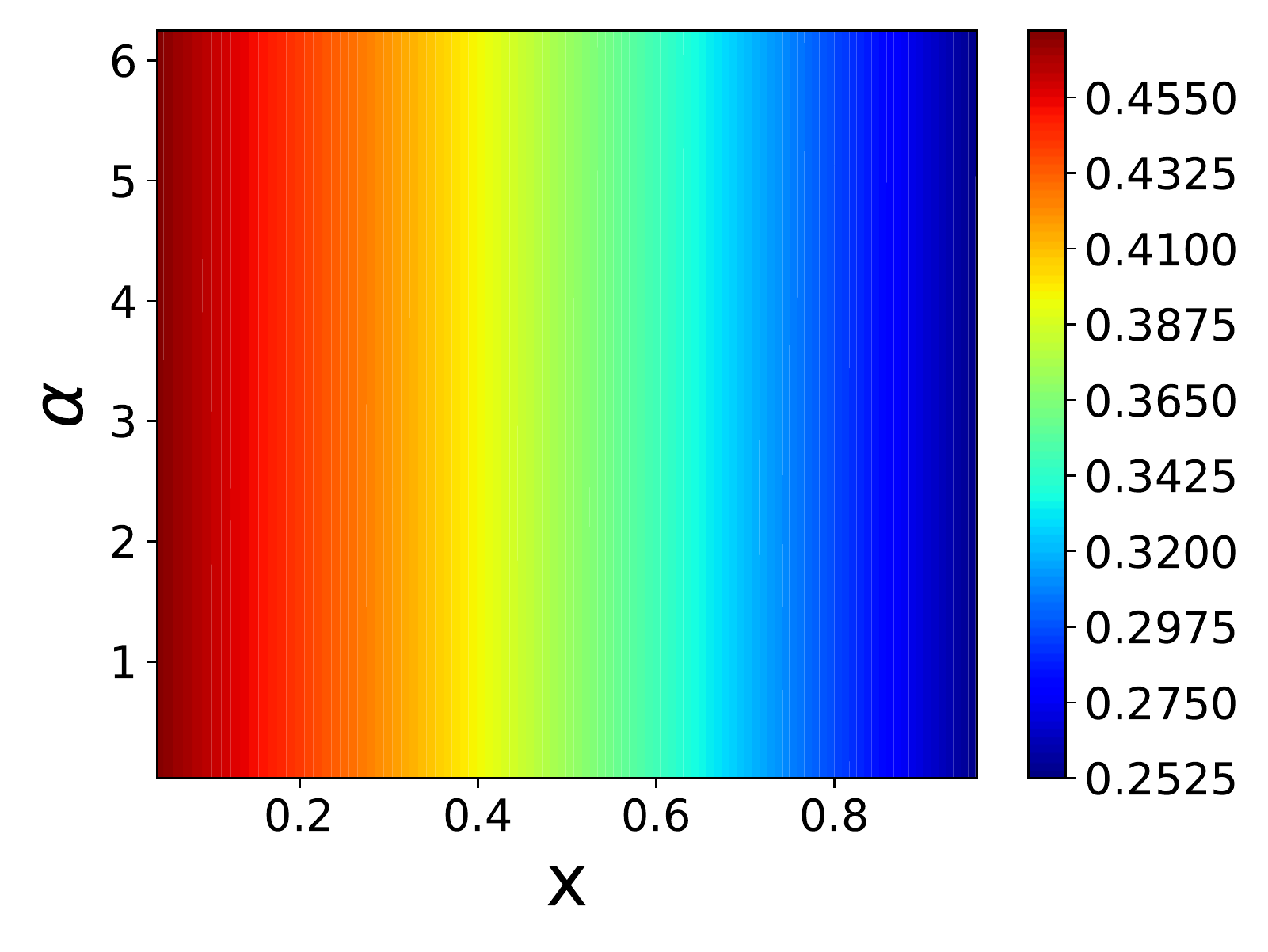}}&
 \raisebox{-.5\height}{\includegraphics[width=0.24\textwidth, height=0.12\textheight]{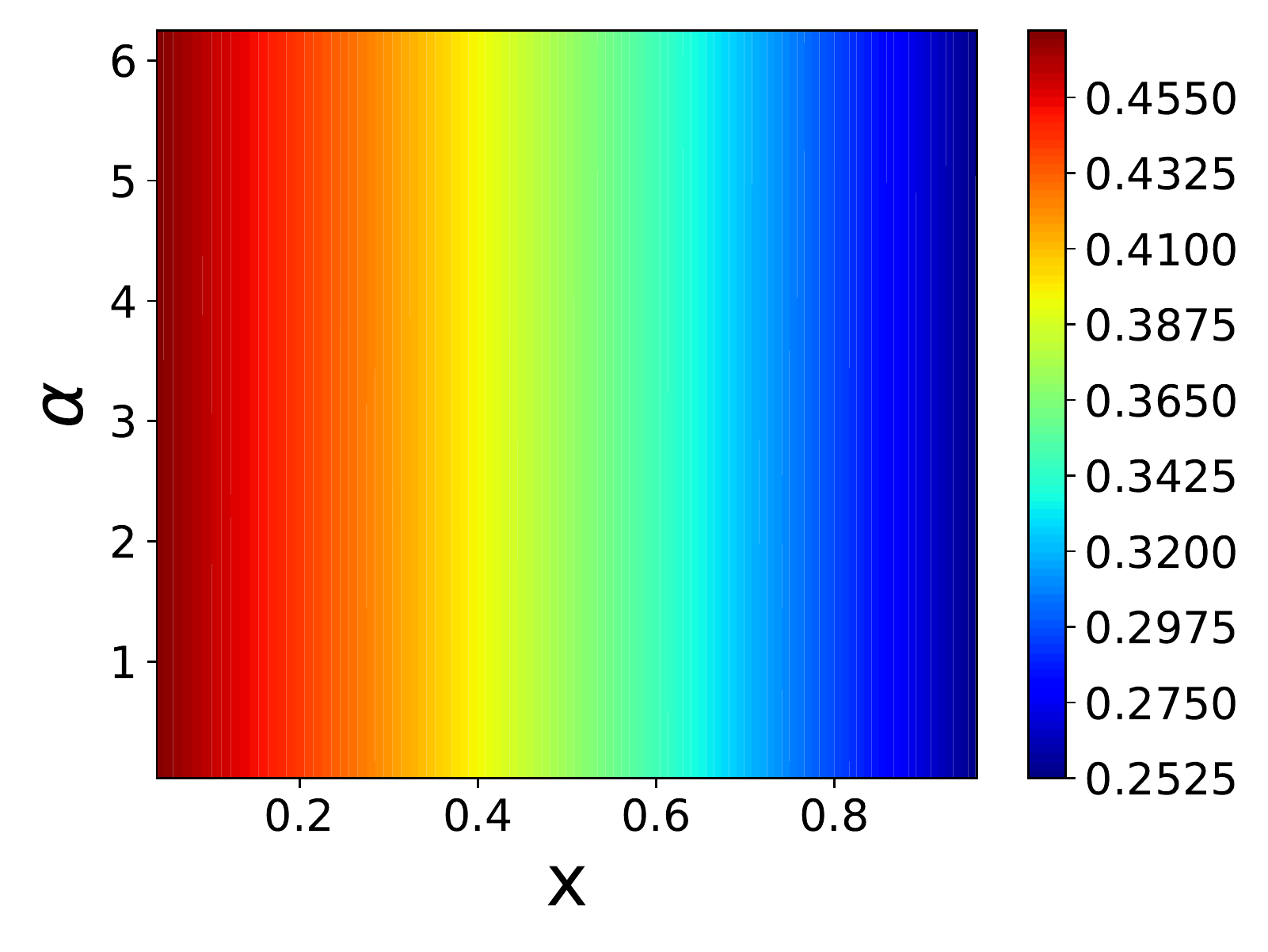}}\\
 \text{TL-DeepONet}&
 \raisebox{-.5\height}{\includegraphics[width=0.24\textwidth, height=0.12\textheight]{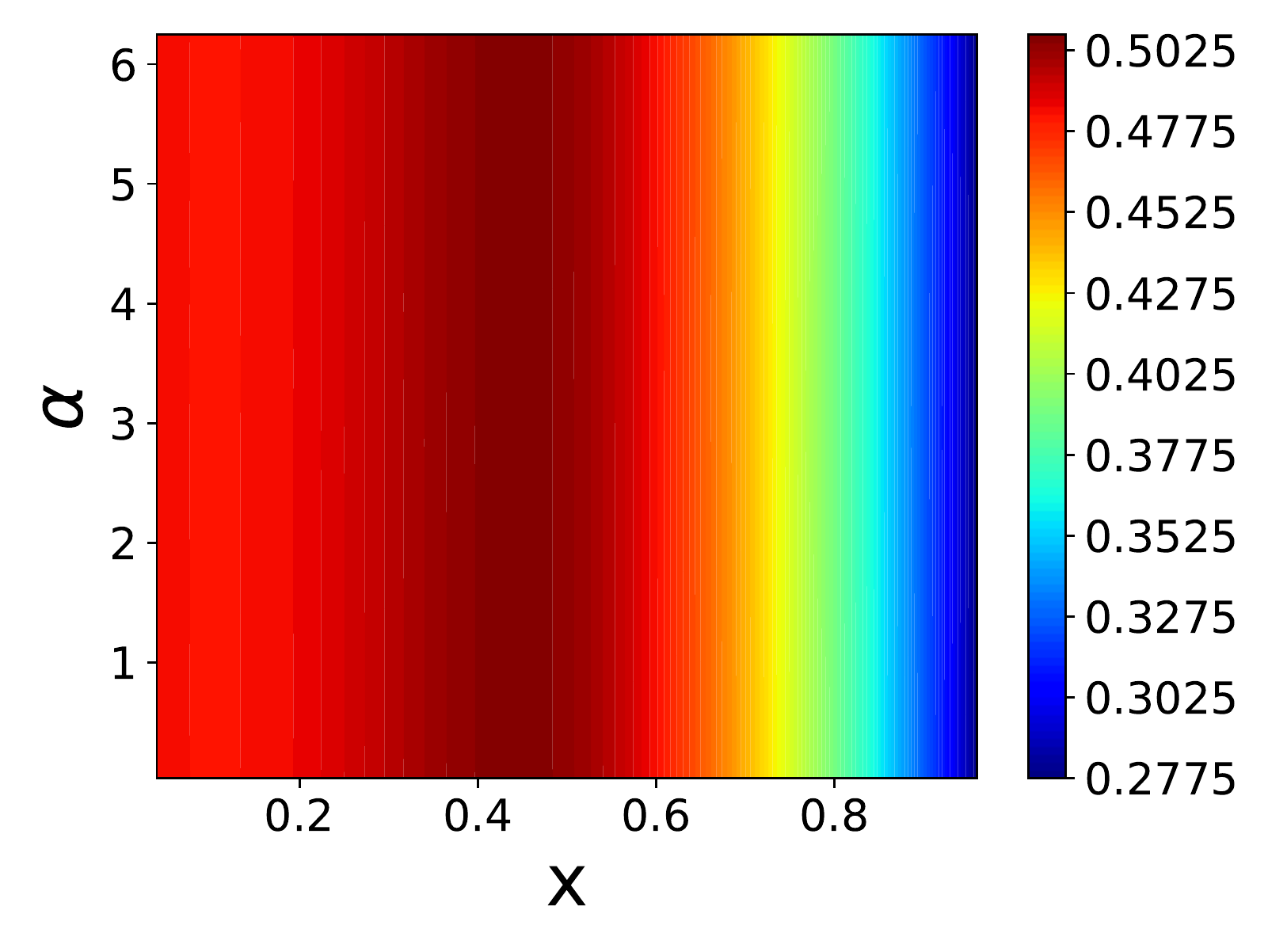}}&
 \raisebox{-.5\height}{\includegraphics[width=0.24\textwidth, height=0.12\textheight]{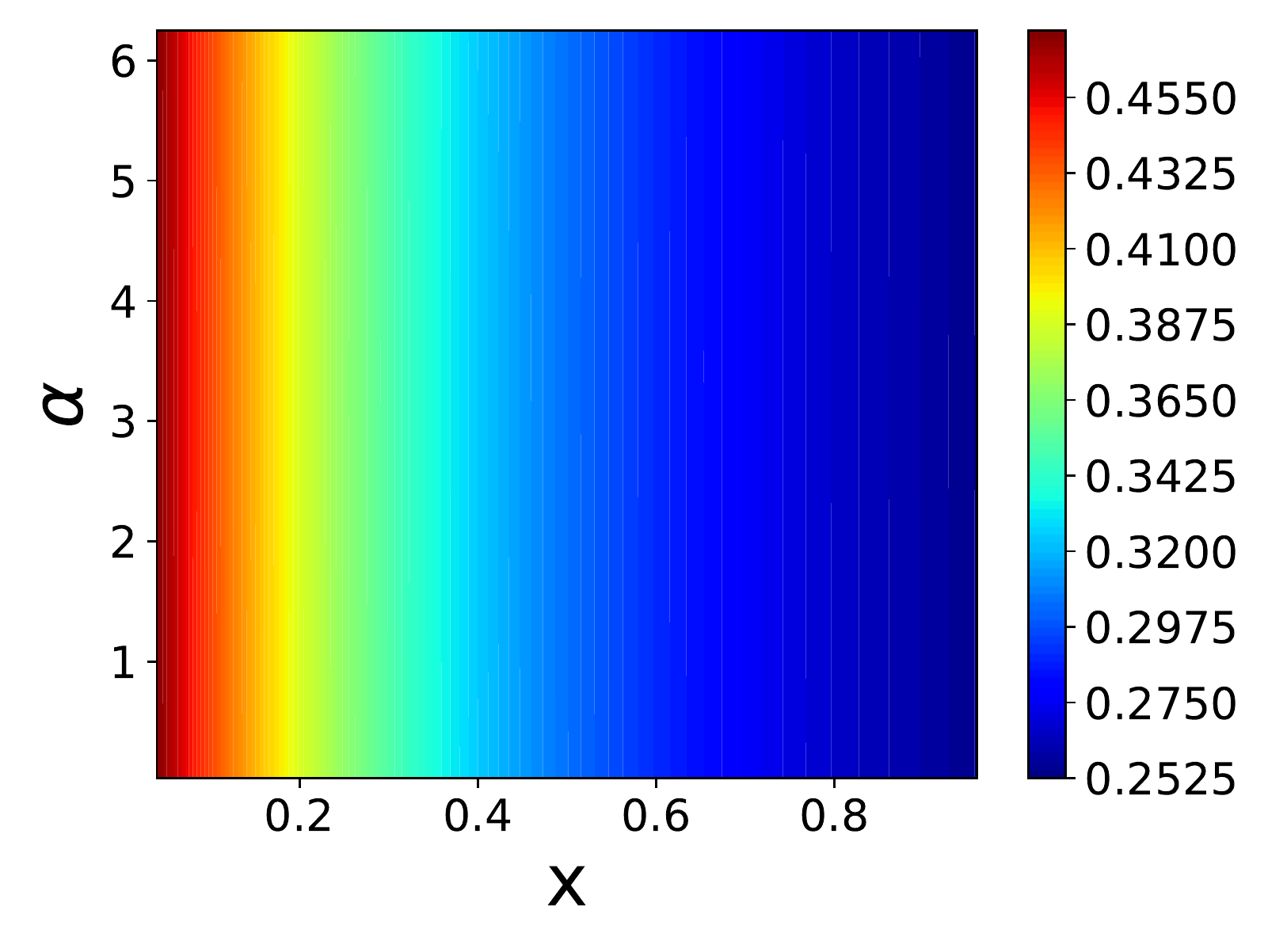}}&
 \raisebox{-.5\height}{\includegraphics[width=0.24\textwidth, height=0.12\textheight]{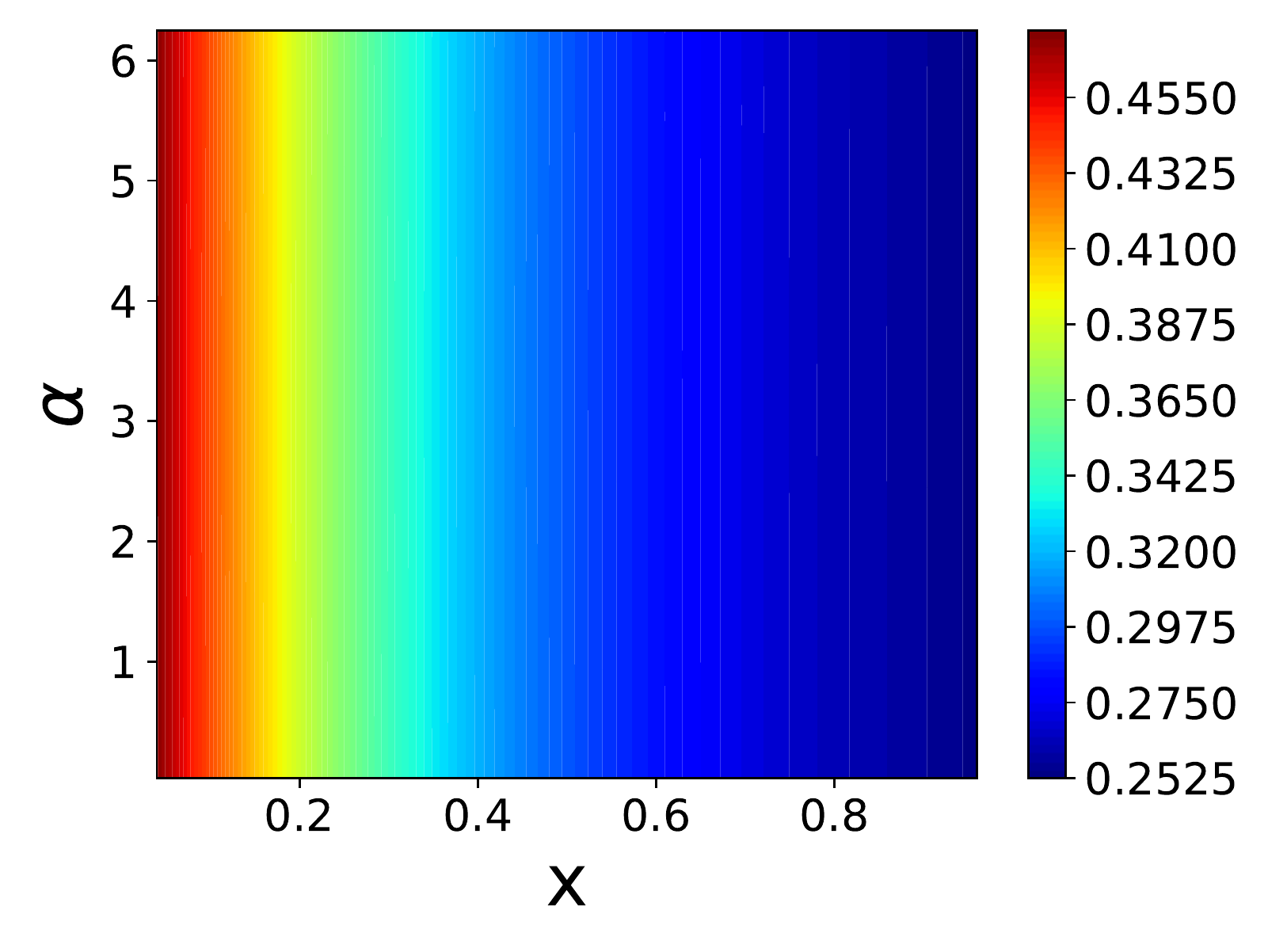}}\\
 \end{tabular}
 \caption{Results on 2D radiative transfer equation with $\eps=$1e-4. The top three rows are snapshots of $\rho(t,x,y)$ of reference solutions, approximate solutions predicted by DeepONet and approximate solutions predicted by TL-DeepONet respectively. The bottom three rows are snapshots of $f(t, x,y=0.5,\alpha)$ of reference solutions, approximate solutions predicted by DeepONet and approximate solutions predicted by TL-DeepONet respectively.}
 \label{fig:rte_eps_zpzzz1} 
 \end{figure*}

\begin{figure*}
 \centering
 \setlength{\tabcolsep}{0pt}
 \renewcommand{\arraystretch}{-1}
 \begin{tabular}{cccc}
 \text{Initical condition} \quad & $t=0.5$ & $t=1$ & $t=10$\\
 \raisebox{-.5\height}{\includegraphics[width=0.24\textwidth, height=0.12\textheight]{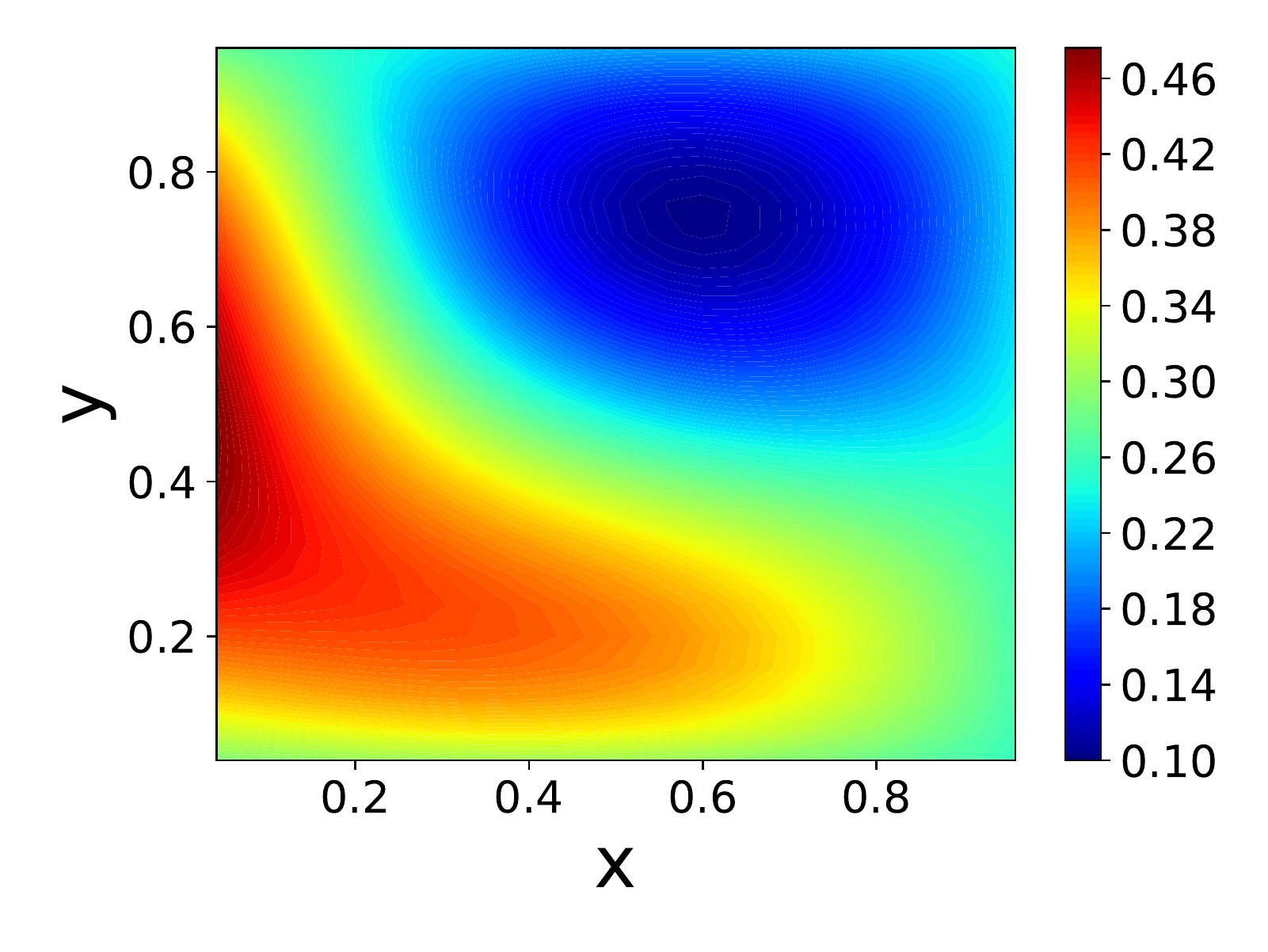}}&
 \raisebox{-.5\height}{\includegraphics[width=0.24\textwidth, height=0.12\textheight]{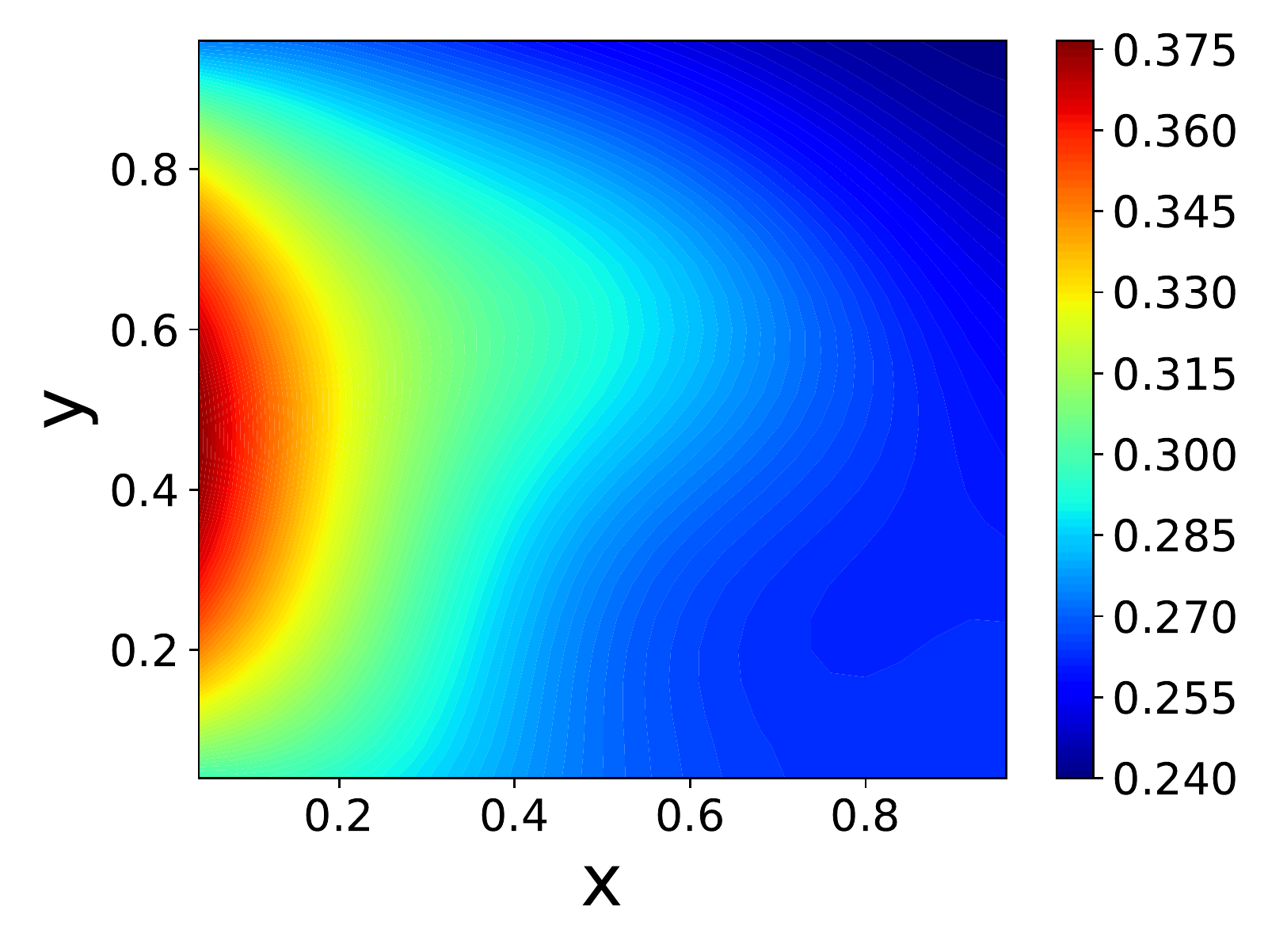}}&
 \raisebox{-.5\height}{\includegraphics[width=0.24\textwidth, height=0.12\textheight]{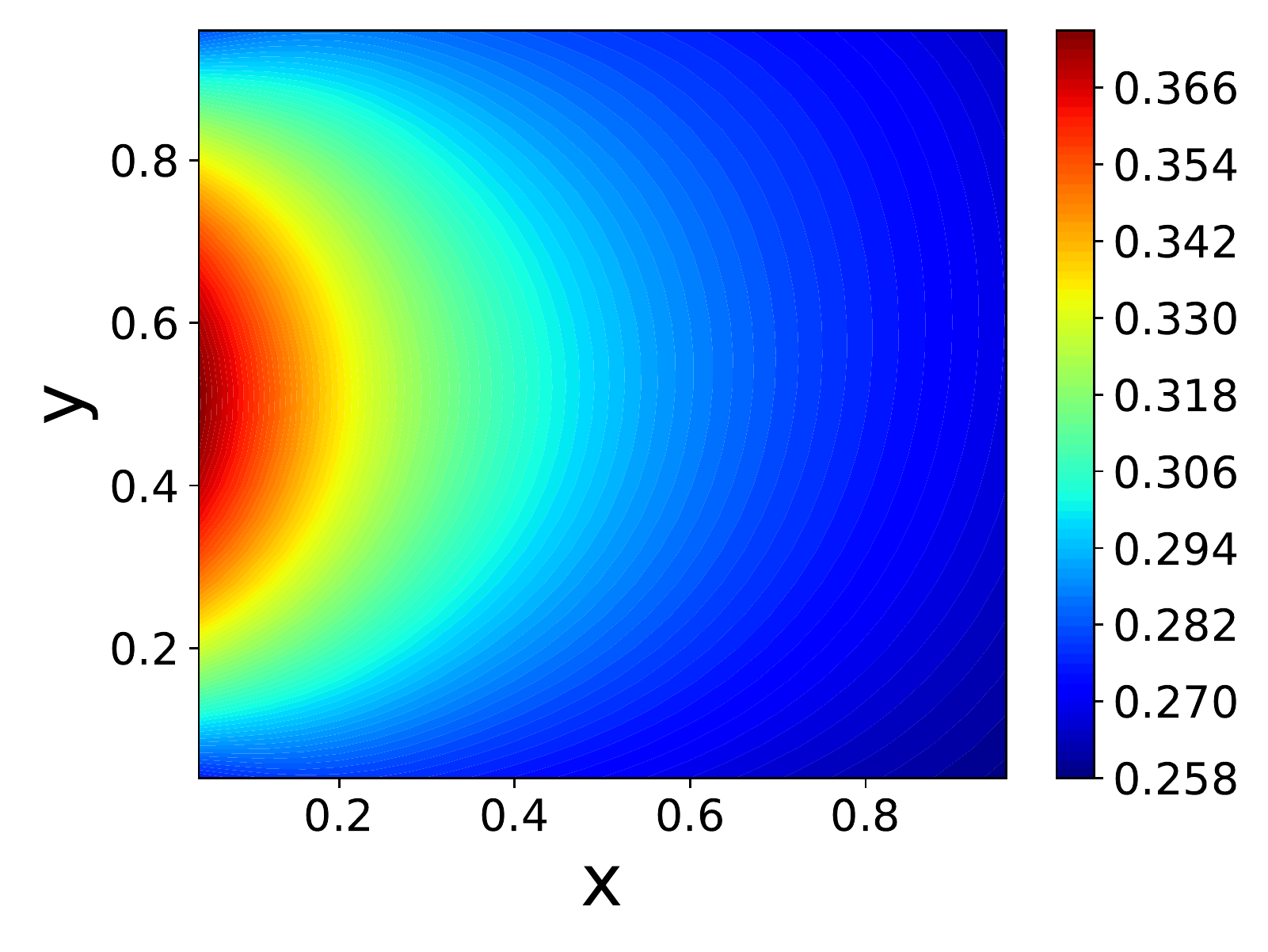}}&
 \raisebox{-.5\height}{\includegraphics[width=0.24\textwidth, height=0.12\textheight]{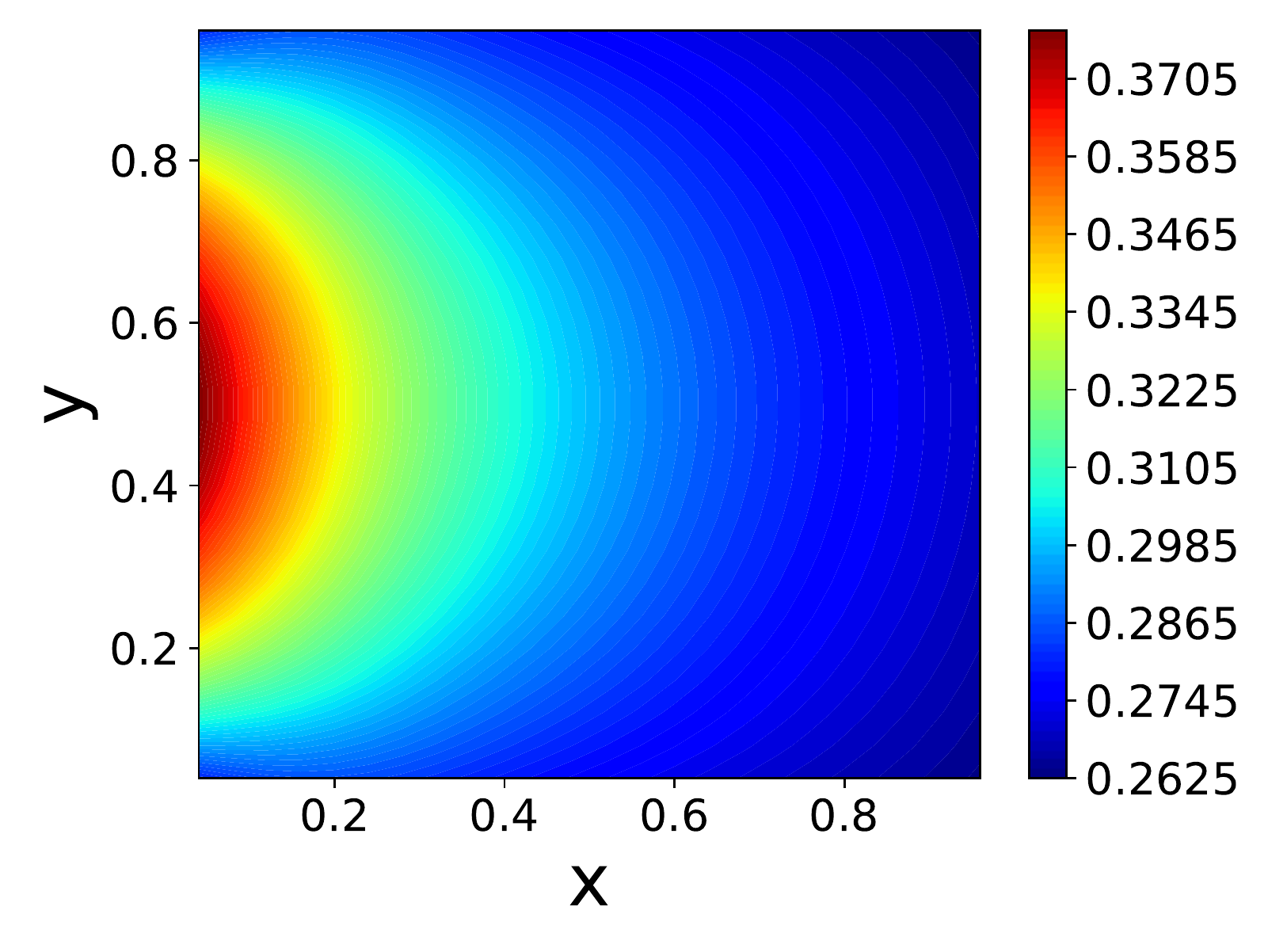}}\\
 \text{DeepONet}&
 \raisebox{-.5\height}{\includegraphics[width=0.24\textwidth, height=0.12\textheight]{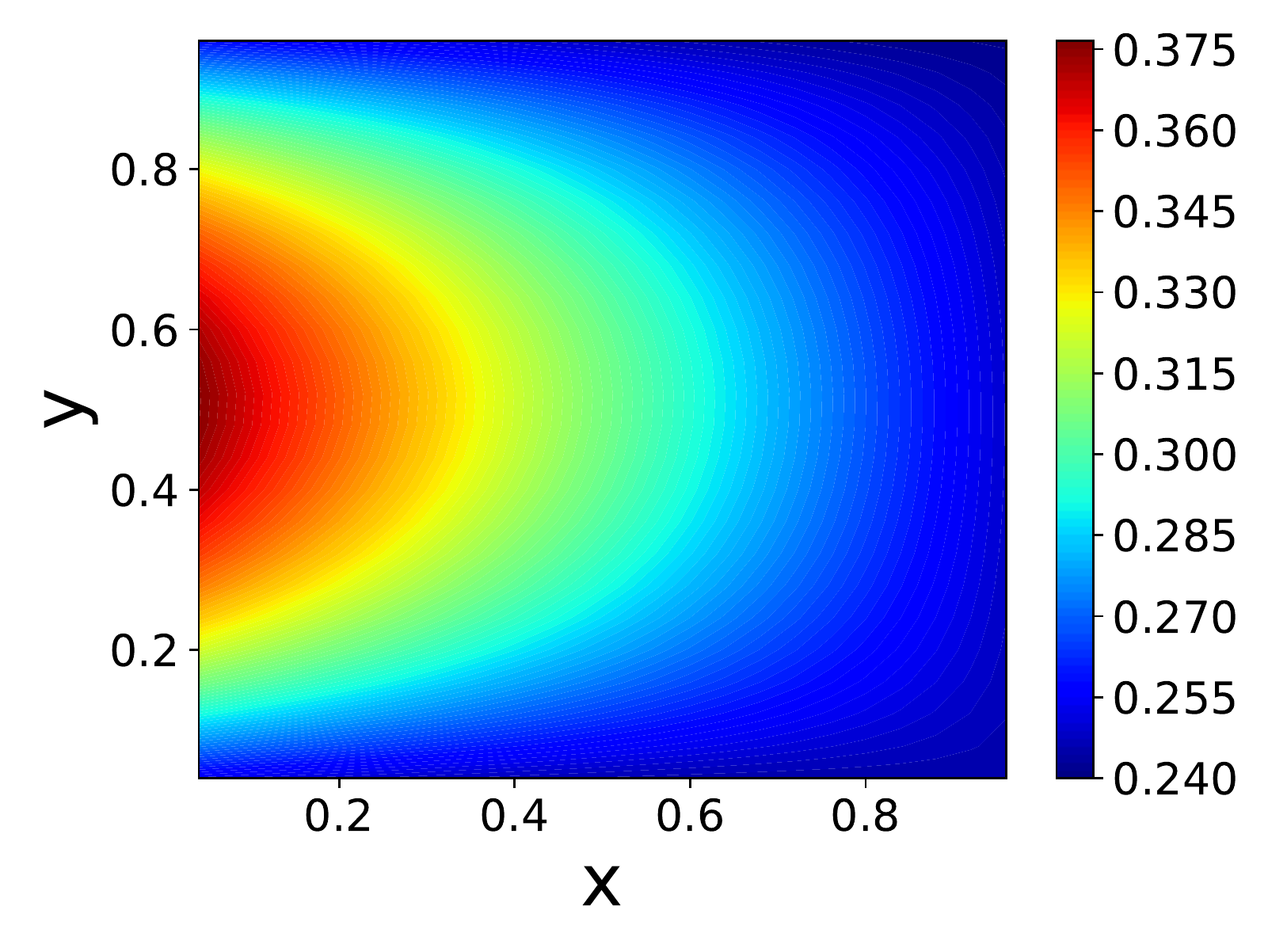}}&
 \raisebox{-.5\height}{\includegraphics[width=0.24\textwidth, height=0.12\textheight]{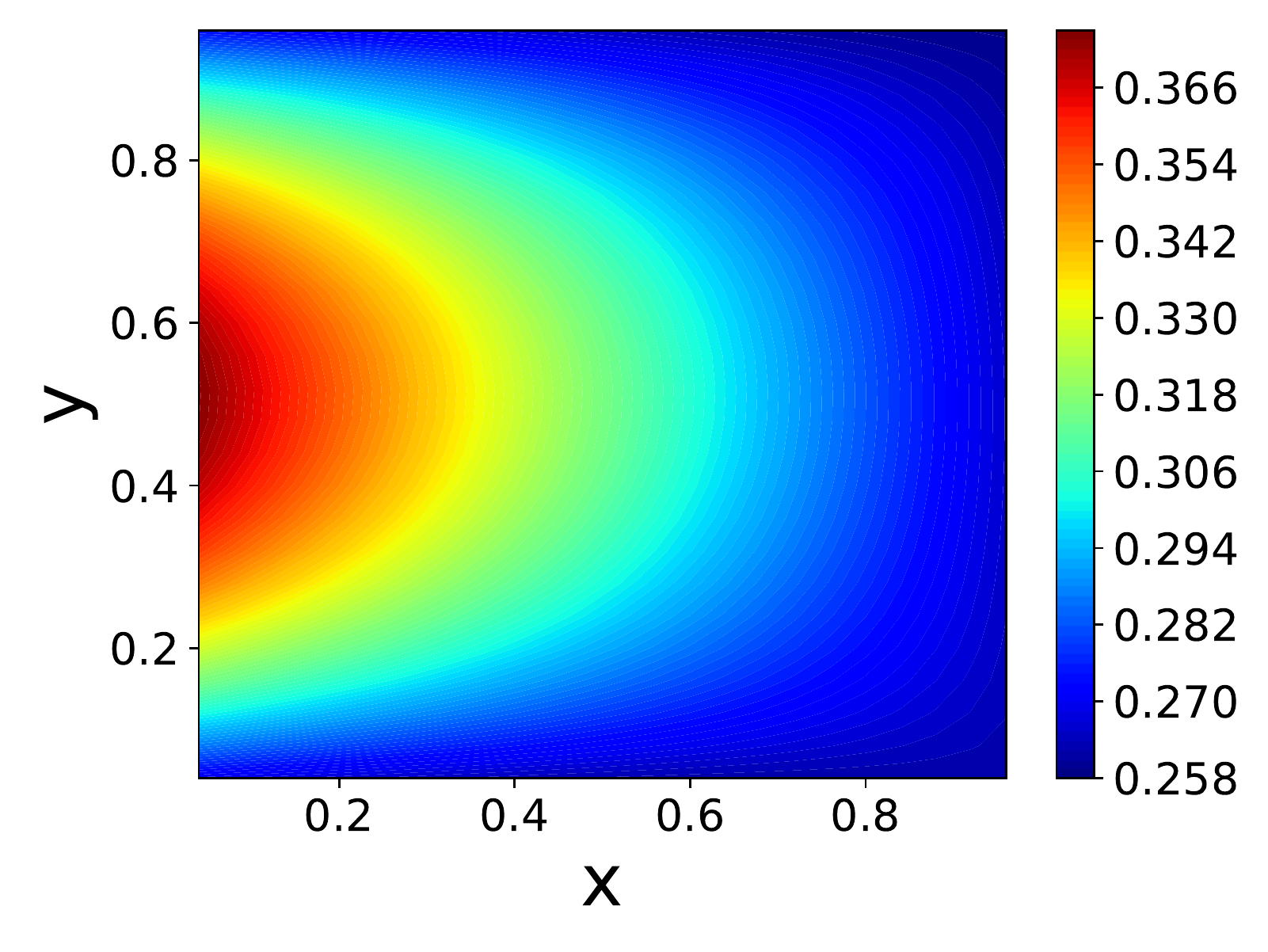}}&
 \raisebox{-.5\height}{\includegraphics[width=0.24\textwidth, height=0.12\textheight]{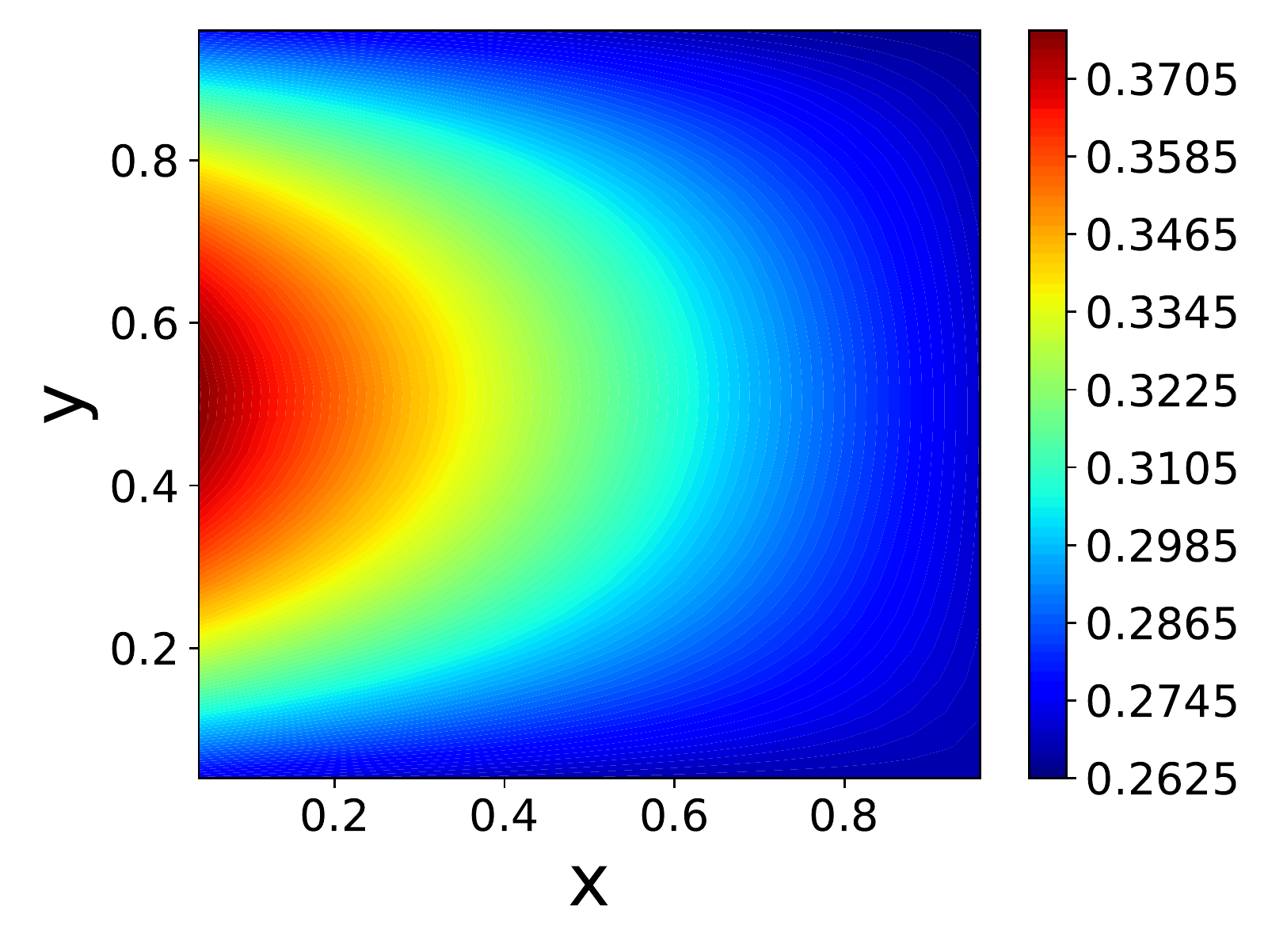}}\\
 \text{TL-DeepONet}&
 \raisebox{-.5\height}{\includegraphics[width=0.24\textwidth, height=0.12\textheight]{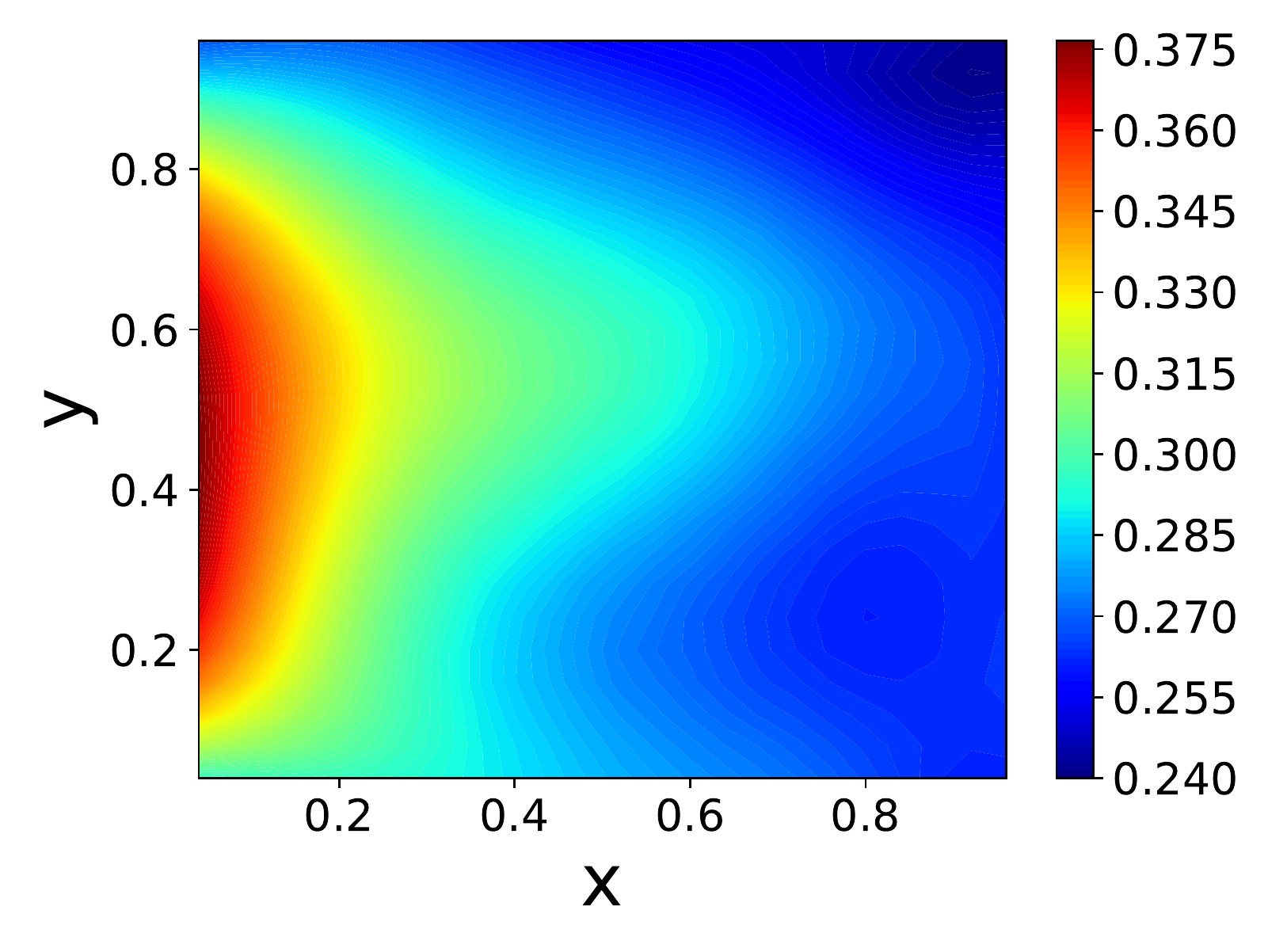}}&
 \raisebox{-.5\height}{\includegraphics[width=0.24\textwidth, height=0.12\textheight]{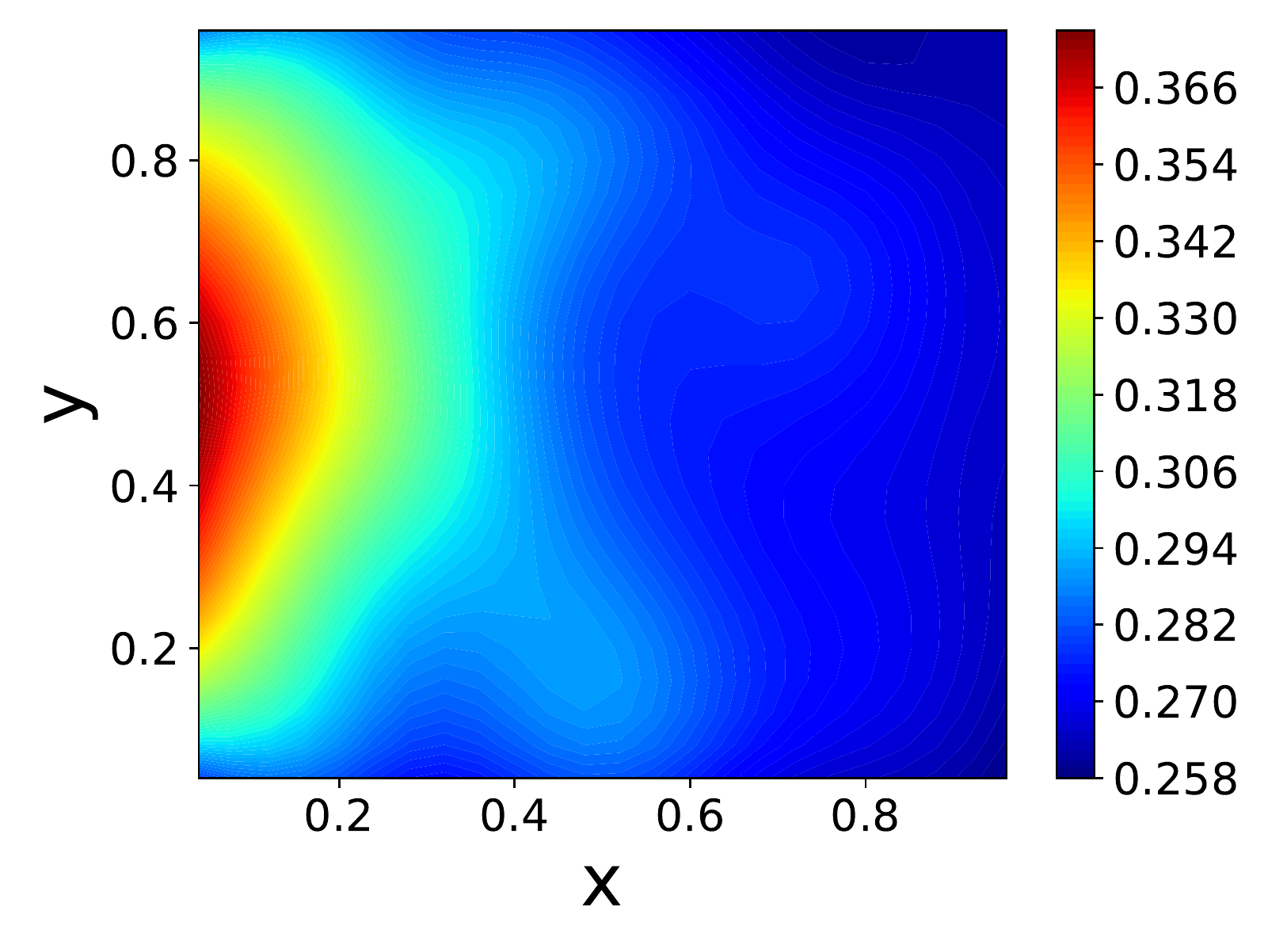}}&
 \raisebox{-.5\height}{\includegraphics[width=0.24\textwidth, height=0.12\textheight]{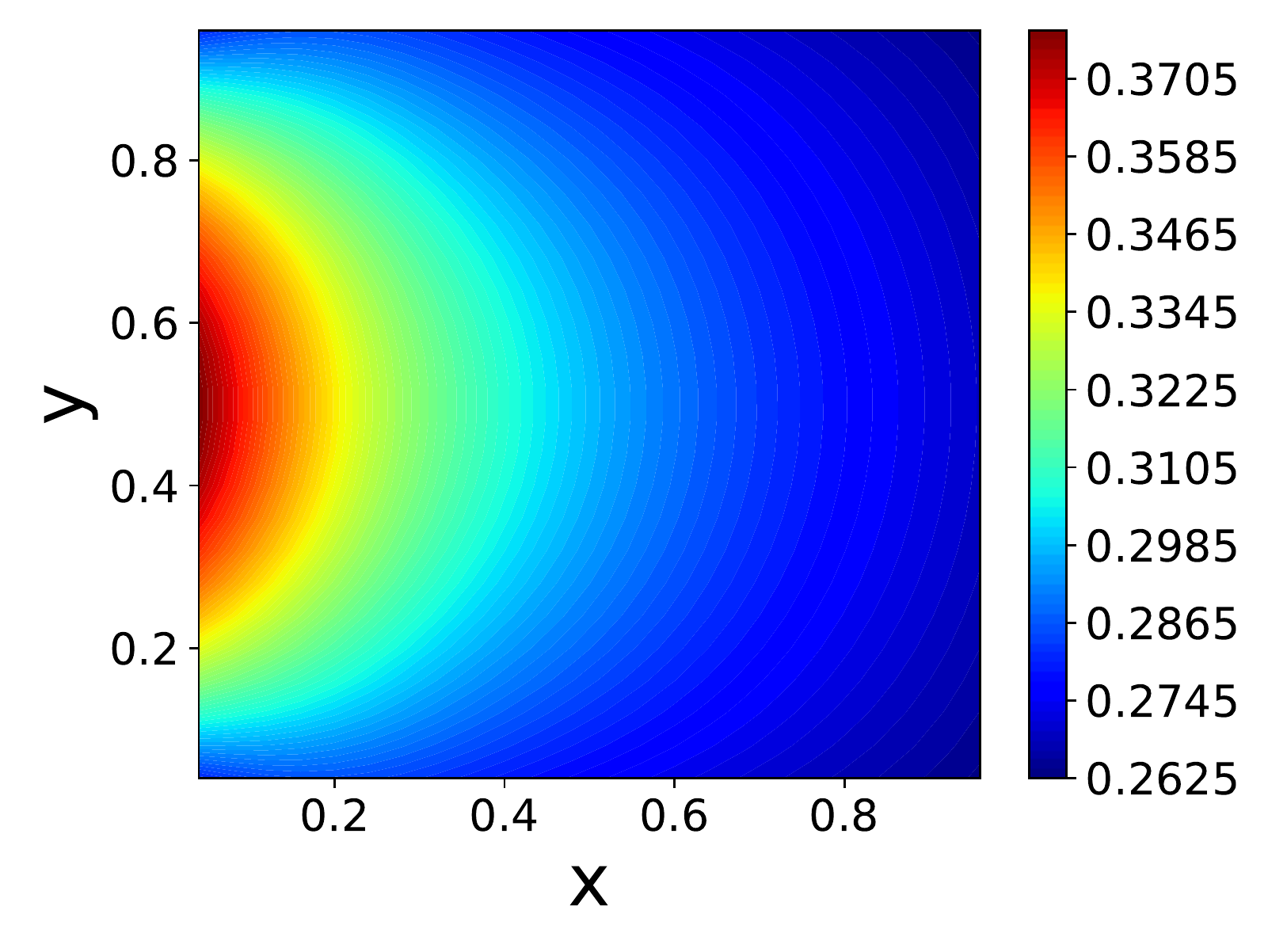}}\\
 \end{tabular}

 \begin{tabular}{cccc}
 \text{Initical condition} \quad & $t=0.5$ & $t=1$ & $t=10$\\
 \raisebox{-.5\height}{\includegraphics[width=0.24\textwidth, height=0.12\textheight]{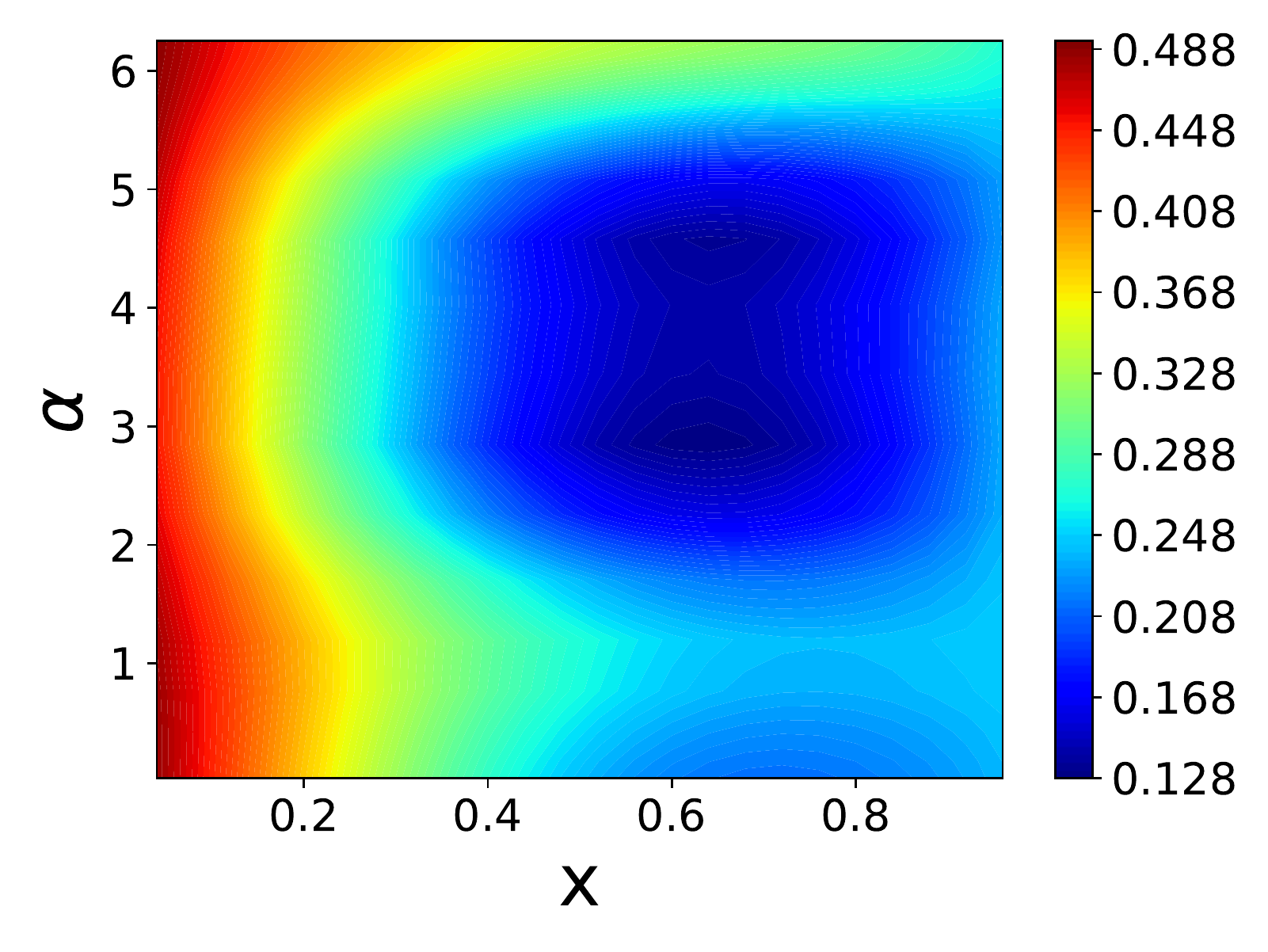}}&
 \raisebox{-.5\height}{\includegraphics[width=0.24\textwidth, height=0.12\textheight]{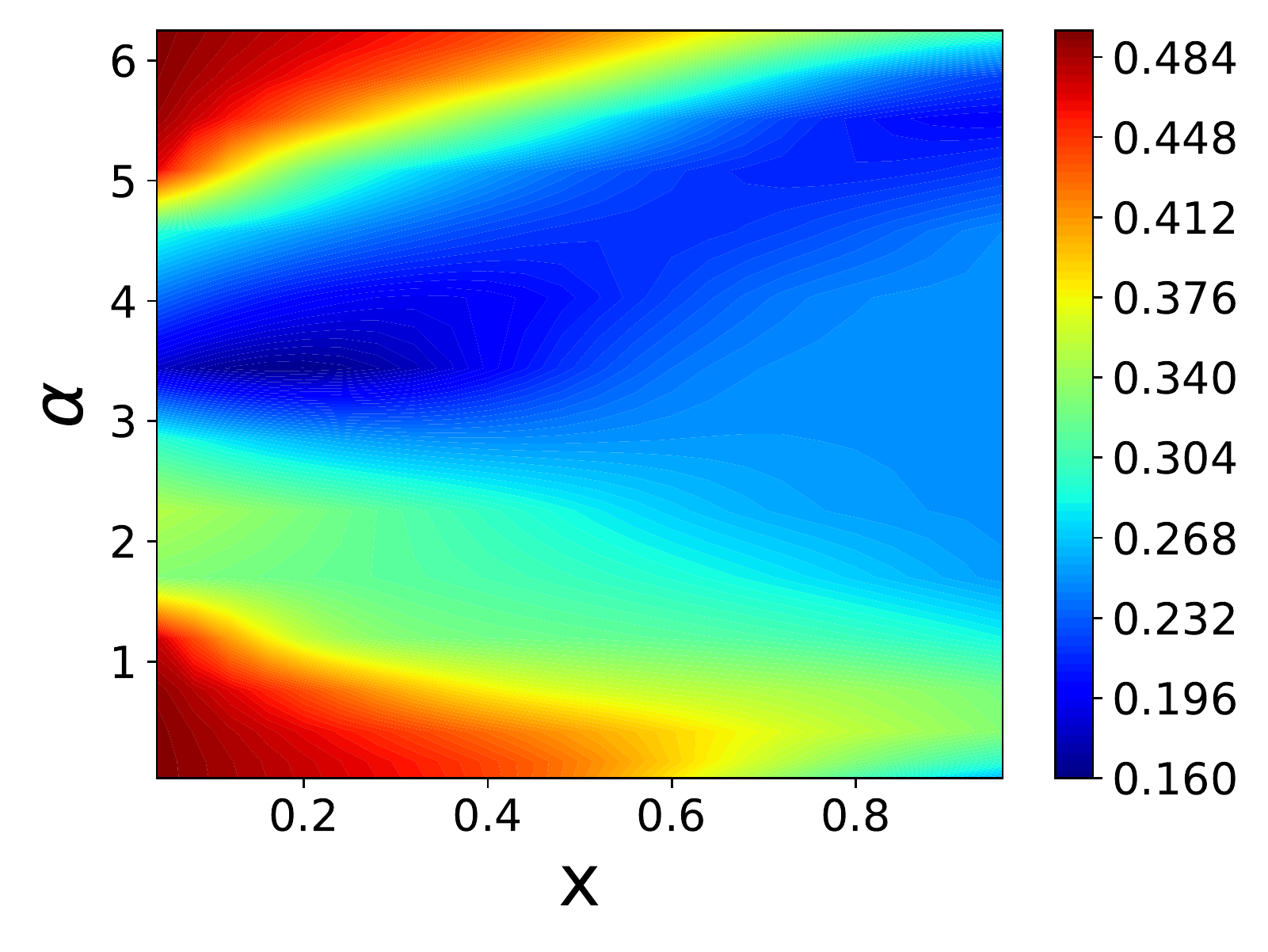}}&
 \raisebox{-.5\height}{\includegraphics[width=0.24\textwidth, height=0.12\textheight]{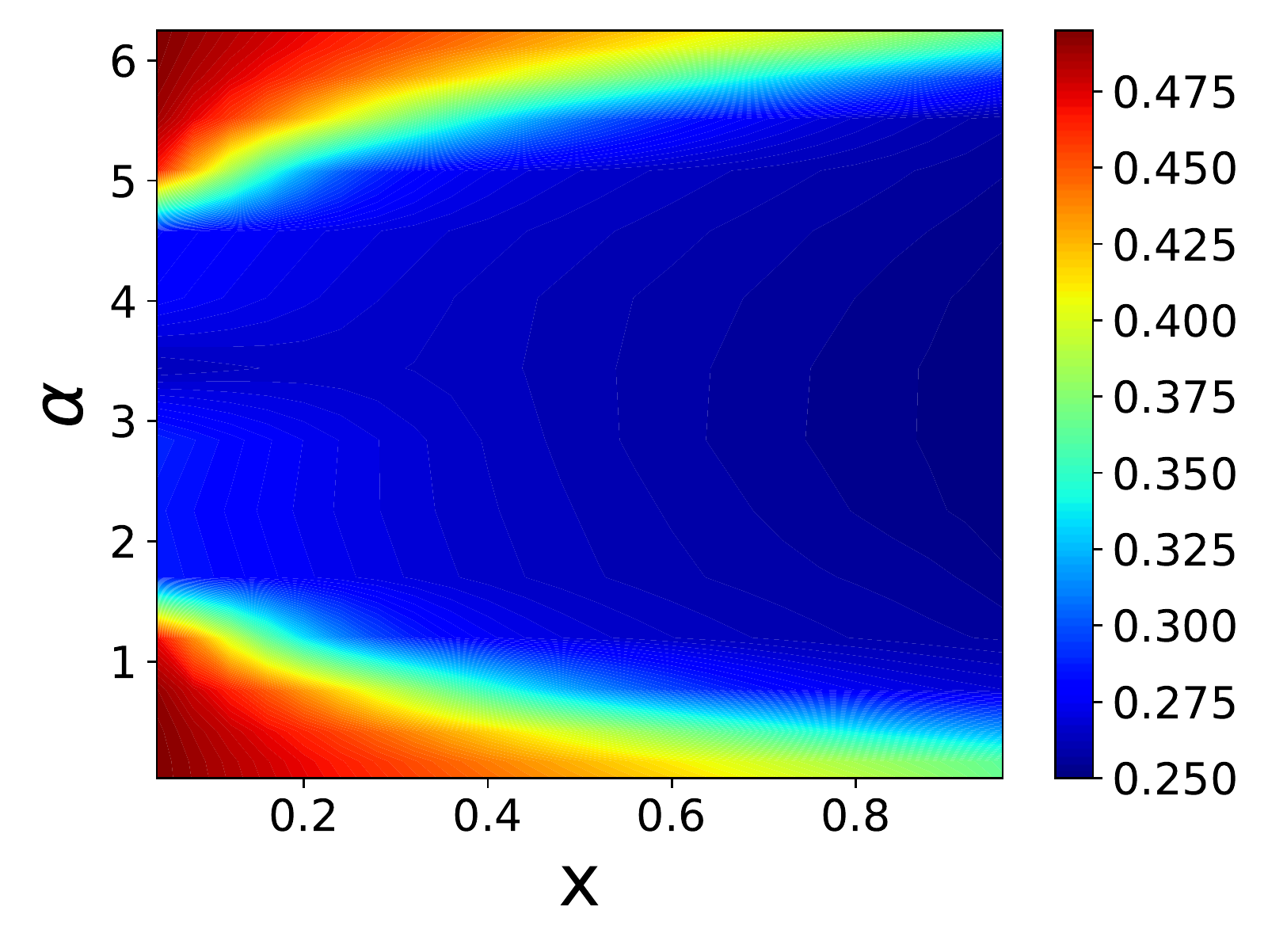}}&
 \raisebox{-.5\height}{\includegraphics[width=0.24\textwidth, height=0.12\textheight]{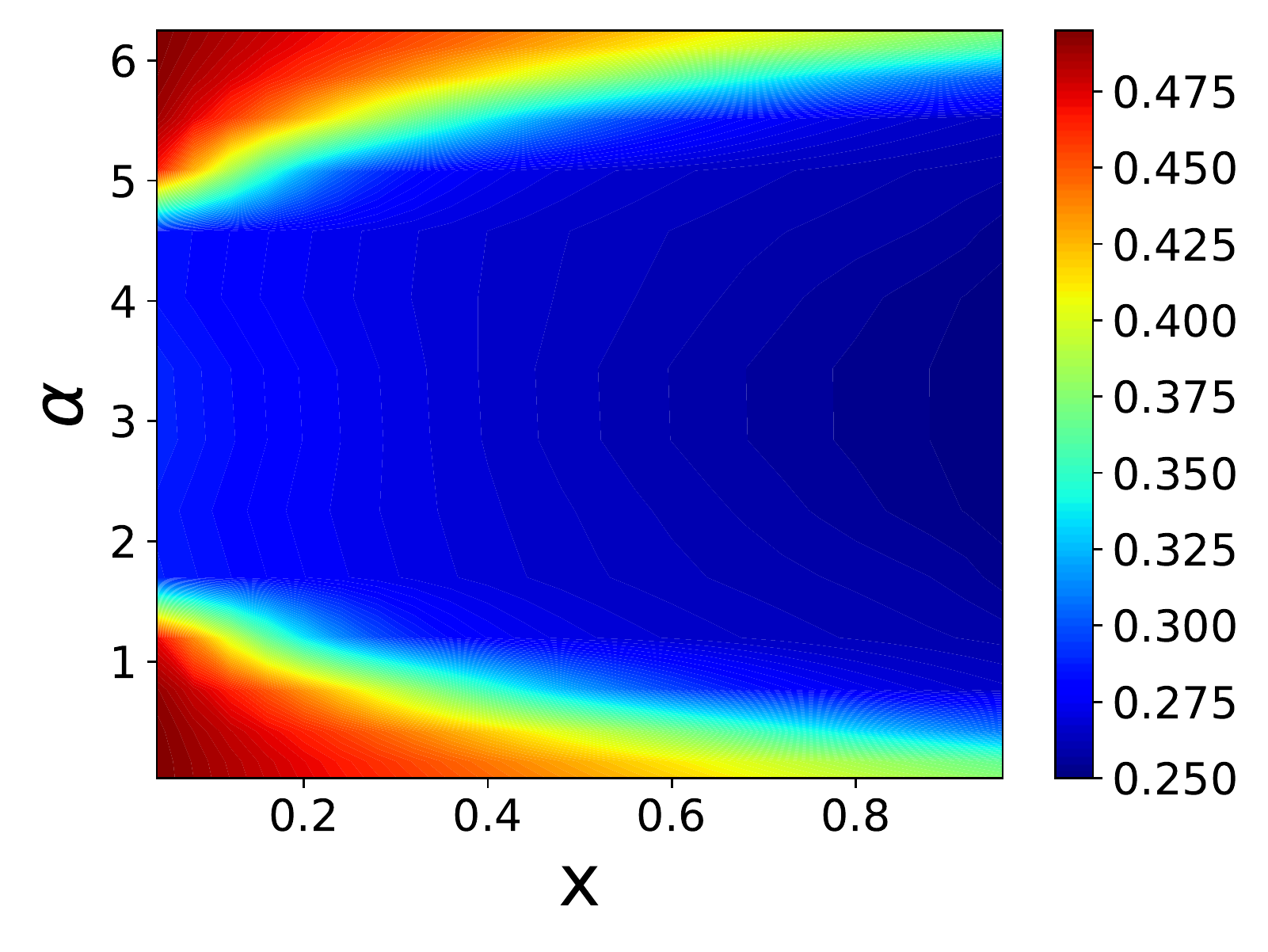}}\\
 \text{DeepONet}&
 \raisebox{-.5\height}{\includegraphics[width=0.24\textwidth, height=0.12\textheight]{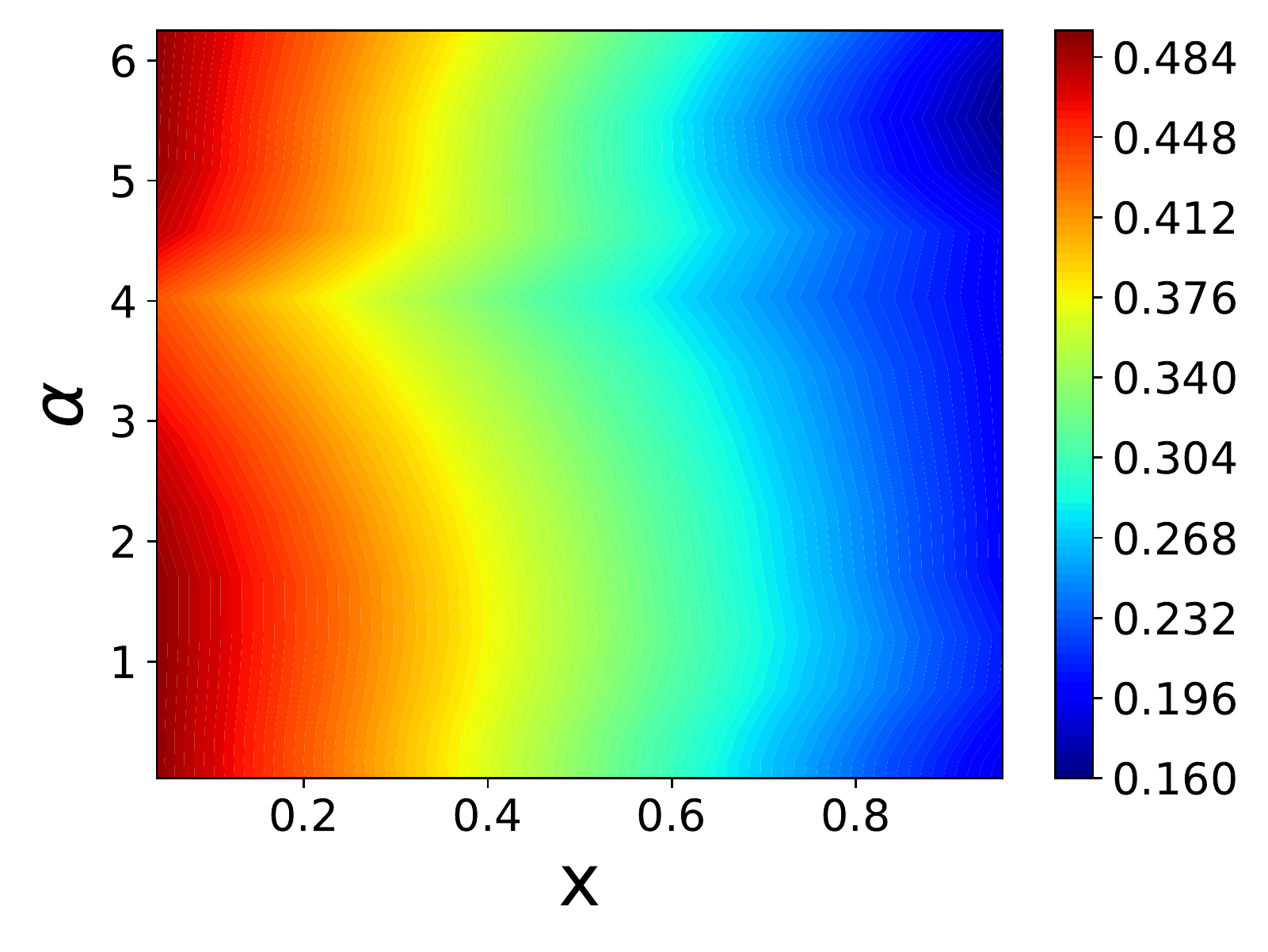}}&
 \raisebox{-.5\height}{\includegraphics[width=0.24\textwidth, height=0.12\textheight]{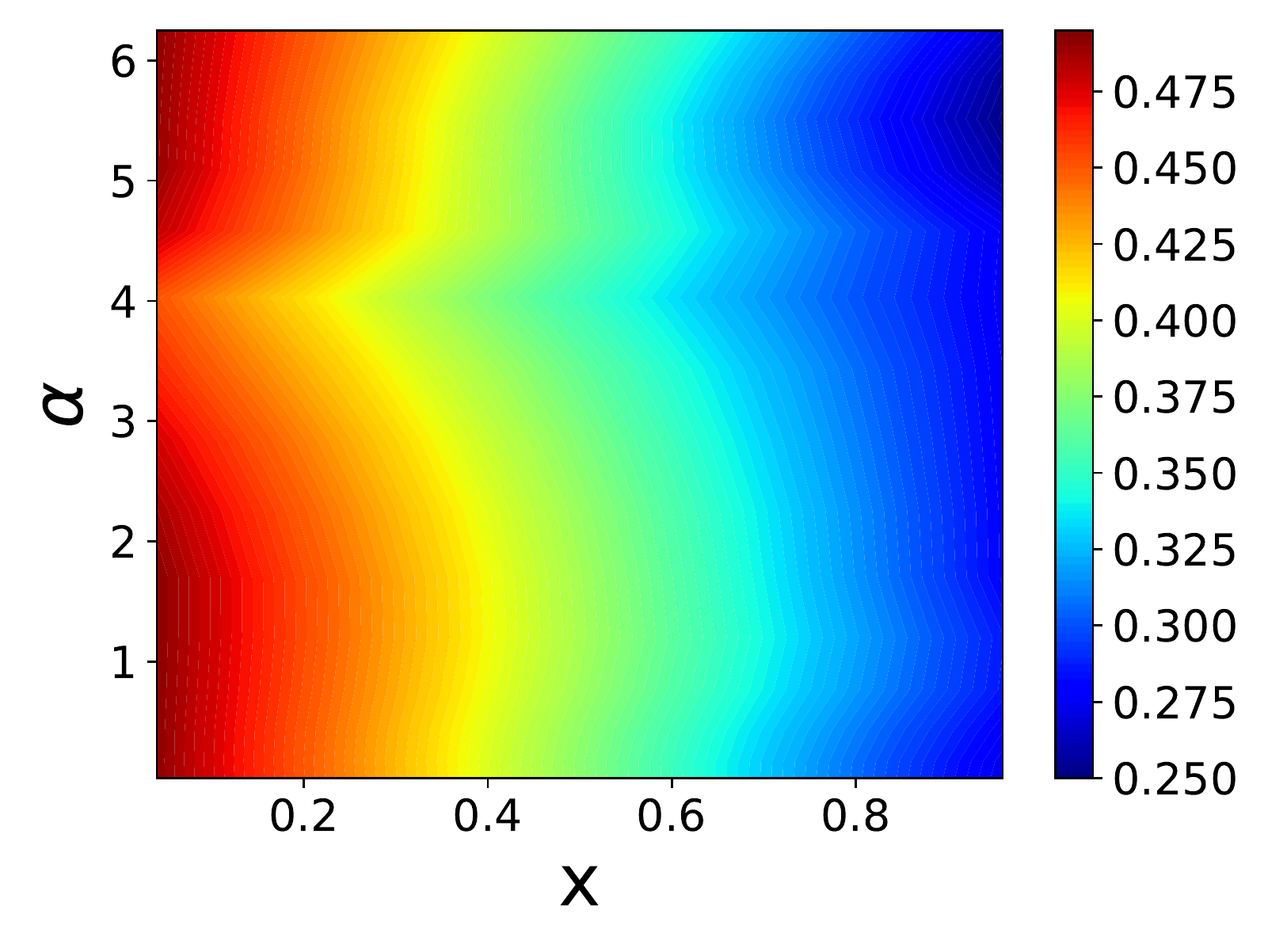}}&
 \raisebox{-.5\height}{\includegraphics[width=0.24\textwidth, height=0.12\textheight]{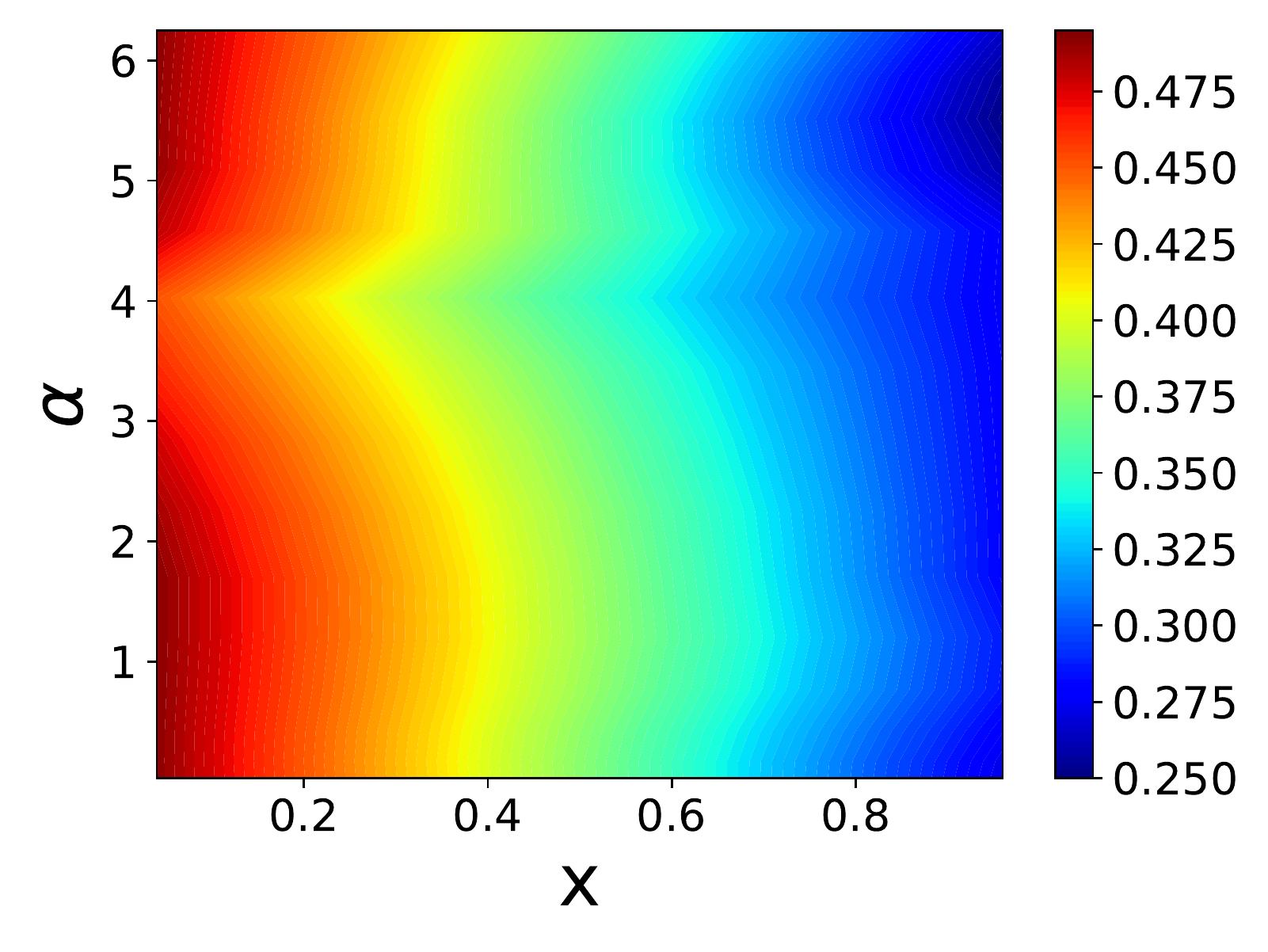}}\\
 \text{TL-DeepONet}&
 \raisebox{-.5\height}{\includegraphics[width=0.24\textwidth, height=0.12\textheight]{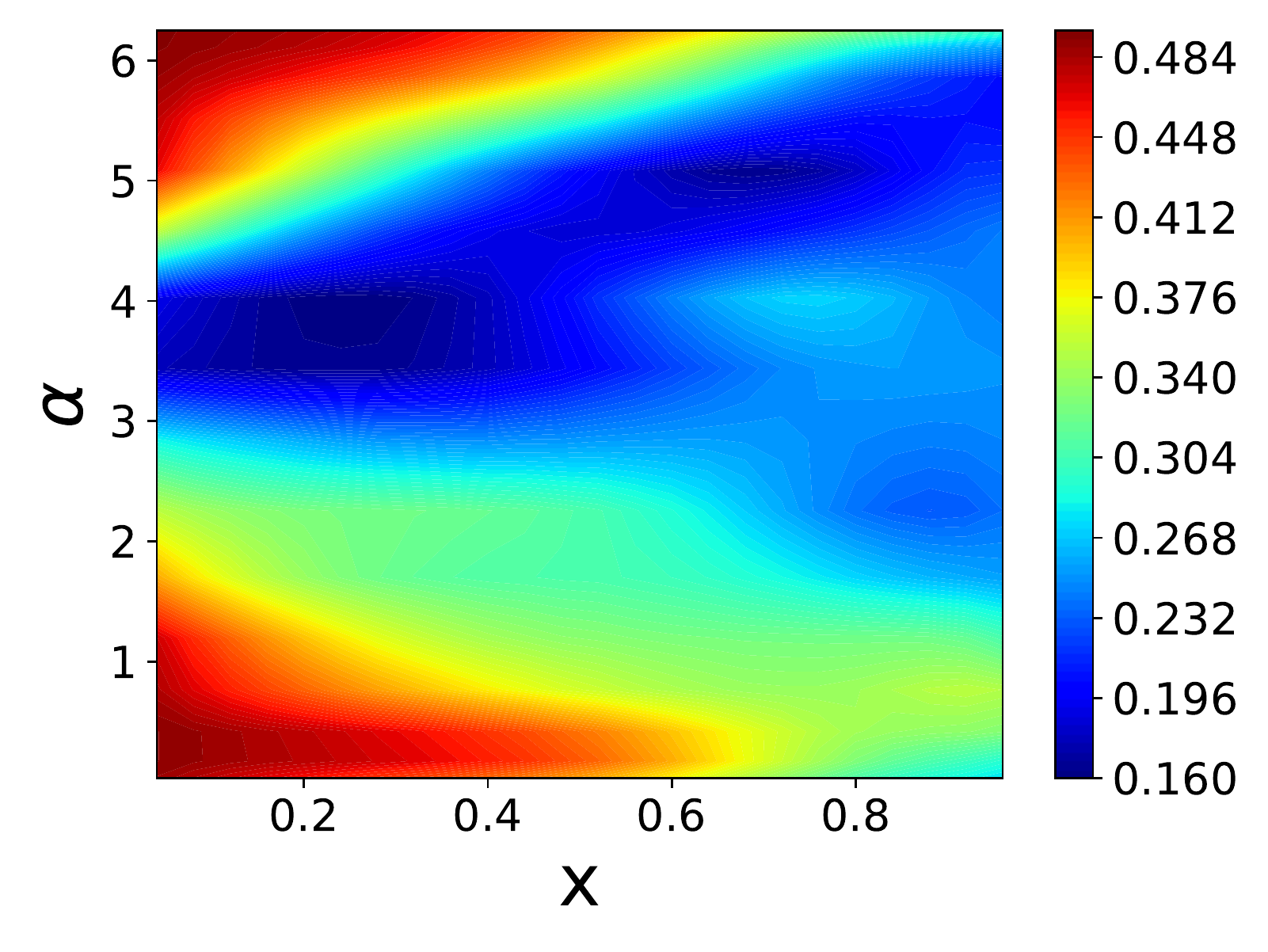}}&
 \raisebox{-.5\height}{\includegraphics[width=0.24\textwidth, height=0.12\textheight]{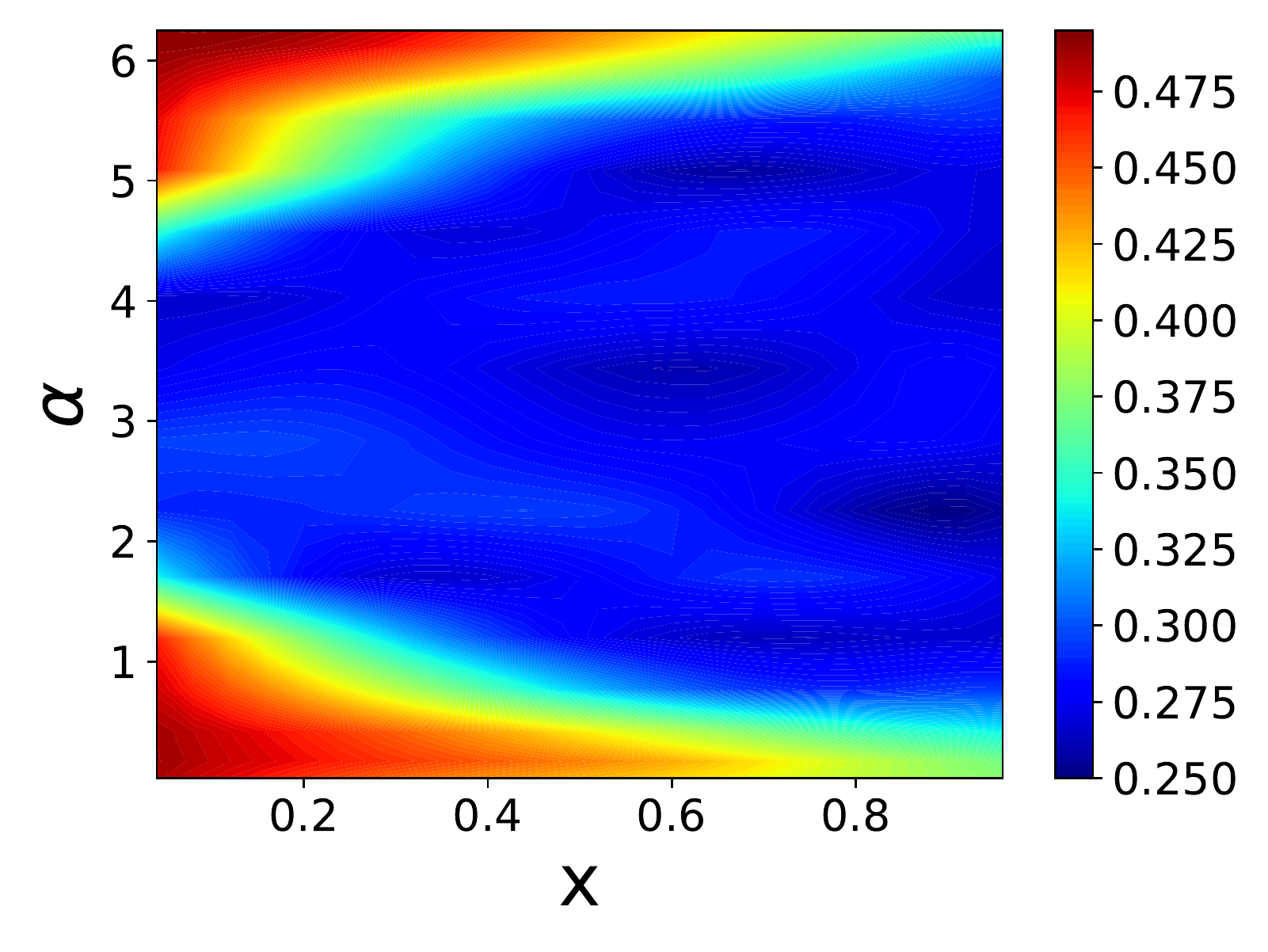}}&
 \raisebox{-.5\height}{\includegraphics[width=0.24\textwidth, height=0.12\textheight]{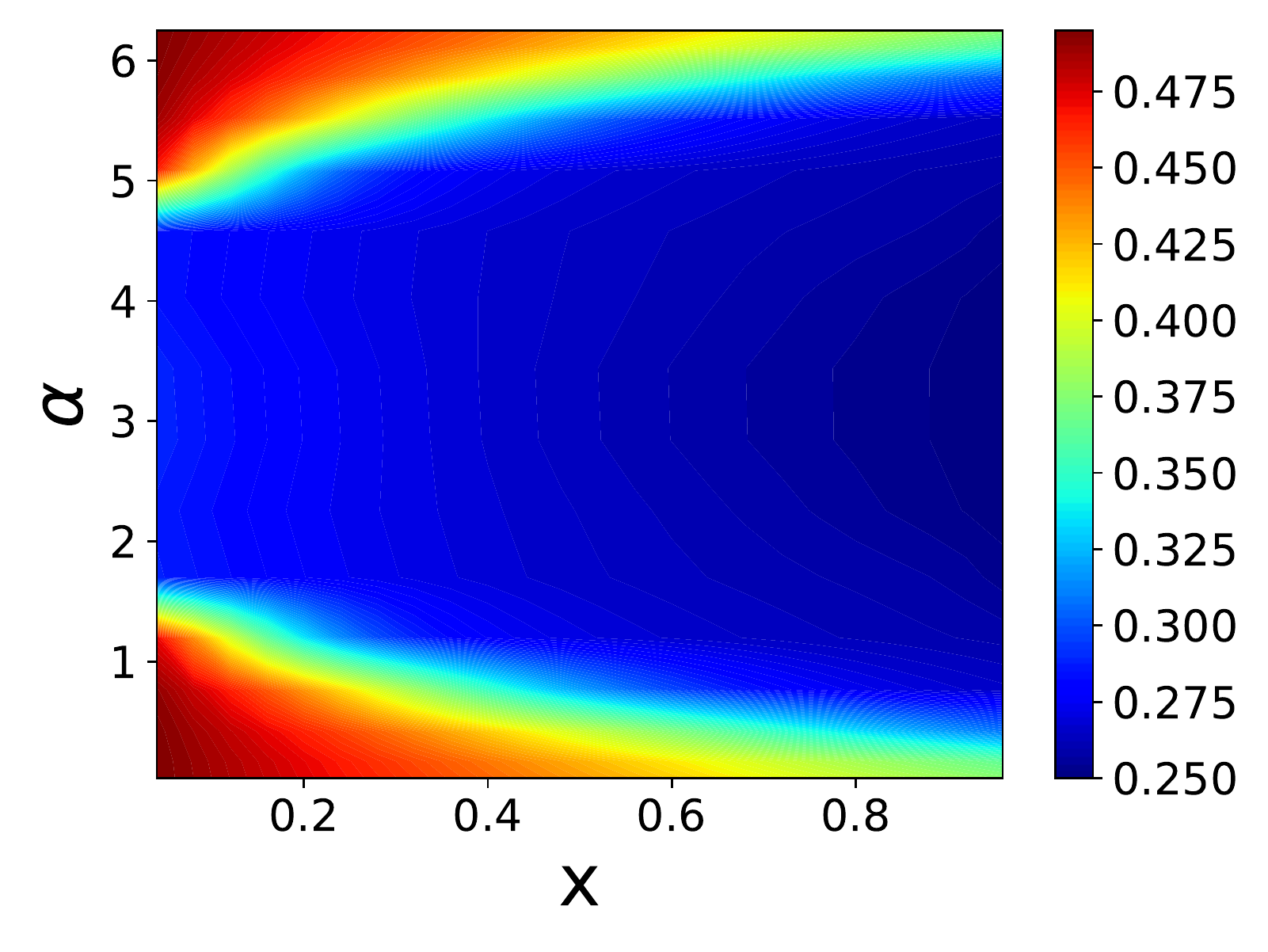}}\\
 \end{tabular}
 \caption{Results on 2D radiative transfer equation with $\eps=1$. The top three rows are snapshots of $\rho(t,x,y)$ of reference solutions, approximate solutions predicted by DeepONet and approximate solutions predicted by TL-DeepONet respectively. The bottom three rows are snapshots of $f(t, x,y=0.5,\alpha)$ of reference solutions, approximate solutions predicted by DeepONet and approximate solutions predicted by TL-DeepONet respectively.}
 \label{fig:rte_eps_1} 
 \end{figure*}

 Here we provide the snapshots of solution to \eqref{eq:rte} in 2D with $\eps=0.0001$ and $1$ in Figure~\ref{fig:rte_eps_zpzzz1} and Figure~\ref{fig:rte_eps_1}, respectively.

 \subsubsection{Implementation details on Table~\ref{tab:rte} and Figures~\ref{fig:rte_eps_zpzzz1},\ref{fig:rte_eps_1}}
 Consider multiscale radiative transfer equation \eqref{eq:rte}. In 1D case, we use $N_e=30$ test functions drawn from the Gaussian process defined above with the length scale $l=1$ in \eqref{eqn:kl_xv_1d}. We use $\Delta t = 0.01$ for time step size, $N_x=32$ uniform spatial grids for spatial discretization and $N_v = 16$ Gaussian quadrature points for velocity discretization. We set the maximum iteration number $N_{iter} = 200000$, adopt the stopping criterion that the empirical loss is below 1e-6 and use number of feature $p=100$. In the transfer learning step, we update $q = 40$ weights defined in \eqref{eqn:approx_trans}. In 2D case, we use $N_e=30$ test functions drawn from the Gaussian process defined above with the length scale $l=1$ in \eqref{eqn:kl_xv_2d}. We use $\Delta t = 0.01$ for time step size, $N_x=N_y = 24$ uniform spatial grids for spatial discretization and $N_v = 16$ Gaussian quadrature points for velocity discretization. We set the maximum iteration number $N_{iter} = 300000$, adopt the stopping criterion that the empirical loss is below 1e-6 and use number of feature $p=150$. In the transfer learning step, we update $q = 120$ weights defined in \eqref{eqn:approx_trans}.

\end{document}